\documentclass[11pt]{article}
\usepackage[top=30truemm,bottom=30truemm,left=25truemm,right=25truemm]{geometry}
\usepackage{parskip}
\usepackage[utf8]{inputenc}
\usepackage[T1]{fontenc}
\usepackage{tgtermes}
\usepackage{microtype}
\usepackage{amsmath}
\usepackage{amssymb}
\usepackage{bbm}
\usepackage{amsthm}
\usepackage{mathtools}

\usepackage[authoryear,round]{natbib}

\usepackage{xcolor}
\usepackage{float}
\usepackage[hypertexnames=false,backref,pagebackref]{hyperref}
\hypersetup{
  colorlinks=true,
  linkcolor=blue,
  citecolor=blue,
}
\usepackage[nameinlink,capitalise,noabbrev]{cleveref}
\newcommand{\fixcrefthm}[1]{%
  \AddToHook{env/#1/begin}{\crefalias{theorem}{#1}}%
}
\fixcrefthm{theorem}
\fixcrefthm{lemma}
\fixcrefthm{proposition}
\fixcrefthm{corollary}
\fixcrefthm{assumption}
\fixcrefthm{definition}
\fixcrefthm{remark}

\crefname{equation}{}{}
\crefname{theorem}{Theorem}{Theorems}
\Crefname{theorem}{Theorem}{Theorems}
\crefname{lemma}{Lemma}{Lemmas}
\Crefname{lemma}{Lemma}{Lemmas}
\crefname{corollary}{Corollary}{Corollaries}
\Crefname{corollary}{Corollary}{Corollaries}
\crefname{definition}{Definition}{Definitions}
\Crefname{definition}{Definition}{Definitions}
\crefname{remark}{Remark}{Remarks}
\Crefname{remark}{Remark}{Remarks}
\crefname{assumption}{Assumption}{Assumptions}
\Crefname{assumption}{Assumption}{Assumptions}
\crefname{proposition}{Proposition}{Propositions}
\Crefname{proposition}{Proposition}{Propositions}
\Crefname{algocf}{Algorithm}{Algorithms}
\crefname{algocf}{Algorithm}{Algorithms}
\crefname{appendix}{Appendix}{Appendices}
\Crefname{appendix}{Appendix}{Appendices}

\usepackage[ruled,vlined,linesnumbered]{algorithm2e}
\SetArgSty{textnormal}
\SetKwInOut{KwRequire}{Require}
\crefname{line}{Line}{Lines}
\Crefname{line}{Line}{Lines}
\renewcommand{\nllabel}[1]{%
  \addtocounter{AlgoLine}{-1}%
  \refstepcounter{AlgoLine}%
  \label[line]{#1}%
}

\usepackage{graphicx}
\graphicspath{{figure/}}
\usepackage[labelfont=bf]{caption}
\usepackage{enumitem}
\usepackage{booktabs}
\usepackage{pdflscape}

\usepackage{wrapfig}
\usepackage{tikz}

\theoremstyle{plain}
\newtheorem{theorem}{Theorem}[section]
\newtheorem{lemma}[theorem]{Lemma}

\theoremstyle{definition}
\newtheorem{definition}[theorem]{Definition}
\theoremstyle{remark}

\renewcommand{\epsilon}{\varepsilon}
\renewcommand{\hat}{\widehat}
\def\E{\mathbb{E}}
\def\P{\mathbb{P}}
\def\R{\mathbb{R}}
\def\N{\mathbb{N}}
\def\I{\mathbb{I}}
\newcommand{\ind}{\mathbbm{1}}

\DeclareMathOperator*{\argmin}{arg\,min}

\DeclareMathOperator{\Reg}{Reg}
\DeclareMathOperator{\Unif}{Unif}
\DeclarePairedDelimiter{\abs}{\lvert}{\rvert}
\DeclarePairedDelimiter{\inpr}{\langle}{\rangle}
\DeclarePairedDelimiter{\brk}{[}{]}
\DeclarePairedDelimiter{\prn}{(}{)}
\DeclarePairedDelimiter{\set}{\{}{\}}
\DeclarePairedDelimiter{\nrm}{\|}{\|}
\DeclarePairedDelimiter{\ceil}{\lceil}{\rceil}
\newcommand{\calA}{\mathcal{A}}
\newcommand{\calE}{\mathcal{E}}
\newcommand{\calF}{\mathcal{F}}
\newcommand{\calK}{\mathcal{K}}
\newcommand{\calM}{\mathcal{M}}

\newcommand{\calQ}{\mathcal{Q}}

\newcommand{\calS}{\mathcal{S}}

\newcommand{\pio}{\mathring{\pi}}
\newcommand{\qo}{\mathring{q}}
\newcommand{\Pbar}{\bar{P}}

\allowdisplaybreaks[1]
\usepackage{autonum}
\usepackage[normalem]{ulem}

\newcommand{\given}{\;\middle|\;}

\usepackage{tabularx,array}
\title{Minimax Optimal Regret Bound for Adversarial MDPs\\
with Unknown Transitions}
\author{
  Mingyi Li\footnote{
    The University of Tokyo;
    \texttt{mingyi-mike@g.ecc.u-tokyo.ac.jp}.
  }
  \and
  Taira Tsuchiya\footnote{
    The University of Tokyo, The University of Osaka, and RIKEN;
    \texttt{tsuchiya@mist.i.u-tokyo.ac.jp}.
  }
}
\date{}

\begin{document}
\maketitle

\begin{abstract}
We study online learning in episodic tabular Markov decision processes (MDPs) with adversarial losses and unknown transitions under bandit feedback.
For this problem, the best known regret bounds are $\widetilde O(HS\sqrt{AT})$ and $\widetilde O(H^{5/2}\sqrt{SAT})$, where $H$ is the horizon, $S$ the number of states, $A$ the number of actions, and $T$ the number of episodes.
However, these bounds do not match the lower bound $\Omega(H\sqrt{SAT})$, and it has long been unknown whether this gap can be closed.
We close this gap by giving the first algorithm that attains regret $\widetilde O(H\sqrt{SAT})$ with high probability, which matches the lower bound up to logarithmic factors.
Our algorithm runs follow-the-regularized-leader (FTRL) over occupancy measures and avoids the two sources of the extra factor $\sqrt S$ in existing algorithms of this type.
First, instead of bounding the mismatch between the estimated and true occupancy measures by the width of a confidence set of transitions, we estimate the loss due to this mismatch by temporal-difference errors and add the estimate to the loss that FTRL uses.
As a result, a fixed bound on the relative error of the transition probability estimates suffices for the regret guarantee.
Second, instead of dividing each observed loss by an upper occupancy bound over a confidence set, we deliver the loss to FTRL gradually over later episodes, refining geometric resampling with rejection.
\end{abstract}
\section{Introduction}
Markov decision processes (MDPs) are a standard model of reinforcement learning and describe how the distribution of the next state depends on the current state and the action taken in that state.
We study episodic tabular MDPs with unknown transitions and adversarial losses under bandit feedback, where the learner observes losses only at the state-action pairs it visits.
In each episode, the learner chooses a policy that assigns a distribution over actions to each state, follows it from the initial state under a fixed but unknown transition kernel, and observes the resulting trajectory.
The learner aims to minimize the regret, which is the difference between the sum of the expected losses of the learner's policies over all episodes and that of the best fixed policy in hindsight.

For MDPs with horizon $H$, $S$ states, and $A$ actions over $T$ episodes, many existing algorithms optimize over occupancy measures.
When the transition kernel is known, \citet{zimin2013online} attain the minimax-optimal regret $\widetilde O(\sqrt{HSAT})$.
When the transition kernel is unknown, the learner also has to learn the transitions, and one of the best known bounds in this case is $\widetilde O(HS\sqrt{AT})$ of \citet{jin2020learning}.
The same order is also attained by \citet{li2026closing} with policy optimization, which optimizes directly over action distributions at each state.
These bounds are larger by a factor of $\sqrt S$ than the lower bound $\Omega(H\sqrt{SAT})$, which holds even under full-information feedback~\citep{jin2018q,domingues2021episodic}, and it has long been unknown whether this gap can be closed.

Recently, \citet{kulkarni2026introducing} attained regret $\widetilde O(H^{5/2}\sqrt{SAT})$ under bandit feedback by delaying the use of observed losses, at the cost of a worse dependence on $H$.
Their bound has the optimal dependence on $S$, while the bound of \citet{jin2020learning} has the optimal dependence on $H$, which raises the following question.
\begin{center}
\emph{For adversarial MDPs with unknown transitions, can we construct an algorithm
that attains a regret upper bound matching the lower bound?}
\end{center}

\begin{table}[t]
\centering
\small
\setlength{\tabcolsep}{8pt}
\setlength{\belowcaptionskip}{8pt}
\setlength{\parskip}{0pt}
\renewcommand{\arraystretch}{1.2}
\caption{Regret bounds for episodic tabular MDPs with adversarial losses and unknown transitions under bandit feedback. Logarithmic factors and terms independent of $T$ are omitted.}
\begin{tabular}{@{}lcc@{}}
\toprule
Reference & Regret & Approach \\
\midrule
\citet{jin2020learning} & $HS\sqrt{AT}$ & Optimization over occupancy measures \\
\citet{luo2021policy} & $H^2S\sqrt{AT}$ & Policy optimization \\
\citet{li2026closing} & $HS\sqrt{AT}$ & Policy optimization \\
\citet{kulkarni2026introducing} & $H^{5/2}\sqrt{SAT}$ & Policy optimization \\
\midrule
\textbf{This work} & $H\sqrt{SAT}$ & Optimization over occupancy measures \\
\midrule
Lower bound~\citep{jin2018q,domingues2021episodic} & $H\sqrt{SAT}$ & --- \\
\bottomrule
\end{tabular}
\label{tab:comparison}
\end{table}

\subsection{Contributions of this paper}\label{sec:contributions}
We answer this question affirmatively.
Our algorithm achieves, with probability at least $1-11\delta$,
\begin{align}
\Reg_T\lesssim H\sqrt{SAT}\log^2\prn*{\frac{SAT}{\delta}}+H^2S^3A^2\max\set{SA,H^2}\log\prn*{\frac{SA}{\delta}}.
\end{align}
This is the first bound that matches the lower bound $\Omega(H\sqrt{SAT})$, and it improves the bound of \citet{jin2020learning} by a factor of $\sqrt S$ and that of \citet{kulkarni2026introducing} by a factor of $H^{3/2}$ (\cref{tab:comparison}).
Our algorithm builds on follow-the-regularized-leader (FTRL) over occupancy measures, which are also used by \citet{jin2020learning}, and the improvement comes from the two ideas below.

The first idea is to incorporate the transition estimation error into the loss that FTRL uses to optimize over occupancy measures.
Existing methods that optimize over occupancy measures separately analyze the regret of the online learning algorithm and the loss due to the mismatch between the estimated and true occupancy measures.
Using the width of a transition confidence set to bound the latter introduces an extra factor of $\sqrt S$.
Our treatment of the transition estimation error has three components.
First, we use temporal-difference (TD) errors to estimate the loss due to the mismatch.
By feeding this estimate to FTRL, we can account for that loss in the FTRL regret bound.
Second, since a fixed bound on the relative error of the transition probability estimates suffices for the regret guarantee, we estimate the transition probability of each state-action-next-state triple after observing that transition a set number of times and reuse the estimate thereafter.
Third, since the TD errors enlarge the stability term of FTRL, we control it by a hybrid regularizer that adds a logarithmic barrier to the negative entropy.

The second idea is to deliver each observed loss to FTRL gradually over later episodes.
Under bandit feedback, a loss is usually estimated by dividing it by the probability of visiting its state-action pair, but this probability depends on the unknown transitions.
Previous algorithms replace this probability by an upper occupancy bound over a confidence set of transitions~\citep{jin2020learning}, and bounding the resulting bias in the loss estimate also gives an extra factor of $\sqrt S$ in the regret bound.
\citet{kulkarni2026introducing} instead use geometric resampling with rejection to estimate the inverse visitation probability from the waiting time for a revisit in later episodes, spreading the loss estimate over the policy updates made during that wait.
We refine this approach by keeping a weight for each observed loss.
After each later episode, we use its trajectory to estimate the probability of visiting the same state-action pair under the policy used when we observed the loss, and reduce the weight using that estimate.
This lets us give FTRL the loss multiplied by its current weight in each later episode without the rules that \citet{kulkarni2026introducing} use to decide when resampling estimates start and stop and how many can run at once.
As a result, summing the weighted contributions to FTRL over later episodes gives a loss estimate whose expectation is at most the original loss, just like the importance-weighted estimator with the true visitation probability.
That said, the visitation probability estimate can become large when the current policy is far from the policy used when we observed the loss.
In that case, we update the weight using an upper bound on the visitation probability, and the resulting contribution to regret is absorbed into the leading term.
The hybrid regularizer also keeps the relative change of the occupancy measure small at every state-action pair, which bounds the cost of the gradual delivery.

\section{Related work}
Early work by \citet{even2009online} and \citet{yu2009Markov} studied infinite-horizon adversarial MDPs.
For episodic MDPs with known transitions, \citet{zimin2013online} optimized over the set of all occupancy measures and attained the minimax regret rate $\widetilde O(\sqrt{HSAT})$ under bandit feedback.
With unknown transitions, the same approach yields regret $\widetilde O(HS\sqrt{AT})$ under full-information feedback~\citep{rosenberg2019online} and bandit feedback~\citep{jin2020learning}.
\citet{dai2022follow} also achieved the same order of regret under bandit feedback using follow-the-perturbed-leader.
\citet{li2026closing} used policy optimization to achieve regret $\widetilde O(HS\sqrt{AT})$ under bandit feedback, improving the $\widetilde O(H^2S\sqrt{AT})$ bound of \citet{luo2021policy}.
These approaches have been widely used in subsequent work to obtain best-of-both-worlds guarantees~\citep{jin2021best,dann2023best} and data-dependent regret bounds~\citep{li2026data,li2026policy}.
The regret lower bound, on the other hand, is $\Omega(H\sqrt{SAT})$~\citep{jin2018q,domingues2021episodic}, and narrowing the $\sqrt S$ gap between this lower bound and the $\widetilde O(HS\sqrt{AT})$ upper bound was a long-standing open problem.

The $\sqrt S$ gap was first narrowed under full-information feedback by \citet{tiapkin2025narrowing}, who obtained a regret rate of $\widetilde O(H^3\sqrt{SAT})$.
\citet{schlisselberg2026improved} later improved the dependence on $H$ to obtain $\widetilde O(H^{3/2}\sqrt{SAT})$.
\citet{kulkarni2026introducing} showed that $\sqrt{S}$ dependence is also attainable under bandit feedback, with regret rate $\widetilde O(H^{5/2}\sqrt{SAT})$.
These results, all obtained using policy optimization, match the lower bound in their dependence on $S$, with a gap in $H$ remaining even under full-information feedback.
With this gap still unresolved for policy optimization, we use FTRL over occupancy measures to obtain a leading regret term of $\widetilde O(H\sqrt{SAT})$ under bandit feedback.

To obtain $\sqrt{S}$ dependence under bandit feedback, \citet{kulkarni2026introducing} combine rejection sampling with geometric resampling~\citep{neu2016importance}.
The expected number of episodes until rejection sampling accepts a trajectory that revisits the same state-action pair is proportional to the inverse visitation probability.
They incorporate the loss estimate gradually into policy updates as resampling proceeds and analyze the resulting delay using techniques for delayed bandit feedback in MDPs~\citep{jin2022near,lancewicki2023delay}.

\section{Preliminaries}\label{sec:prelim}
For $N\in\N$, let $[N]=\{1,\ldots,N\}$. For a finite set $\calK$, let $\Delta(\calK)$ denote the set of probability distributions on $\calK$. We denote the indicator of an event $E$ by $\ind\{E\}$, use $\log$ for the natural logarithm, and write $\inpr{x,y}=\sum_ix(i)y(i)$. For nonnegative functions $f$ and $g$, we write $f\lesssim g$ or $g\gtrsim f$ if $f\leq Cg$ for some universal constant $C>0$, and use $\widetilde O(\cdot)$ to hide polylogarithmic factors.
Let $\calF_t$ be the sigma-algebra generated by the history up to the end of episode $t-1$, and write $\E_t[\cdot]=\E[\cdot\mid\calF_t]$.

\paragraph{Episodic tabular MDPs.}
We study an episodic tabular MDP $\calM=(\calS,\calA,P,H,s_0)$, where $\calS$ and $\calA$ are the finite state and action spaces of sizes $S=\abs{\calS}$ and $A=\abs{\calA}$, respectively, $H$ is the horizon, and $s_0$ is the initial state.
After the learner takes action $a$ in state $s$, the next state is drawn from $P(\cdot\mid s,a)$, where the transition kernel $P$ is fixed across episodes and unknown to the learner.
We use the standard layered formulation~\citep[\textit{e.g.,}][]{neu2010online,jin2020learning,luo2021policy}, in which $\calS$ is partitioned into $H$ nonempty layers $\calS_0,\ldots,\calS_{H-1}$ ($S\geq H$), with $\calS_0=\{s_0\}$.
The terminal layer $\calS_H=\{s_H\}$ consists of a single absorbing state.
We write $h(s)=h$ for $s\in\calS_h$, and for every $(s,a)\in\calS_h\times\calA$ with $h<H$, $P(\cdot\mid s,a)$ is supported on $\calS_{h+1}$.
We call each triple $(s,a,s')\in\calS_h\times\calA\times\calS_{h+1}$ with $h<H$ an \emph{edge}, and there are at most $S^2A$ edges.

A policy $\pi:\calS\to\Delta(\calA)$ assigns probability $\pi(a\mid s)$ to action $a$ in state $s$.
The learner interacts with the MDP for $T$ episodes, with the loss functions $\ell_1,\ldots,\ell_T:\calS\times\calA\to[0,1]$ fixed in advance by the adversary.
At the start of episode $t$, the learner selects $\pi_t$ based on past observations and follows it under $P$ from $s_{t,0}=s_0$ to $s_{t,H}=s_H$.
Under bandit feedback, the learner observes the trajectory and the loss at each visited state-action pair, given by $\{(s_{t,h},a_{t,h},\ell_t(s_{t,h},a_{t,h}))\}_{h=0}^{H-1}$.

We call $\widetilde P$ a transition subkernel if its entries are nonnegative and $\sum_{s'}\widetilde P(s'\mid s,a)\leq1$ for every $(s,a)$. Under $\widetilde P$, after taking action $a$ in state $s$, the episode ends with probability $1-\sum_{s'}\widetilde P(s'\mid s,a)$.
Given $\widetilde P$, a policy $\pi$, and a function $f:\calS\times\calA\to\R$, we define the value function $V^{\widetilde P,\pi}(s;f)$ and the $Q$-function $Q^{\widetilde P,\pi}(s,a;f)$ by the backward recursion below, with the value function set to zero at termination, including $V^{\widetilde P,\pi}(s_H;f)=0$.
\begin{align}
V^{\widetilde P,\pi}(s;f)&=\E_{a\sim\pi(\cdot\mid s)}\brk*{Q^{\widetilde P,\pi}(s,a;f)},\qquad
Q^{\widetilde P,\pi}(s,a;f)=f(s,a)+\sum_{s'}\widetilde P(s'\mid s,a)V^{\widetilde P,\pi}(s';f).
\end{align}
The state-action occupancy measure $q^{\widetilde P,\pi}$ gives the probability $q^{\widetilde P,\pi}(s,a)$ of visiting $(s,a)$ when following $\pi$ from $s_0$ under $\widetilde P$, with corresponding state and edge occupancy measures $q^{\widetilde P,\pi}(s)=\sum_a q^{\widetilde P,\pi}(s,a)$ and $q^{\widetilde P,\pi}(s,a,s')=q^{\widetilde P,\pi}(s,a)\widetilde P(s'\mid s,a)$, respectively.
Then we can write the value function at $s_0$ as $V^{\widetilde P,\pi}(s_0;f)=\sum_{s,a}q^{\widetilde P,\pi}(s,a)f(s,a)$.
When $\widetilde P=P$, we write $V^\pi$, $Q^\pi$, and $q^\pi$.
The learner's goal is to minimize regret against the best fixed policy in hindsight,
\begin{align}\label{eq:regret}
\Reg_T\coloneqq\sum_{t=1}^T V^{\pi_t}(s_0;\ell_t)-\min_{\pi}\sum_{t=1}^T V^{\pi}(s_0;\ell_t).
\end{align}
Let $\pio$ be a fixed policy that minimizes $\sum_{t=1}^T V^\pi(s_0;\ell_t)$.

\section{Algorithm}\label{sec:algorithm}
We present our algorithm in \cref{alg:main}.
In each episode, the algorithm computes a policy by FTRL over occupancy measures, executes it, and uses the observed trajectory to estimate the transitions and to construct the next loss vector.
We describe the estimation of the transitions in \cref{sec:calibration}, the FTRL update in \cref{sec:ftrl}, and the construction of the loss vector in \cref{sec:delivery,sec:gradient}.

\begingroup
\csletcs{equation}{autonum@equationOld}
\csletcs{endequation}{autonum@endequationOld}
\begin{algorithm}[!t]
\caption{FTRL over occupancy measures with gradual loss delivery}\label{alg:main}
\DontPrintSemicolon
\KwRequire{
$\iota=\log(\mathrm{e}SAT/\delta)$, $c_0=256$, $\epsilon=1/(100\sqrt{HSA\max\set{SA,H^2}})$, $B=\ceil{25\iota/\epsilon^2}$, \\
$\eta=1/(2500c_0(\sqrt{SAT}+\sqrt{HS^3}A)\iota)$, $\gamma=1/(\sqrt{SAT}+SA)$, and $\calE_1=\emptyset$.}
\For{$t=1,\ldots,T$}{
  Compute
  \begin{equation}
    \hat q_t=\argmin_{q\in\calQ(\calE_t)}
      \set*{\inpr*{\sum_{k=1}^{t-1}\hat g_k,q}+\frac{\Psi_{\calE_t}(q)}{\eta}}\notag
  \end{equation}
  using the regularizer $\Psi_{\calE_t}$ in \cref{eq:regularizer}.\nllabel{line:ftrl}\;
  Compute $\pi_t(a\mid s)\propto\hat q_t(s,a)$ and
  $\hat P_{\calE_t}(s'\mid s,a)=\hat q_t(s,a,s')/\hat q_t(s,a)$ on $\calE_t$, with $\hat P_{\calE_t}=0$ outside $\calE_t$.\nllabel{line:policy-model}\;
  Execute $\pi_t$ and observe the trajectory $\{(s_{t,h},a_{t,h},\ell_t(s_{t,h},a_{t,h}))\}_{h=0}^{H-1}$.\nllabel{line:execute}\;
  If $t=1$ or $R(\pi_t,\pi_{r_{t-1}})>\sqrt2$ for $R$ in \cref{eq:epoch}, set $r_t=t$, and otherwise set $r_t=r_{t-1}$.\nllabel{line:epoch}\;
  Compute the received loss $G_t(s,a)=\sum_{k=1}^{t-1}w_{k,t}(s,a)\ell_k(s,a)/(c_0\iota)$ for all $(s,a)$.\nllabel{line:deliver}\;
  For all edges $(s,a,s')$, compute
  \begingroup\small
  \begin{equation}
    \hat g_t(s,a,s')=G_t(s,a)+\sum_{s''}\frac{\I_t(s,a,s'';\calE_t)V^{\hat P_{\calE_t},\pi_t}(s'';G_t)}{\hat q_t(s,a)+\gamma}-\frac{V^{\hat P_{\calE_t},\pi_t}(s';G_t)}{\hat q_t(s,a)}\I_t(s,a;\calE_t)\ind\set{(s,a,s')\in\calE_t}.\notag
  \end{equation}
  \endgroup
  \nllabel{line:gradient}\;
  For all $k<t$ and all $(s,a)$, compute
  $w_{k,t+1}(s,a)=w_{k,t}(s,a)\prn*{1-(\tilde q_{k,t}(s,a)+\gamma)/(c_0\iota)}$, where
  \begin{equation}
    \tilde q_{k,t}(s,a)=
    \begin{cases}
      \displaystyle\I_t(s,a;\calE_k)\prod_{h'=0}^{h(s)}
        \frac{\pi_k(a_{t,h'}\mid s_{t,h'})}{\pi_t(a_{t,h'}\mid s_{t,h'})}
        & \text{if }k\geq r_t,\\
      \mathrm{e}^{4H\epsilon}q^{\hat P_{\calE_k},\pi_k}(s,a) & \text{if }k<r_t.
    \end{cases}\notag
  \end{equation}
  \nllabel{line:service}\;
  Initialize $w_{t,t+1}(s,a)=\I_t(s,a;\calE_t)$ for all $(s,a)$.\nllabel{line:birth}\;
  Update $n(s,a)$ and $n(s,a,s')$ from the trajectory. When $n(s,a,s')$ first reaches $B$, activate the edge for the next episode and fix $\Pbar(s'\mid s,a)=B/n(s,a)$.\nllabel{line:activate}\;
}
\end{algorithm}
\endgroup

\subsection{Active edges and truncated kernels}\label{sec:calibration}
Let $n(s,a)$ and $n(s,a,s')$ denote the numbers of visits to $(s,a)$ and of transitions along the edge $(s,a,s')$ observed so far.
When $n(s,a,s')$ reaches $B$, the edge is declared \emph{active}, and its transition probability is estimated by $\Pbar(s'\mid s,a)=B/n(s,a)$ (\cref{line:activate}).
Once an edge becomes active, its estimate $\Pbar(s'\mid s,a)$ is never changed.
Let $\calE_t$ denote the set of active edges at the beginning of episode $t$, with $\calE_1=\emptyset$.
Since $B\gtrsim \iota/\epsilon^2$, the empirical Bernstein inequality implies that, with high probability (\cref{lem:calibration}),
\begin{align}\label{eq:calibration}
\abs*{\frac{P(s'\mid s,a)}{\Pbar(s'\mid s,a)}-1}\leq\epsilon\quad\text{for all }t\in[T]\text{ and }(s,a,s')\in\calE_t.
\end{align}
The algorithm uses each trajectory only up to its first transition along an inactive edge, since inactive edges have no estimate.
To formalize this, for a set $\calE$ of edges and a kernel $\widetilde P$, we define the truncated kernel
\begin{align}\label{eq:truncation}
\widetilde P_\calE(s'\mid s,a)=\widetilde P(s'\mid s,a)\ind\set{(s,a,s')\in\calE},
\end{align}
which assigns zero probability to every edge outside $\calE$.
The occupancy $q^{P_\calE,\pi}(s,a)$ is then the probability that $\pi$ reaches $(s,a)$ without traversing any edge outside $\calE$.
The corresponding observations in episode $t$ are defined, for $(s,a)\in\calS_h\times\calA$, by
\begin{align}
\I_t(s,a;\calE)&=\ind\set*{(s_{t,h},a_{t,h})=(s,a)\text{ and }(s_{t,h'},a_{t,h'},s_{t,h'+1})\in\calE\text{ for all }h'<h},\\
\I_t(s,a,s';\calE)&=\I_t(s,a;\calE)\ind\set{s_{t,h+1}=s'}\ind\set{(s,a,s')\in\calE}.
\end{align}
For every $\calF_t$-measurable edge set $\calE$, they then satisfy
\begin{align}\label{eq:indicator-mean}
\E_t[\I_t(s,a;\calE)]=q^{P_\calE,\pi_t}(s,a),\qquad
\E_t[\I_t(s,a,s';\calE)]=q^{P_\calE,\pi_t}(s,a,s').
\end{align}
Each edge is traversed at most $B$ times while inactive, and hence the discarded losses contribute to the regret only a lower-order term that depends on $T$ only logarithmically.

\subsection{FTRL over occupancy measures}\label{sec:ftrl}
FTRL runs over occupancy measures $q=(q(s,a,s'))$ indexed by the edges $(s,a,s')\in\calS_h\times\calA\times\calS_{h+1}$ for $0\leq h<H$, and we write $q(s,a)=\sum_{s'}q(s,a,s')$.
A function $f$ on $\calS\times\calA$ is identified with the function $f(s,a,s')=f(s,a)$ on the edges, and we write $\inpr{q,f}=\sum_{s,a}q(s,a)f(s,a)$ for such $f$ and $\inpr{q,g}=\sum_{(s,a,s')}q(s,a,s')g(s,a,s')$ for a function $g$ on the edges.
For an edge set $\calE$, the domain of FTRL is
\begin{align}\label{eq:domain}
\calQ(\calE)=\set*{q>0\given
\begin{aligned}
&\textstyle\sum_aq(s,a)=\ind\set{s=s_0}+\sum_{(s',a')\in\calS_{h(s)-1}\times\calA}q(s',a',s)&&\text{for all }s,\\
&\Pbar^-(s'\mid s,a)\,q(s,a)<q(s,a,s')<\Pbar^+(s'\mid s,a)\,q(s,a)&&\text{for all }(s,a,s')\in\calE
\end{aligned}},
\end{align}
where $\Pbar^\pm=(1\pm2\epsilon)\Pbar$.
The first condition is the flow constraint, which states that the occupancy of each state other than $s_0$ is the sum of $q(s',a',s)$ over the preceding layer.
The last condition is the interval constraint, which states that the transition probabilities $q(s,a,s')/q(s,a)$ on active edges lie between $\Pbar^-$ and $\Pbar^+$.
The width of the interval is twice the accuracy in \cref{eq:calibration}, and hence the true transition probabilities lie in the interior of the interval with margin $\epsilon$, which we use in the analysis.
We write $b^-(s,a,s';q)=q(s,a,s')-\Pbar^-(s'\mid s,a)\,q(s,a)$ and $b^+(s,a,s';q)=\Pbar^+(s'\mid s,a)\,q(s,a)-q(s,a,s')$ for the slacks of the interval constraint.
FTRL computes the update (\cref{line:ftrl}),
\begin{align}\label{eq:ftrl}
\hat q_t=\argmin_{q\in\calQ(\calE_t)}\set*{\inpr*{\sum_{k=1}^{t-1}\hat g_k,q}+\frac1\eta\Psi_{\calE_t}(q)},
\end{align}
where $\hat g_k$ is an estimate of the loss of episode $k$ on the edges, constructed in \cref{sec:gradient}, and the regularizer is
\begin{align}
\Psi_\calE(q)&=\sum_{s,a}q(s,a)\log q(s,a)
+\frac{H}{SA}\sum_{s,a}\log\frac{1}{q(s,a)}
+\frac{H}{S^2A}\sum_{(s,a,s')}\log\frac{1}{q(s,a,s')}\\
&\quad+\frac{H}{S^2A}\sum_{(s,a,s')\in\calE}\prn*{\log\frac{2}{b^-(s,a,s';q)}+\log\frac{2}{b^+(s,a,s';q)}}.\label{eq:regularizer}
\end{align}
This hybrid regularizer adds logarithmic barriers to the negative entropy, which keeps FTRL stable under the loss estimates constructed in \cref{sec:delivery,sec:gradient}.%
\footnote{If $\calQ(\calE_t)$ is empty, the algorithm plays an arbitrary fixed policy from episode $t$ on. This happens only if \cref{eq:calibration} fails, which occurs with probability at most $\delta$ by \cref{lem:calibration}.}

The algorithm computes the policy and the estimated kernel of episode $t$ by normalizing $\hat q_t$ (\cref{line:policy-model}),
\begin{align}\label{eq:policy-model}
\pi_t(a\mid s)=\frac{\hat q_t(s,a)}{\sum_b\hat q_t(s,b)},\qquad
\hat P_{\calE_t}(s'\mid s,a)=\frac{\hat q_t(s,a,s')}{\hat q_t(s,a)}\ind\set{(s,a,s')\in\calE_t}.
\end{align}
The estimated kernel $\hat P_{\calE_t}$ is a subkernel that keeps only the active edges, and by the interval constraint in \cref{eq:domain} it lies between $\Pbar^-$ and $\Pbar^+$ on them.

\subsection{Gradual delivery of observed losses}\label{sec:delivery}
For an episode $k$ and a state-action pair $(s,a)$, the probability that $\pi_k$ reaches $(s,a)$ before traversing an inactive edge is $q^{P_{\calE_k},\pi_k}(s,a)$.
The importance-weighted estimator $\I_k(s,a;\calE_k)\ell_k(s,a)/q^{P_{\calE_k},\pi_k}(s,a)$ is unbiased for $\ell_k(s,a)$, but it is not computable, since it depends on the unknown transition kernel.
To address this issue, \citet{kulkarni2026introducing} estimate the inverse visitation probability $1/q^{\pi_k}(s,a)$ under the true kernel from subsequent trajectories using geometric resampling with rejection, which builds on geometric resampling~\citep{neu2016importance}.
We refine this technique and use it to estimate $1/q^{P_{\calE_k},\pi_k}(s,a)$.
Instead of counting the trials of geometric resampling, we keep each observed loss with a weight, and the weight decreases in every later episode by an estimate of $q^{P_{\calE_k},\pi_k}(s,a)$.

We write $w_{k,t}(s,a)$ for the weight of $\ell_k(s,a)$ at the beginning of episode $t>k$, with $w_{k,k+1}(s,a)=\I_k(s,a;\calE_k)$ (\cref{line:birth}).
In episode $t$, FTRL receives the loss (\cref{line:deliver})
\begin{align}\label{eq:received}
G_t(s,a)=\frac{1}{c_0\iota}\sum_{k<t}w_{k,t}(s,a)\ell_k(s,a).
\end{align}
After observing the trajectory of episode $t$, the algorithm updates each weight as $w_{k,t+1}(s,a)=w_{k,t}(s,a)(1-(\tilde q_{k,t}(s,a)+\gamma)/(c_0\iota))$ (\cref{line:service}), where $\gamma$ in \cref{alg:main} is the implicit exploration rate.
Here $\tilde q_{k,t}(s,a)$ is an estimate of $q^{P_{\calE_k},\pi_k}(s,a)$ from the trajectory of episode $t$, given by
\begin{align}\label{eq:service}
\tilde q_{k,t}(s,a)=\I_t(s,a;\calE_k)\prod_{h'=0}^{h(s)}\frac{\pi_k(a_{t,h'}\mid s_{t,h'})}{\pi_t(a_{t,h'}\mid s_{t,h'})}.
\end{align}
The likelihood ratio removes the dependence on the transitions, as in \citet{kulkarni2026introducing}, and yields an unbiased estimate.
\begin{lemma}\label{lem:unbiased}
For every $k<t$ and $(s,a)\in\calS\times\calA$, the quantity in \cref{eq:service} satisfies $\E_t[\tilde q_{k,t}(s,a)]=q^{P_{\calE_k},\pi_k}(s,a)$.
\end{lemma}
\begin{proof}
Let $h=h(s)$.
The quantity in \cref{eq:service} is nonzero only if episode $t$ follows a prefix $(s_0,a_0,\ldots,s_h,a_h)$ with $(s_h,a_h)=(s,a)$ whose edges lie in $\calE_k$.
Hence
\begin{align}
\E_t[\tilde q_{k,t}(s,a)]
&=\sum_{s_0,a_0,\ldots,s_{h-1},a_{h-1}}\prod_{h'=0}^{h}\pi_t(a_{h'}\mid s_{h'})\prod_{h'=0}^{h-1}P_{\calE_k}(s_{h'+1}\mid s_{h'},a_{h'})\cdot\prod_{h'=0}^{h}\frac{\pi_k(a_{h'}\mid s_{h'})}{\pi_t(a_{h'}\mid s_{h'})}\\
&=\sum_{s_0,a_0,\ldots,s_{h-1},a_{h-1}}\prod_{h'=0}^{h}\pi_k(a_{h'}\mid s_{h'})\prod_{h'=0}^{h-1}P_{\calE_k}(s_{h'+1}\mid s_{h'},a_{h'})
=q^{P_{\calE_k},\pi_k}(s,a),
\end{align}
where the product $\prod_{h'=0}^{h}\pi_t(a_{h'}\mid s_{h'})\prod_{h'=0}^{h-1}P_{\calE_k}(s_{h'+1}\mid s_{h'},a_{h'})$ is the probability that episode $t$ follows the prefix $(s_0,a_0,\ldots,s_h,a_h)$ with all its edges in $\calE_k$.
\end{proof}

By \cref{lem:unbiased}, each update multiplies the weight by $1-(q^{P_{\calE_k},\pi_k}(s,a)+\gamma)/(c_0\iota)$ in conditional expectation, and this makes the delivered estimate unbiased when $\gamma = 0$.
Indeed, if there were infinitely many episodes, given that $(s,a)$ is visited in episode $k$, the expected total amount of $\ell_k(s,a)$ received by FTRL would be
\begin{align}
\frac{\ell_k(s,a)}{c_0\iota}\sum_{m=0}^\infty\prn*{1-\frac{q^{P_{\calE_k},\pi_k}(s,a)+\gamma}{c_0\iota}}^m=\frac{\ell_k(s,a)}{q^{P_{\calE_k},\pi_k}(s,a)+\gamma}.
\end{align}
Since $(s,a)$ is visited with probability $q^{P_{\calE_k},\pi_k}(s,a)$, the total contribution of $\ell_k(s,a)$ to FTRL is at most $\ell_k(s,a)$ in expectation (\cref{lem:delayed-estimate}), just like the importance-weighted estimator.

However, if $\pi_t$ moves far from $\pi_k$, the likelihood ratio in \cref{eq:service} becomes large, and the factor  $1-(\tilde q_{k,t}(s,a)+\gamma)/(c_0\iota)$ can be negative.
To prevent this, the algorithm maintains a reference episode $r_t\leq t$, with $r_1=1$, and measures how far $\pi_t$ is from $\pi_{r_{t-1}}$ by
\begin{align}\label{eq:epoch}
R(\pi,\pi')=\prod_{h=0}^{H-1}\max_{(s,a)\in\calS_h\times\calA}\max\set*{\frac{\pi(a\mid s)}{\pi'(a\mid s)},\frac{\pi'(a\mid s)}{\pi(a\mid s)}},
\end{align}
and sets $r_t=t$ if $R(\pi_t,\pi_{r_{t-1}})>\sqrt2$ and $r_t=r_{t-1}$ otherwise (\cref{line:epoch}).
Then the likelihood ratio in \cref{eq:service} is at most $R(\pi_k,\pi_{r_t})R(\pi_t,\pi_{r_t})\leq2$ for $k\geq r_t$.
For $k<r_t$, the algorithm instead uses
\begin{align}\label{eq:upper-probability}
\tilde q_{k,t}(s,a)=\mathrm{e}^{4H\epsilon}q^{\hat P_{\calE_k},\pi_k}(s,a),
\end{align}
which inflates the visitation probability under the estimated kernel of episode $k$ and hence upper bounds $q^{P_{\calE_k},\pi_k}(s,a)$ under \cref{eq:calibration}.
This replacement contributes at most the order of the leading term of our bound.

\subsection{Loss vector with temporal-difference errors}\label{sec:gradient}
If FTRL were run with the loss $G_t$, it would bound $\sum_t\inpr{\hat q_t-q,G_t}$ for a comparator $q$.
In the regret, however, the learner incurs the loss $\inpr{q^{P_{\calE_t},\pi_t},G_t}$ by running $\pi_t$ under the true kernel, and hence we need to bound $\sum_t\inpr{q^{P_{\calE_t},\pi_t}-q,G_t}$.
The two sums differ by $\sum_t\inpr{q^{P_{\calE_t},\pi_t}-\hat q_t,G_t}$, and each term of this sum is at most the following gap since $\hat q_t\geq q^{\hat P_{\calE_t},\pi_t}$ by the truncation of the estimated kernel,
\begin{align}\label{eq:simulation}
\inpr*{q^{P_{\calE_t},\pi_t}-q^{\hat P_{\calE_t},\pi_t},G_t}
&=\sum_{s,a}q^{P_{\calE_t},\pi_t}(s,a)\sum_{s'}\prn*{P_{\calE_t}(s'\mid s,a)-\hat P_{\calE_t}(s'\mid s,a)}V^{\hat P_{\calE_t},\pi_t}(s';G_t)\\
&=\E_t\brk*{\sum_{s,a}\mathrm{TD}_t(s,a)},
\end{align}
where
\begin{align}\label{eq:td}
\mathrm{TD}_t(s,a)=\sum_{s'}\prn*{\I_t(s,a,s';\calE_t)-\hat P_{\calE_t}(s'\mid s,a)\I_t(s,a;\calE_t)}V^{\hat P_{\calE_t},\pi_t}(s';G_t)
\end{align}
is the temporal-difference (TD) error.
Bounding this gap by the confidence width of $\hat P_{\calE_t}$ leads to the extra factor $\sqrt S$ in the regret bound of \citet{jin2020learning}.
Instead, since $\mathrm{TD}_t$ is computable from the trajectory of episode $t$, we add it to the loss of FTRL and charge the error of the estimated transitions to the learner (\cref{line:gradient}),
\begin{align}\label{eq:gradient}
\begin{aligned}
\hat g_t(s,a,s')&=G_t(s,a)+\sum_{s''}\frac{\I_t(s,a,s'';\calE_t)V^{\hat P_{\calE_t},\pi_t}(s'';G_t)}{\hat q_t(s,a)+\gamma}\\
&\quad-\frac{V^{\hat P_{\calE_t},\pi_t}(s';G_t)}{\hat q_t(s,a)}\I_t(s,a;\calE_t)\ind\set{(s,a,s')\in\calE_t}.
\end{aligned}
\end{align}
Since $\hat q_t(s,a,s')=\hat q_t(s,a)\hat P_{\calE_t}(s'\mid s,a)$ on active edges, $\sum_{s'}\hat q_t(s,a,s')\hat g_t(s,a,s')=\hat q_t(s,a)G_t(s,a)+\mathrm{TD}_t(s,a)-\beta_t(s,a)$ for every $(s,a)$, where
\begin{align}\label{eq:ix-term}
\beta_t(s,a)=\frac{\gamma}{\hat q_t(s,a)+\gamma}\sum_{s''}\I_t(s,a,s'';\calE_t)V^{\hat P_{\calE_t},\pi_t}(s'';G_t)\geq0
\end{align}
is the bias caused by the implicit exploration rate $\gamma$, which is needed only for the high-probability bound.
This identity gives the following lemma, which we prove in \cref{app:ftrl}.
\begin{lemma}\label{lem:pairing}
For every episode $t$, $\E_t[\inpr{\hat q_t,\hat g_t}]=\inpr{\hat q_t+q^{P_{\calE_t},\pi_t}-q^{\hat P_{\calE_t},\pi_t},G_t}-\sum_{s,a}\E_t[\beta_t(s,a)]$.
\end{lemma}
By the lemma and $\hat q_t\geq q^{\hat P_{\calE_t},\pi_t}$, we have $\E_t[\inpr{\hat q_t,\hat g_t}]\geq\inpr{q^{P_{\calE_t},\pi_t},G_t}-\sum_{s,a}\E_t[\beta_t(s,a)]$.
On the other hand, $\E_t[\inpr{q,\hat g_t}]\leq\inpr{q,G_t}$ for a comparator $q$ that follows the true transitions.
Thus, by including the transition estimation error in the loss vector $\hat g_t$ of FTRL, we obtain $\sum_t\inpr{q^{P_{\calE_t},\pi_t}-q,G_t}\leq\sum_t\E_t[\inpr{\hat q_t-q,\hat g_t}]+\sum_t\sum_{s,a}\E_t[\beta_t(s,a)]$, which reduces the regret to the analysis of FTRL.

\section{Regret guarantee}\label{sec:main-result}
In this section, we present the regret guarantee of \cref{alg:main} and sketch its proof, whose details are given in \cref{app:events,app:regularizer,app:auxiliary,app:terms}.
\begin{theorem}\label{thm:main}
\cref{alg:main} achieves, with probability at least $1-11\delta$,
\begin{align}\label{eq:main-rate}
\Reg_T\lesssim H\sqrt{SAT}\log^2\prn*{\frac{SAT}{\delta}}+H^2S^3A^2\max\set{SA,H^2}\log\prn*{\frac{SA}{\delta}}.
\end{align}
\end{theorem}
\cref{thm:main} gives the first regret bound that matches the lower bound $\Omega(H\sqrt{SAT})$~\citep{jin2018q,domingues2021episodic} up to logarithmic factors.
Our bound improves the previous bounds $\widetilde O(HS\sqrt{AT})$ of \citet{jin2020learning} by a factor of $\sqrt S$ and $\widetilde O(H^{5/2}\sqrt{SAT})$ of \citet{kulkarni2026introducing} by a factor of $H^{3/2}$.
It is even better than the best known bound $\widetilde O(H^{3/2}\sqrt{SAT})$ under full-information feedback~\citep{schlisselberg2026improved}, which is an easier setting.

\subsection{Proof sketch}\label{sec:sketch}
We give an overview of the proof of \cref{thm:main}.
We first define the stopping time $\tau$ as the first episode $t$ in which the relative error of some estimated transition probability exceeds $\epsilon$ or the sum of the weights $w_{k,t}(s,a)$ over $k<t$ exceeds $c_0^2\iota^2$ for some $(s,a)$,
\begin{align}
\tau=\min\set*{t\in[T]:\max_{(s,a,s')\in\calE_t}\abs*{\frac{P(s'\mid s,a)}{\Pbar(s'\mid s,a)}-1}>\epsilon\text{ or }\max_{s,a}\sum_{k<t}w_{k,t}(s,a)>c_0^2\iota^2},
\end{align}
with $\tau=T+1$ if there is no such episode.
As $\tau=T+1$ holds with probability at least $1-3\delta$ (\cref{lem:stopping}), we assume in the rest of the proof that both conditions hold in all episodes.
On this event, we define the total amount of $\ell_t(s,a)$ that FTRL receives over the later episodes as
\begin{align}\label{eq:source-estimate}
\hat\ell_t(s,a)=\ell_t(s,a)\sum_{u=t+1}^{\tau-1}\frac{w_{t,u}(s,a)}{c_0\iota},
\end{align}
and we also write $\bar q_t=\hat q_t+q^{P_{\calE_t},\pi_t}-q^{\hat P_{\calE_t},\pi_t}$.
The regret then decomposes into the following five terms (\cref{lem:decomposition}), which we bound in turn,
\begin{align}
\Reg_T&=\underbrace{\sum_t\inpr*{\bar q_t-q^{\pio},G_t}}_{\text{(I)}}+\underbrace{\sum_t\prn*{\inpr*{q^{P_{\calE_t},\pi_t},\hat\ell_t}-\inpr*{\bar q_t,G_t}}}_{\text{(II)}}+\underbrace{\sum_t\inpr*{q^{P_{\calE_t},\pi_t},\ell_t-\hat\ell_t}}_{\text{(III)}}\\
&\quad+\underbrace{\sum_t\inpr*{q^{\pio},\hat\ell_t-\ell_t}}_{\text{(IV)}}+\underbrace{\sum_t\inpr*{q^{\pi_t}-q^{P_{\calE_t},\pi_t},\ell_t}}_{\text{(V)}}.\label{eq:sketch-decomposition}
\end{align}
Term (I) is the regret on the received losses $G_t$, where the learner is evaluated by $\bar q_t$.
By \cref{lem:pairing} and the inequality $\E_t[\inpr{q^{\pio},\hat g_t}]\leq\inpr{q^{\pio},G_t}$ in \cref{sec:gradient}, we have $\text{(I)}\leq\sum_t\inpr{\hat q_t-q^{\pio},\hat g_t}+\widetilde O(H\sqrt{SAT})$.
The first sum is the regret of FTRL with the loss vectors $\hat g_t$, and the standard analysis of FTRL bounds it by $\widetilde O(H/\eta+\eta\sum_t\nrm{\hat g_t}_{t,*}^2)$ after moving $q^{\pio}$ slightly into the interior of the domain, where $\nrm{\cdot}_{t,*}$ is the dual of the local norm of $\Psi_{\calE_t}$ at $\hat q_t$.
Since $\hat g_t$ contains the TD errors, the second term can be large in general.
The logarithmic barrier on $q(s,a)$ prevents this and gives $\E_t[\nrm{\hat g_t}_{t,*}^2]=\widetilde O(HSA)$ (\cref{lem:gradient-moments}), and hence $\text{(I)}=\widetilde O(H/\eta+\eta HSAT+H\sqrt{SAT})$ (\cref{lem:ftrl}).

Term (II) is the cost of the gradual delivery.
We can write $\text{(II)}=\sum_{k<t}\sum_{s,a}w_{k,t}(s,a)\ell_k(s,a)(q^{P_{\calE_k},\pi_k}(s,a)-\bar q_t(s,a))/(c_0\iota)$.
Here a loss observed in episode $k$ is measured by the occupancy measure of episode $k$, while FTRL receives it in a later episode $t$ and measures it by $\bar q_t$.
The logarithmic barrier on $q(s,a)$ keeps this difference small by keeping the relative change of $\bar q_t(s,a)$ between episodes small for every $(s,a)$ (\cref{lem:motion}), and hence $\text{(II)}=\widetilde O(\eta HSAT)$ (\cref{lem:delay}).

Term (III) compares the losses $\ell_t$ with the total amounts $\hat\ell_t$ received by FTRL, weighted by the occupancy measure of the learner.
If the weights always decreased by the unbiased estimate in \cref{eq:service}, we would have $\E_t[\hat\ell_t]=\ell_t$ apart from $\gamma$, and this term would be small.
The bias comes only from losses observed before an update of the reference episode, whose weights then decrease by the upper bound in \cref{eq:upper-probability} with relative error $O(H\epsilon)$.
The policy changes little in each episode (\cref{lem:policy-motion}), so such updates are rare, and hence $\text{(III)}=\widetilde O(\eta HSAT)$ by the choice of $\epsilon$ (\cref{lem:source}).

Term (IV) compares the same quantities, weighted by the occupancy measure of the comparator.
By $\E_t[\hat\ell_t]\leq\ell_t$ (\cref{lem:delayed-estimate}), it is nonpositive in expectation.
With the implicit exploration rate $\gamma$, it is also at most $\widetilde O(H/\gamma)=\widetilde O(H\sqrt{SAT})$ with high probability (\cref{lem:comparator}).

Term (V) is the loss caused by the inactive edges.
Each edge is traversed at most $B$ times while inactive, and hence $\text{(V)}\lesssim HS^2AB$ (\cref{lem:inactive}), which depends on $T$ only through the logarithmic factor in $B$.

Substituting the bounds on (I)--(V) into the decomposition \cref{eq:sketch-decomposition}, we obtain $\Reg_T=\widetilde O(H/\eta+\eta HSAT+H\sqrt{SAT})$, and the choice of $\eta$ in \cref{alg:main} balances the first two terms, which gives \cref{thm:main} (\cref{app:main-proof}).

\section{Conclusion}\label{sec:conclusion}
For adversarial MDPs with unknown transitions under bandit feedback, we gave an algorithm with regret $\widetilde O(H\sqrt{SAT})$, which matches the lower bound up to logarithmic factors.
The key ideas are to include the transition estimation error in the loss of FTRL and to feed each observed loss to FTRL gradually over later episodes.
As a future direction, it would be interesting to see whether policy optimization can achieve the same regret rate of $\widetilde O(H\sqrt{SAT})$.
Other directions include extending our techniques to linear MDPs and obtaining best-of-both-worlds guarantees.

\section*{AI use statement}
We used GPT-6-Astra, GPT-6.2-Astra, GPT-6.2-Sol, and Claude Fable 5.2 and Opus 5.5 to help develop and check proofs, search the literature, and edit the manuscript.
We reviewed all AI-assisted work and substantially revised AI-generated text for clarity.
We take responsibility for the final manuscript.

\bibliographystyle{plainnat}
\bibliography{ref}

\clearpage
\appendix
\crefalias{section}{appendix}
\crefalias{subsection}{appendix}

\tableofcontents
\clearpage
\section{Notation}\label{app:notation}
In this section, we first define the additional notation used in the appendix and then summarize the main notation of the paper in \cref{tab:notation} for convenience.

\begin{table}[H]
\caption{Summary of notation.}\label{tab:notation}
\centering
\small
\setlength{\parskip}{0pt}
\renewcommand{\arraystretch}{1.0}
\begin{tabular}{@{}ll@{}}
\toprule
\textbf{Symbol} & \textbf{Meaning} \\
\midrule
\multicolumn{2}{@{}l}{\textbf{Tabular MDPs}} \\
$\calM=(\calS,\calA,P,H,s_0)$ & Episodic tabular MDP \\
$\calS$, $S$ & State space and its size $S=\abs{\calS}$ \\
$\calA$, $A$ & Action space and its size $A=\abs{\calA}$ \\
$P$ & Transition kernel, unknown to the learner \\
$H$ & Horizon \\
$T$ & Number of episodes \\
$\calS_h$, $h(s)$ & Layer $h$ and layer index of state $s$ \\
$(s,a,s')$ & Edge with $s\in\calS_h$, $a\in\calA$, and $s'\in\calS_{h+1}$ \\
$s_{t,h}$, $a_{t,h}$ & State and action at step $h$ in episode $t$ \\
$\ell_t(s,a)$ & Loss of $(s,a)$ in episode $t$ \\
$\pi_t$ & Policy in episode $t$ \\
$V^{\widetilde P,\pi}(s;f)$, $Q^{\widetilde P,\pi}(s,a;f)$ & Value function and $Q$-function of $f$ under $\widetilde P$ and policy $\pi$ \\
$q^{\widetilde P,\pi}(s,a)$, $q^{\widetilde P,\pi}(s,a,s')$ & Occupancy measure under $\widetilde P$ and $\pi$ \\
$\Reg_T$ & Regret over $T$ episodes \\
$\pio$ & Optimal fixed policy in hindsight \\
\midrule
\multicolumn{2}{@{}l}{\textbf{Notation for \cref{alg:main} (\cref{sec:algorithm})}} \\
$\iota$, $c_0$, $\epsilon$, $B$, $\eta$, $\gamma$ & Parameters in \cref{alg:main} \\
$n(s,a)$, $n(s,a,s')$ & Numbers of visits to $(s,a)$ and transitions along $(s,a,s')$ observed so far \\
$\calE_t$ & Set of active edges before episode $t$ \\
$\Pbar(s'\mid s,a)$ & Estimate of the transition probability $P(s'\mid s,a)$ \\
$\widetilde P_\calE$ & Transition kernel $\widetilde P$ truncated to $\calE$ \cref{eq:truncation} \\
$\I_t(s,a;\calE)$, $\I_t(s,a,s';\calE)$ & Visit and traversal indicators in episode $t$ before leaving $\calE$ \\
$\calQ(\calE)$, $\Psi_\calE$ & FTRL domain \cref{eq:domain} and regularizer \cref{eq:regularizer} \\
$\hat q_t$, $\hat P_{\calE_t}$ & FTRL iterate \cref{eq:ftrl} and its estimated transition kernel \cref{eq:policy-model} \\
$w_{k,t}(s,a)$ & Weight of $\ell_k(s,a)$ at the beginning of episode $t$ \\
$G_t$, $\hat g_t$ & Received loss \cref{eq:received} and FTRL loss vector \cref{eq:gradient} \\
$\tilde q_{k,t}(s,a)$ & Estimate of $q^{P_{\calE_k},\pi_k}(s,a)$ in episode $t$ \cref{eq:service,eq:upper-probability} \\
$R(\pi,\pi')$, $r_t$ & Product of maximum action probability ratios across layers \cref{eq:epoch} and reference episode \\
$\mathrm{TD}_t$, $\beta_t(s,a)$ & TD error \cref{eq:td} and nonnegative bias from $\gamma$ \cref{eq:ix-term} \\
\midrule
\multicolumn{2}{@{}l}{\textbf{Notation for the analysis (\cref{sec:main-result} and the appendix)}} \\
$\tau$ & Stopping time \cref{eq:stopping} \\
$\hat\ell_t$ & Total amount of $\ell_t$ received by FTRL \cref{eq:source-estimate} \\
$\bar q_t$ & $\hat q_t+q^{P_{\calE_t},\pi_t}-q^{\hat P_{\calE_t},\pi_t}$ \\
$\hat q_t^+$, $\pi_t^+$, $\hat P^+_{\calE_t}$ & Next FTRL iterate over $\calQ(\calE_t)$, with its policy and estimated transition kernel \\
$\bar q_t^+$ & $\hat q_t^+ + q^{P_{\calE_t},\pi_t^+}-q^{\hat P^+_{\calE_t},\pi_t^+}$ \\
$\nrm{\cdot}_t$, $\nrm{\cdot}_{t,*}$ & Local norm induced by $\nabla^2\Psi_{\calE_t}(\hat q_t)$ and its dual \\
\bottomrule
\end{tabular}
\end{table}
\newpage
The analysis of FTRL uses the following derivatives and local norms.
For a function $f$, a point $x$ in its domain, and vectors $u_1,\ldots,u_k$, $\nabla^kf(x)[u_1,\ldots,u_k]$ is the $k$-th derivative of $f$ at $x$ in the directions $u_1,\ldots,u_k$.
For a convex function $f$, a point $x$ in its domain, and a vector $u$, the local norm is $\nrm{u}_{x,f}=(\nabla^2f(x)[u,u])^{1/2}$.
For vectors $u$ and $g$ on the edges, we write
$
\nrm{u}_t=\nrm{u}_{\hat q_t,\Psi_{\calE_t}},\qquad
\nrm{g}_{t,*}=\sup\set*{\inpr{g,u}\given \nrm{u}_t\leq1},
$
and by definition, $\abs{\inpr{g,u}}\leq\nrm{g}_{t,*}\nrm{u}_t$.

We also write $\bar q_t=\hat q_t+q^{P_{\calE_t},\pi_t}-q^{\hat P_{\calE_t},\pi_t}$.
Let $\hat q_t^+$ be the minimizer of $\inpr{\sum_{k\leq t}\hat g_k,q}+\Psi_{\calE_t}(q)/\eta$ over $\calQ(\calE_t)$, which is the next FTRL iterate before new edges are activated.
Let $\pi_t^+$ and $\hat P^+_{\calE_t}$ be the policy and the estimated kernel obtained from $\hat q_t^+$ by \cref{eq:policy-model}, and let $\bar q_t^+=\hat q_t^++q^{P_{\calE_t},\pi_t^+}-q^{\hat P^+_{\calE_t},\pi_t^+}$.
If no edge is activated in episode $t$, then $\hat q_{t+1}=\hat q_t^+$ and $\bar q_{t+1}=\bar q_t^+$.

\section{Concentration Inequalities and High-Probability Events}\label{app:events}
This section first gives the concentration inequalities used in the analysis (\cref{app:concentration}) and then shows that the conditions in \cref{eq:good-event} hold for every episode with probability at least $1-3\delta$ (\cref{app:good-event}).

\subsection{Concentration inequalities}\label{app:concentration}
\begin{lemma}[{Empirical Bernstein inequality~\citep[Theorem~4]{maurer2009empirical}}]\label{lem:empirical-bernstein}
Let $m\geq2$ and let $Z_1,\ldots,Z_m$ be i.i.d.\ random variables with values in $[0,1]$ and mean $\mu$.
Then, with probability at least $1-2\delta$,
\begin{align}
\abs*{\frac1m\sum_{i=1}^mZ_i-\mu}\leq\sqrt{\frac{2V_m\log(2/\delta)}{m}}+\frac{7\log(2/\delta)}{3(m-1)},
\end{align}
where $V_m=\frac{1}{m(m-1)}\sum_{1\leq i<j\leq m}(Z_i-Z_j)^2$ is the sample variance.
\end{lemma}

\begin{lemma}\label{lem:bounded-sum}
Let $X_1,\ldots,X_T$ be random variables with values in $[0,R]$ such that $X_t$ is $\calF_{t+1}$-measurable.
Each of the following inequalities holds with probability at least $1-\delta$,
\begin{align}\label{eq:bounded-sum}
\sum_{t=1}^TX_t\leq\frac32\sum_{t=1}^T\E_t[X_t]+2R\log\prn*{\frac1\delta},\qquad
\sum_{t=1}^TX_t\geq\frac12\sum_{t=1}^T\E_t[X_t]-2R\log\prn*{\frac1\delta}.
\end{align}
\end{lemma}
\begin{proof}
Let $Y_1,\ldots,Y_T$ be random variables such that $Y_t$ is $\calF_{t+1}$-measurable with $\E_t[Y_t]=0$ and $\abs{Y_t}\leq R$.
For $\alpha\in(0,1/R)$, \citet[Lemma~E.2]{lancewicki2025near} showed that with probability at least $1-\delta$
\begin{align}\label{eq:freedman}
\sum_{t=1}^TY_t\leq\alpha\sum_{t=1}^T\E_t\brk*{Y_t^2}+\frac1\alpha\log\prn*{\frac1\delta}.
\end{align}
We take $Y_t=\pm\prn{X_t-\E_t[X_t]}$ and $\alpha=1/(2R)$.
Then $\abs{Y_t}\leq R$ and $\E_t[Y_t^2]\leq\E_t[X_t^2]\leq R\,\E_t[X_t]$. Hence, with probability at least $1-\delta$
\begin{align}
\pm\sum_{t=1}^T\prn*{X_t-\E_t[X_t]}\leq\frac12\sum_{t=1}^T\E_t[X_t]+2R\log\prn*{\frac1\delta}.
\end{align}
\end{proof}

\begin{lemma}\label{lem:martingale}
Let $X_1,\ldots,X_T$ be random variables such that $X_t$ is $\calF_{t+1}$-measurable, $\E_t[X_t]=0$, and $\abs{X_t}\leq R$, and suppose that $\sum_{t=1}^T\E_t[X_t^2]\leq V$ for a constant $V$.
Then, with probability at least $1-\delta$,
\begin{align}\label{eq:martingale}
\sum_{t=1}^TX_t\leq\sqrt{2V\log(1/\delta)}+\frac23R\log(1/\delta).
\end{align}
\end{lemma}
\begin{proof}
By Bernstein's inequality for martingales \citep[Lemma~A.8]{cesa2006prediction} and $\sum_{t=1}^T\E_t[X_t^2]\leq V$, for every $x>0$,
\begin{align}\label{eq:bernstein}
\P\prn*{\sum_{t=1}^TX_t>x}\leq\exp\prn*{-\frac{x^2}{2(V+Rx/3)}}.
\end{align}
Let $x=\sqrt{2V\log(1/\delta)}+\frac23R\log(1/\delta)$.
Then $x\geq\sqrt{2V\log(1/\delta)}$ gives
\begin{align}
x^2=\prn*{\sqrt{2V\log(1/\delta)}+\frac23R\log(1/\delta)}x\geq2V\log(1/\delta)+\frac23Rx\log(1/\delta)=2\prn*{V+\frac{Rx}{3}}\log\frac1\delta,
\end{align}
and hence \cref{eq:bernstein} gives $\P(\sum_{t=1}^TX_t>x)\leq\delta$.
\end{proof}

\begin{lemma}\label{lem:supermartingale}
Let $X_1,\ldots,X_T$ be random variables such that $X_t$ is $\calF_{t+1}$-measurable and $\E_t[\exp(X_t)]\leq1$.
Then, with probability at least $1-\delta$, $\sum_{t=1}^TX_t\leq\log(1/\delta)$.
\end{lemma}
\begin{proof}
Since $\E_t[\exp(X_t)]\leq1$ for every $t$,
\begin{align}
\E\brk*{\exp\prn*{\sum_{t=1}^TX_t}}=\E\brk*{\exp\prn*{\sum_{t=1}^{T-1}X_t}\E_T\brk*{\exp(X_T)}}\leq\E\brk*{\exp\prn*{\sum_{t=1}^{T-1}X_t}}\leq\cdots\leq1,
\end{align}
and hence Markov's inequality gives
\begin{align}
\P\prn*{\sum_{t=1}^TX_t>\log\frac1\delta}=\P\prn*{\exp\prn*{\sum_{t=1}^TX_t}>\frac1\delta}\leq\delta\,\E\brk*{\exp\prn*{\sum_{t=1}^TX_t}}\leq\delta.
\end{align}
\end{proof}

\subsection{High-probability events}\label{app:good-event}
For $t'\in[T]$, consider the conditions
\begin{align}\label{eq:good-event}
\abs*{\frac{P(s'\mid s,a)}{\Pbar(s'\mid s,a)}-1}\leq\epsilon\quad\text{for all }(s,a,s')\in\calE_{t'}\qquad\text{and}\qquad\max_{s,a}\sum_{k<t'}w_{k,t'}(s,a)\leq c_0^2\iota^2,
\end{align}
and let $\tau$ be the first episode in which \cref{eq:good-event} fails,
\begin{align}\label{eq:stopping}
\tau=\min\set*{t\in[T]:\text{\cref{eq:good-event} is violated for }t'=t}\text{ with }\min\emptyset=T+1.
\end{align}
The event $\set{t<\tau}$ is $\calF_t$-measurable, since it is determined by the conditions for $t'\leq t$, which depend only on the history up to the end of episode $t-1$.
Note that $\tau\geq2$, as the conditions hold trivially for $t'=1$ by $\calE_1=\emptyset$.
The first condition in \cref{eq:good-event} follows from \cref{lem:calibration} and the second from \cref{lem:total-weight}.

\begin{lemma}\label{lem:calibration}
With probability at least $1-\delta$,
\begin{align}
\abs*{\frac{P(s'\mid s,a)}{\Pbar(s'\mid s,a)}-1}\leq\epsilon\quad\text{for all }t\in[T]\text{ and }(s,a,s')\in\calE_t.
\end{align}
\end{lemma}
\begin{proof}
Fix an edge $(s,a,s')$, and let $Z_i$ be the indicator that the $i$-th visit to $(s,a)$ moves to $s'$.
Then, $Z_1,Z_2,\ldots$ are i.i.d.\ Bernoulli variables with mean $P(s'\mid s,a)$.
For $m\geq2$, the sample variance of $Z_1,\ldots,Z_m$ in \cref{lem:empirical-bernstein} satisfies
\begin{align}\label{eq:sample-variance}
V_m=\frac{1}{m(m-1)}\sum_{1\leq i<j\leq m}(Z_i-Z_j)^2\leq\frac{1}{m(m-1)}\sum_{1\leq i<j\leq m}(Z_i+Z_j)=\frac1m\sum_{i=1}^mZ_i.
\end{align}
By \cref{lem:empirical-bernstein} and a union bound, with probability at least $1-\delta$, for all edges $(s,a,s')$ at all times $t$ with $n(s,a)\geq2$,
\begin{align}
\abs*{\frac{n(s,a,s')}{n(s,a)}-P(s'\mid s,a)}
&\leq\sqrt{\frac{2V_{n(s,a)}}{n(s,a)}\log\prn*{\frac{4S^2AT}{\delta}}}+\frac{7}{3(n(s,a)-1)}\log\prn*{\frac{4S^2AT}{\delta}}\\
&\leq\frac{1}{n(s,a)}\sqrt{2n(s,a,s')\log\prn*{\frac{4S^2AT}{\delta}}}+\frac{7}{3(n(s,a)-1)}\log\prn*{\frac{4S^2AT}{\delta}},
\end{align}
where we used \cref{eq:sample-variance} and $\sum_{i=1}^{n(s,a)}Z_i=n(s,a,s')$.
When the edge is activated, $n(s,a,s')=B$, so that $n(s,a)\geq B\geq2$, and $\Pbar(s'\mid s,a)=B/n(s,a)$.
Hence
\begin{align}
\abs*{\frac{P(s'\mid s,a)}{\Pbar(s'\mid s,a)}-1}
&=\frac{\abs{\Pbar(s'\mid s,a)-P(s'\mid s,a)}}{\Pbar(s'\mid s,a)}
=\frac{n(s,a)}{B}\abs*{\frac{n(s,a,s')}{n(s,a)}-P(s'\mid s,a)}\\
&\leq\sqrt{\frac2B\log\prn*{\frac{4S^2AT}{\delta}}}+\frac{14}{3B}\log\prn*{\frac{4S^2AT}{\delta}}\leq\epsilon,
\end{align}
where the last inequality follows from $B\geq25\iota/\epsilon^2$, $\log(4S^2AT/\delta)\leq2\iota$, and $\epsilon\leq1$.
\end{proof}

\begin{lemma}\label{lem:upper-comparison}
Let $k\in[T]$, and suppose that \cref{eq:calibration} holds for $\calE_k$.
Then, for every $(s,a)$,
\begin{align}\label{eq:upper-comparison}
&q^{P_{\calE_k},\pi_k}(s,a)\leq \mathrm{e}^{4H\epsilon}q^{\hat P_{\calE_k},\pi_k}(s,a)\leq \mathrm{e}^{8H\epsilon}q^{P_{\calE_k},\pi_k}(s,a)\leq2q^{P_{\calE_k},\pi_k}(s,a),\\
&\prn*{\mathrm{e}^{4H\epsilon}q^{\hat P_{\calE_k},\pi_k}(s,a)}^2\leq2q^{P_{\calE_k},\pi_k}(s,a).
\end{align}
\end{lemma}
\begin{proof}
By the definition of $\epsilon$ in \cref{alg:main}, $H\epsilon\leq1/100$, so that $\epsilon\leq1/8$ and $12H\epsilon\leq\log2$, that is, $\mathrm{e}^{12H\epsilon}\leq2$.
Let $(s,a,s')\in\calE_k$.
By \cref{eq:calibration} and \cref{eq:domain},
\begin{align}
(1-\epsilon)\Pbar(s'\mid s,a)&\leq P(s'\mid s,a)\leq(1+\epsilon)\Pbar(s'\mid s,a),\\
(1-2\epsilon)\Pbar(s'\mid s,a)&\leq\hat P_{\calE_k}(s'\mid s,a)\leq(1+2\epsilon)\Pbar(s'\mid s,a).
\end{align}
Since $\epsilon\leq1/8$, and $\log(1+x)\leq x$ and $\log(1-x)\geq-x/(1-x)$ for $0\leq x<1$,
\begin{align}
\log\prn*{\frac{1+\epsilon}{1-2\epsilon}}\leq\epsilon+\frac{2\epsilon}{1-2\epsilon}\leq4\epsilon,\qquad
\log\prn*{\frac{1-\epsilon}{1+2\epsilon}}\geq-\frac{\epsilon}{1-\epsilon}-2\epsilon\geq-4\epsilon.
\end{align}
Hence
\begin{align}\label{eq:kernel-comparison}
\mathrm{e}^{-4\epsilon}\leq\frac{1-\epsilon}{1+2\epsilon}\leq\frac{P(s'\mid s,a)}{\hat P_{\calE_k}(s'\mid s,a)}\leq\frac{1+\epsilon}{1-2\epsilon}\leq \mathrm{e}^{4\epsilon}.
\end{align}
Since $P_{\calE_k}$ and $\hat P_{\calE_k}$ are zero outside $\calE_k$ by \cref{eq:truncation,eq:policy-model}, it follows that $P_{\calE_k}(s'\mid s,a)\leq \mathrm{e}^{4\epsilon}\hat P_{\calE_k}(s'\mid s,a)$ and $\hat P_{\calE_k}(s'\mid s,a)\leq \mathrm{e}^{4\epsilon}P_{\calE_k}(s'\mid s,a)$ for all $(s,a,s')$.
With $h=h(s)$ and $(s_h,a_h)=(s,a)$ we have
\begin{align}
q^{P_{\calE_k},\pi_k}(s,a)
&=\sum_{s_0,a_0,\ldots,s_{h-1},a_{h-1}}\prod_{h'=0}^{h}\pi_k(a_{h'}\mid s_{h'})\prod_{h'=0}^{h-1}P_{\calE_k}(s_{h'+1}\mid s_{h'},a_{h'})\label{eq:trajectory-sum}\\
&\leq \mathrm{e}^{4H\epsilon}\sum_{s_0,a_0,\ldots,s_{h-1},a_{h-1}}\prod_{h'=0}^{h}\pi_k(a_{h'}\mid s_{h'})\prod_{h'=0}^{h-1}\hat P_{\calE_k}(s_{h'+1}\mid s_{h'},a_{h'})
=\mathrm{e}^{4H\epsilon}q^{\hat P_{\calE_k},\pi_k}(s,a),
\end{align}
and exchanging the roles of $P_{\calE_k}$ and $\hat P_{\calE_k}$ also gives $q^{\hat P_{\calE_k},\pi_k}(s,a)\leq \mathrm{e}^{4H\epsilon}q^{P_{\calE_k},\pi_k}(s,a)$.
Thus, we obtain
\begin{align}
q^{P_{\calE_k},\pi_k}(s,a)&\leq \mathrm{e}^{4H\epsilon}q^{\hat P_{\calE_k},\pi_k}(s,a)\leq \mathrm{e}^{8H\epsilon}q^{P_{\calE_k},\pi_k}(s,a)
\end{align}
and
\begin{align}
\prn*{\mathrm{e}^{4H\epsilon}q^{\hat P_{\calE_k},\pi_k}(s,a)}^2&=\mathrm{e}^{8H\epsilon}q^{\hat P_{\calE_k},\pi_k}(s,a)^2\leq \mathrm{e}^{8H\epsilon}q^{\hat P_{\calE_k},\pi_k}(s,a)\leq \mathrm{e}^{12H\epsilon}q^{P_{\calE_k},\pi_k}(s,a).
\end{align}
Combining these with $\mathrm{e}^{12H\epsilon}\leq2$ completes the proof.
\end{proof}

\begin{lemma}\label{lem:service}
Let $k<t$ and $(s,a)\in\calS\times\calA$.
Then $0\leq\tilde q_{k,t}(s,a)\leq2$, so that $0\leq1-(\tilde q_{k,t}(s,a)+\gamma)/(c_0\iota)\leq1$, and $\tilde q_{k,t}(s,a)\leq2\I_t(s,a;\calE_k)$ if $k\geq r_t$.
If \cref{eq:calibration} holds for $\calE_k$, then
\begin{align}
\E_t[\tilde q_{k,t}(s,a)]\geq q^{P_{\calE_k},\pi_k}(s,a),\qquad\E_t\brk*{\tilde q_{k,t}(s,a)^2}\leq2q^{P_{\calE_k},\pi_k}(s,a),
\end{align}
and the first inequality holds with equality if $k\geq r_t$.
\end{lemma}
\begin{proof}
Suppose that $k\geq r_t$.
Then $r_k=r_t$, and \cref{line:epoch} gives $R(\pi_k,\pi_{r_t})\leq\sqrt2$ and $R(\pi_t,\pi_{r_t})\leq\sqrt2$.
Hence for every $h$ and $(s_0,a_0,\ldots,s_h,a_h)$
\begin{align}
\prod_{h'=0}^h\frac{\pi_k(a_{h'}\mid s_{h'})}{\pi_t(a_{h'}\mid s_{h'})}
=\prod_{h'=0}^h\frac{\pi_k(a_{h'}\mid s_{h'})}{\pi_{r_t}(a_{h'}\mid s_{h'})}\prod_{h'=0}^h\frac{\pi_{r_t}(a_{h'}\mid s_{h'})}{\pi_t(a_{h'}\mid s_{h'})}
\leq R(\pi_k,\pi_{r_t})R(\pi_t,\pi_{r_t})
\leq2.
\end{align}
Thus $0\leq\tilde q_{k,t}(s,a)\leq2\I_t(s,a;\calE_k)\leq2\leq c_0\iota-\gamma$ by \cref{eq:service}, and \cref{lem:unbiased} gives
\begin{align}
\E_t[\tilde q_{k,t}(s,a)]=q^{P_{\calE_k},\pi_k}(s,a),\qquad
\E_t\brk*{\tilde q_{k,t}(s,a)^2}\leq2\E_t[\tilde q_{k,t}(s,a)]=2q^{P_{\calE_k},\pi_k}(s,a).
\end{align}
For $k<r_t$, $0\leq\tilde q_{k,t}(s,a)=\mathrm{e}^{4H\epsilon}q^{\hat P_{\calE_k},\pi_k}(s,a)\leq \mathrm{e}^{4H\epsilon}\leq2$ by \cref{eq:upper-probability} and the definition of $\epsilon$ in \cref{alg:main}, and if \cref{eq:calibration} holds for $\calE_k$, then \cref{lem:upper-comparison} gives
\begin{align}
q^{P_{\calE_k},\pi_k}(s,a)\leq\tilde q_{k,t}(s,a)\leq2q^{P_{\calE_k},\pi_k}(s,a)\leq2,\qquad
\tilde q_{k,t}(s,a)^2\leq2q^{P_{\calE_k},\pi_k}(s,a),
\end{align}
which completes the proof.
\end{proof}

\begin{lemma}\label{lem:total-weight}
With probability at least $1-2\delta$, for all $(s,a)$ and all $t\in[T]$ such that \cref{eq:calibration} holds for $\calE_k$ for every $k<t$,
\begin{align}\label{eq:total-weight}
\sum_{k<t}w_{k,t}(s,a)\leq c_0^2\iota^2.
\end{align}
\end{lemma}
\begin{proof}
For $k<u<t$, $0\leq(\tilde q_{k,u}(s,a)+\gamma)/(c_0\iota)\leq1$ by \cref{lem:service}, $1-x\leq \mathrm{e}^{-x}$ for $0\leq x\leq1$ and $\gamma\geq0$ give
\begin{align}
w_{k,t}(s,a)=\I_k(s,a;\calE_k)\prod_{u=k+1}^{t-1}\prn*{1-\frac{\tilde q_{k,u}(s,a)+\gamma}{c_0\iota}}\leq\exp\prn*{-\frac{1}{c_0\iota}\sum_{u=k+1}^{t-1}\tilde q_{k,u}(s,a)}.
\end{align}
By \cref{lem:bounded-sum} and a union bound, with probability at least $1-\delta$, for all $(s,a)$ and $k<t$
\begin{align}
\sum_{u=k+1}^{t-1}\tilde q_{k,u}(s,a)
\geq\frac12\sum_{u=k+1}^{t-1}\E_u[\tilde q_{k,u}(s,a)]-4\log\prn*{\frac{SAT^2}{\delta}}
\geq\frac{q^{P_{\calE_k},\pi_k}(s,a)(t-k-1)}{2}-8\iota,
\end{align}
where the second inequality uses \cref{lem:service} and $SAT^2/\delta\leq(\mathrm{e}SAT/\delta)^2$.
Combining this with $\mathrm{e}^{8/c_0}\leq2$ gives, with probability at least $1-\delta$, for all $(s,a)$ and $k<t$
\begin{align}\label{eq:weight-envelope}
w_{k,t}(s,a)\leq2\exp\prn*{-\frac{q^{P_{\calE_k},\pi_k}(s,a)(t-k-1)}{2c_0\iota}}.
\end{align}

Let $X_k=\exp\prn*{-\frac{q^{P_{\calE_k},\pi_k}(s,a)(t-k-1)}{2c_0\iota}}\I_k(s,a;\calE_k)$ for $k<t$.
Since $w_{k,t}(s,a)=0$ if $\I_k(s,a;\calE_k)=0$, \cref{eq:weight-envelope} gives $\sum_{k<t}w_{k,t}(s,a)\leq2\sum_{k<t}X_k$.
Each $X_k$ takes values in $[0,1]$ and is $\calF_{k+1}$-measurable, and
\begin{align}
\sum_{k<t}\E_k[X_k]
&=\sum_{k<t}q^{P_{\calE_k},\pi_k}(s,a)\exp\prn*{-\frac{q^{P_{\calE_k},\pi_k}(s,a)(t-k-1)}{2c_0\iota}}\\
&\leq1+\sum_{r=1}^{T}\frac{2c_0\iota}{r}
\leq1+2c_0\iota(1+\log T)
\leq5c_0\iota^2,
\end{align}
where the first inequality uses $p\mathrm{e}^{-pr/(2c_0\iota)}\leq2c_0\iota/r$ for $p\in[0,1]$ and $r\geq1$, and the last inequality uses $1\leq \iota$ and $\log T\leq \iota$.
By \cref{lem:bounded-sum} and a union bound, with probability at least $1-\delta$, for all $t\in[T]$ and $(s,a)$
\begin{align}
\sum_{k<t}X_k\leq\frac32\cdot5c_0\iota^2+2\log\prn*{\frac{SAT}{\delta}}\leq8c_0\iota^2,
\end{align}
where the last inequality uses $\log(SAT/\delta)\leq \iota$ and $c_0\iota\geq4$.
Together with \cref{eq:weight-envelope}, with probability at least $1-2\delta$, we obtain $\sum_{k<t}w_{k,t}(s,a)\leq16c_0\iota^2\leq c_0^2\iota^2$ for all such $t$ and $(s,a)$.
\end{proof}

\begin{lemma}\label{lem:stopping}
It holds that $\P(\tau=T+1)\geq1-3\delta$.
\end{lemma}
\begin{proof}
This follows from \cref{lem:calibration,lem:total-weight} and a union bound.
\end{proof}

\section{Regret Analysis of Follow-the-Regularized-Leader}\label{app:regularizer}
This appendix bounds $\sum_{t<\tau}\inpr{\hat q_t-q^{\pio},\hat g_t}$ (\cref{lem:ftrl-regret}), which is used in \cref{app:ftrl}.

\begin{definition}[{\citet[Definition~5.1.1]{nesterov2018lectures}}]\label{def:self-concordant}
Let $\calK\subseteq\R^d$ be an open convex set and $M\geq0$.
A three times continuously differentiable convex function $f:\calK\to\R$ that diverges at the boundary of $\calK$ is $M$-self-concordant if
\begin{align}
\abs*{\nabla^3f(x)[u,u,u]}\leq2M\prn*{\nabla^2f(x)[u,u]}^{3/2}\qquad\text{for all }x\in\calK\text{ and }u\in\R^d.
\end{align}
\end{definition}

\begin{lemma}[{\citet[Example~5.1.1 and Theorems~5.1.1--5.1.2]{nesterov2018lectures}}]\label{lem:sc-calculus}
The following hold.
\begin{enumerate}
\item The function $x\mapsto-\log x$ on $(0,\infty)$ is $1$-self-concordant.
\item If $f_i:\calK_i\to\R$ is $M_i$-self-concordant and $\alpha_i>0$ for $i=1,\ldots,n$, and $\bigcap_{i=1}^n\calK_i\neq\emptyset$, then $\sum_{i=1}^n\alpha_if_i$ on $\bigcap_{i=1}^n\calK_i$ is $\max_iM_i/\sqrt{\alpha_i}$-self-concordant.
\item If $f:\calK\to\R$ is $M$-self-concordant and $x\mapsto Ax+b$ is an affine map with $\set{x:Ax+b\in\calK}\neq\emptyset$, then $x\mapsto f(Ax+b)$ on $\set{x:Ax+b\in\calK}$ is $M$-self-concordant.
\end{enumerate}
\end{lemma}

\begin{lemma}[{\citet[Lemma~5.1.2 and Theorems~5.1.7--5.1.8]{nesterov2018lectures}}]\label{lem:self-concordant}
Let $f$ be $M$-self-concordant with $M>0$, and let $x$ and $y$ be in the domain of $f$.
Then the following hold.
\begin{enumerate}
\item For all $u_1,u_2,u_3\in\R^d$,
\begin{align}
\abs*{\nabla^3f(x)[u_1,u_2,u_3]}\leq2M\nrm{u_1}_{x,f}\nrm{u_2}_{x,f}\nrm{u_3}_{x,f}.
\end{align}
\item If $M\nrm{y-x}_{x,f}<1$, then
\begin{align}
\prn*{1-M\nrm{y-x}_{x,f}}^2\nabla^2f(x)\preceq\nabla^2f(y)\preceq\prn*{1-M\nrm{y-x}_{x,f}}^{-2}\nabla^2f(x).
\end{align}
\item It holds that
\begin{align}
\inpr{\nabla f(y)-\nabla f(x),y-x}\geq\frac{\nrm{y-x}_{x,f}^2}{1+M\nrm{y-x}_{x,f}}.
\end{align}
\end{enumerate}
\end{lemma}

\begin{lemma}\label{lem:third}
The regularizer $\Psi_\calE$ is $\frac32\sqrt{S^2A/H}$-self-concordant.
\end{lemma}
\begin{proof}
By the definition of $\Psi_\calE$,
\begin{align}
\Psi_\calE(q)&=\sum_{s,a}\prn*{q(s,a)\log q(s,a)+\frac{H}{SA}\log\frac{1}{q(s,a)}}+\frac{H}{S^2A}\sum_{(s,a,s')}\log\frac{1}{q(s,a,s')}\\
&\qquad+\frac{H}{S^2A}\sum_{(s,a,s')\in\calE}\prn*{\log\frac{2}{b^-(s,a,s';q)}+\log\frac{2}{b^+(s,a,s';q)}}.
\end{align}
Let $\varphi(x)=x\log x+(H/(SA))\log(1/x)$.
For $x>0$, $\varphi''(x)=1/x+H/(SAx^2)\geq\max\set{1/x,H/(SAx^2)}$, and hence
\begin{align}
\abs{\varphi'''(x)}=\frac1{x^2}+\frac{2H}{SAx^3}
=\sqrt{\frac{SA}{H}}\,\frac1x\prn*{\frac{H}{SAx^2}}^{1/2}+2\sqrt{\frac{SA}{H}}\prn*{\frac{H}{SAx^2}}^{3/2}
\leq3\sqrt{\frac{SA}{H}}\,\varphi''(x)^{3/2}.
\end{align}
Thus $\varphi$ is $\frac32\sqrt{SA/H}$-self-concordant on $(0,\infty)$.

Since $\log(2/x)=-\log(x/2)$, \cref{lem:sc-calculus} shows that $\log(1/x)$ and $\log(2/x)$ are $1$-self-concordant and that their multiples by $H/(S^2A)$ are $\sqrt{S^2A/H}$-self-concordant.
Since $q(s,a)$, $q(s,a,s')$, and $b^\pm(s,a,s';q)$ are linear in $q$, \cref{lem:sc-calculus} shows that $\Psi_\calE$ is self-concordant with constant
\begin{align}
\max\set*{\frac32\sqrt{\frac{SA}{H}},\sqrt{\frac{S^2A}{H}}}\leq\frac32\sqrt{\frac{S^2A}{H}}.
\end{align}
\end{proof}

\begin{lemma}\label{lem:ftrl-general}
Let $D_1\supseteq D_2\supseteq\cdots\supseteq D_T$ be nonempty sets, let $\psi_t:D_t\to\R$ satisfy $\psi_{t+1}\geq\psi_t$ on $D_{t+1}$ for $t=1,\ldots,T-1$, and let $g_1,\ldots,g_T$ be vectors.
For $t=1,\ldots,T$, let
\begin{align}
x_t\in\argmin_{x\in D_t}\set*{\psi_t(x)+\sum_{k=1}^{t-1}\inpr{g_k,x}},\qquad
x_t^+\in\argmin_{x\in D_t}\set*{\psi_t(x)+\sum_{k=1}^{t}\inpr{g_k,x}}.
\end{align}
Then, for every $u\in D_T$,
\begin{align}\label{eq:ftrl-general}
\sum_{t=1}^T\inpr{g_t,x_t-u}\leq\psi_T(u)-\psi_1(x_1)+\sum_{t=1}^T\inpr{g_t,x_t-x_t^+}.
\end{align}
\end{lemma}
\begin{proof}
Fix $u\in D_T$, and set $\psi_{T+1}=\psi_T$ and $x_{T+1}=u$.
For any $t=1,\ldots,T$, it holds that
\begin{align}
\psi_{t+1}(x_{t+1})+\sum_{k=1}^{t}\inpr{g_k,x_{t+1}}
&\geq\psi_t(x_{t+1})+\sum_{k=1}^{t}\inpr{g_k,x_{t+1}}\tag{by $\psi_{t+1}\geq\psi_t$}\\
&\geq\psi_t(x_t^+)+\sum_{k=1}^{t}\inpr{g_k,x_t^+}\notag\\
&=\psi_t(x_t^+)+\sum_{k=1}^{t-1}\inpr{g_k,x_t^+}+\inpr{g_t,x_t^+}\notag\\
&\geq\psi_t(x_t)+\sum_{k=1}^{t-1}\inpr{g_k,x_t}+\inpr{g_t,x_t^+},\notag
\end{align}
where the second and last inequalities follow from the definitions of $x_t^+$ and $x_t$ and from $x_{t+1},x_t^+\in D_t$.
Summing this over $t=1,\ldots,T$,
\begin{align}
\sum_{t=1}^T\inpr{g_t,x_t^+-u}\leq\psi_T(u)-\psi_1(x_1).
\end{align}
Hence
\begin{align}
\sum_{t=1}^T\inpr{g_t,x_t-u}
=\sum_{t=1}^T\inpr{g_t,x_t^+-u}+\sum_{t=1}^T\inpr{g_t,x_t-x_t^+}
\leq\psi_T(u)-\psi_1(x_1)+\sum_{t=1}^T\inpr{g_t,x_t-x_t^+}.
\end{align}
\end{proof}

The assumption of the following lemma holds by \cref{lem:local-ftrl}.
\begin{lemma}\label{lem:ftrl-regret}
If $\eta\nrm{\hat g_t}_{t,*}\leq\frac13\sqrt{H/(S^2A)}$ for every $t<\tau$, then
\begin{align}\label{eq:ftrl-regret}
\sum_{t<\tau}\inpr*{\hat q_t-q^{\pio},\hat g_t}\lesssim\frac{H\iota}{\eta}+\eta\sum_{t<\tau}\nrm{\hat g_t}_{t,*}^2+\sum_{t<\tau}\inpr*{\qo-q^{\pio},\hat g_t},
\end{align}
where $\qo=q^{P^\zeta,\pi^\zeta}$
with $\pi^\zeta=(1-\zeta)\pio+\zeta\Unif(\calA)$, $P^\zeta=(1-\zeta)P+\zeta\Unif(\calS_{h(s)+1})$, and $\zeta=\epsilon/(2T)$.
\end{lemma}
\begin{proof}
We decompose
\begin{align}\label{eq:comparator-split}
\sum_{t<\tau}\inpr*{\hat q_t-q^{\pio},\hat g_t}=\sum_{t<\tau}\inpr*{\hat q_t-\qo,\hat g_t}+\sum_{t<\tau}\inpr*{\qo-q^{\pio},\hat g_t}.
\end{align}
Below, we bound the first sum.
To this end, we first verify $\qo\in\calQ(\calE_t)$ for $t<\tau$.
Since $P^\zeta(s'\mid s,a)\geq\zeta/S$ and $\pi^\zeta(a\mid s)\geq\zeta/A$ for all $(s,a,s')$, it holds that
\begin{align}
\qo(s,a)\geq\frac{\zeta^2}{SA},\qquad\qo(s,a,s')\geq\frac{\zeta^3}{S^2A}\qquad\text{for all }(s,a,s').
\end{align}
Let $t<\tau$ and $(s,a,s')\in\calE_t$.
Then \cref{eq:good-event} gives $\abs{P(s'\mid s,a)-\Pbar(s'\mid s,a)}\leq\epsilon\Pbar(s'\mid s,a)$.
Moreover, $\abs{P^\zeta(s'\mid s,a)-P(s'\mid s,a)}\leq\zeta$, and $\zeta\leq\epsilon\Pbar(s'\mid s,a)/2$ by $\Pbar(s'\mid s,a)\geq1/T$ for active edges.
By the triangle inequality,
\begin{align}
\abs*{P^\zeta(s'\mid s,a)-\Pbar(s'\mid s,a)}\leq\frac32\epsilon\Pbar(s'\mid s,a).
\end{align}
Then, by the definition of $\qo$,
\begin{align}
\qo(s,a,s')&=\qo(s,a)P^\zeta(s'\mid s,a)\geq\prn*{1-\frac32\epsilon}\Pbar(s'\mid s,a)\qo(s,a)>\Pbar^-(s'\mid s,a)\qo(s,a),\\
\qo(s,a,s')&=\qo(s,a)P^\zeta(s'\mid s,a)\leq\prn*{1+\frac32\epsilon}\Pbar(s'\mid s,a)\qo(s,a)<\Pbar^+(s'\mid s,a)\qo(s,a),
\end{align}
and hence $\qo\in\calQ(\calE_t)$.
Moreover, $\calQ(\calE_{t+1})\subseteq\calQ(\calE_t)$ since $\calE_t\subseteq\calE_{t+1}$, and, for $q\in\calQ(\calE_{t+1})$,
\begin{align}
\Psi_{\calE_{t+1}}(q)-\Psi_{\calE_t}(q)=\frac{H}{S^2A}\sum_{(s,a,s')\in\calE_{t+1}\setminus\calE_t}\prn*{\log\frac{2}{b^-(s,a,s';q)}+\log\frac{2}{b^+(s,a,s';q)}}\geq0
\end{align}
since $b^\pm(s,a,s';q)\leq2$.
Thus, \cref{lem:ftrl-general} gives
\begin{align}\label{eq:ftrl-split}
\sum_{t<\tau}\inpr*{\hat q_t-\qo,\hat g_t}
\leq\frac{\Psi_{\calE_{\tau-1}}(\qo)-\inf_{q\in\calQ(\calE_1)}\Psi_{\calE_1}(q)}{\eta}+\sum_{t<\tau}\inpr{\hat g_t,\hat q_t-\hat q_t^+}.
\end{align}
Now, we bound the first term.
By the bounds on $\qo(s,a,s')$, for $(s,a,s')\in\calE_{\tau-1}$,
\begin{align}
b^\pm(s,a,s';\qo)\geq\prn*{2\epsilon-\frac32\epsilon}\Pbar(s'\mid s,a)\qo(s,a)\geq\frac{\epsilon}{2T}\cdot\frac{\zeta^2}{SA}.
\end{align}
Then, we have
\begin{align}
\Psi_{\calE_{\tau-1}}(\qo)\leq H\log\frac{SA}{\zeta^2}+H\log\frac{S^2A}{\zeta^3}+2H\log\frac{4SAT}{\epsilon\zeta^2}\lesssim H\iota.
\end{align}
Combining this with $\Psi_{\calE_1}(q)\geq-H\log(SA)$ for $q\in\calQ(\calE_1)$, we obtain
\begin{align}\label{eq:range}
\Psi_{\calE_{\tau-1}}(\qo)-\inf_{q\in\calQ(\calE_1)}\Psi_{\calE_1}(q)\lesssim H\iota.
\end{align}
For the second term, let $t<\tau$.
By the optimality of $\hat q_t$ and $\hat q_t^+$ over the convex set $\calQ(\calE_t)$,
\begin{align}
\inpr*{\nabla\Psi_{\calE_t}(\hat q_t)+\eta\sum_{k<t}\hat g_k,\hat q_t^+-\hat q_t}\geq0,\qquad
\inpr*{\nabla\Psi_{\calE_t}(\hat q_t^+)+\eta\sum_{k\leq t}\hat g_k,\hat q_t-\hat q_t^+}\geq0.
\end{align}
Let $M=\frac32\sqrt{S^2A/H}$.
Then
\begin{align}
\frac{\nrm{\hat q_t^+-\hat q_t}_t^2}{1+M\nrm{\hat q_t^+-\hat q_t}_t}
\leq\inpr{\nabla\Psi_{\calE_t}(\hat q_t^+)-\nabla\Psi_{\calE_t}(\hat q_t),\hat q_t^+-\hat q_t}
\leq\eta\inpr{\hat g_t,\hat q_t-\hat q_t^+}
\leq\eta\nrm{\hat g_t}_{t,*}\nrm{\hat q_t^+-\hat q_t}_t,
\end{align}
where the first inequality follows from \cref{lem:third} and the third statement of \cref{lem:self-concordant}, and the second from adding the two inequalities above.
Then, we have
\begin{align}
\nrm{\hat q_t^+-\hat q_t}_t\leq\eta\nrm{\hat g_t}_{t,*}\prn*{1+M\nrm{\hat q_t^+-\hat q_t}_t}\leq\eta\nrm{\hat g_t}_{t,*}+\frac12\nrm{\hat q_t^+-\hat q_t}_t,
\end{align}
where the last inequality follows from $\eta\nrm{\hat g_t}_{t,*}\leq\frac{1}{2M}$ by assumption.
Thus, we obtain
\begin{align}\label{eq:stability}
\nrm{\hat q_t^+-\hat q_t}_t\leq2\eta\nrm{\hat g_t}_{t,*},\qquad
\inpr{\hat g_t,\hat q_t-\hat q_t^+}\leq\nrm{\hat g_t}_{t,*}\nrm{\hat q_t^+-\hat q_t}_t\leq2\eta\nrm{\hat g_t}_{t,*}^2.
\end{align}
Substituting \cref{eq:range,eq:stability} into \cref{eq:ftrl-split}, we obtain
\begin{align}\label{eq:ftrl-smoothed}
\sum_{t<\tau}\inpr*{\hat q_t-\qo,\hat g_t}\lesssim\frac{H\iota}{\eta}+\eta\sum_{t<\tau}\nrm{\hat g_t}_{t,*}^2.
\end{align}
Combining this with \cref{eq:comparator-split} completes the proof.
\end{proof}

\section{Auxiliary Lemmas}\label{app:auxiliary}

\begin{lemma}\label{lem:effective}
Suppose that \cref{eq:calibration} holds for $\calE_t$.
Then, for every $(s,a)$,
\begin{align}
&q^{\hat P_{\calE_t},\pi_t}(s,a)\leq\hat q_t(s,a),\qquad q^{P_{\calE_t},\pi_t}(s,a)\leq2\hat q_t(s,a),\\
&q^{P_{\calE_t},\pi_t}(s,a)\leq\bar q_t(s,a),\qquad\frac12\hat q_t(s,a)\leq\bar q_t(s,a).
\end{align}
\end{lemma}
\begin{proof}
Since $\hat q_t$ is the occupancy measure of $\pi_t$ under $\hat P_t(s'\mid s,a)=\hat q_t(s,a,s')/\hat q_t(s,a)$ and $\hat P_{\calE_t}(s'\mid s,a)\leq\hat P_t(s'\mid s,a)$ for all $(s,a,s')$,
\begin{align}\label{eq:effective-hat}
q^{\hat P_{\calE_t},\pi_t}(s,a)\leq\hat q_t(s,a).
\end{align}
By \cref{lem:upper-comparison}, $q^{P_{\calE_t},\pi_t}(s,a)\leq \mathrm{e}^{4H\epsilon}q^{\hat P_{\calE_t},\pi_t}(s,a)\leq2q^{P_{\calE_t},\pi_t}(s,a)$, and $1\leq \mathrm{e}^{4H\epsilon}\leq2$ gives
\begin{align}\label{eq:effective-ratio}
q^{P_{\calE_t},\pi_t}(s,a)\leq2q^{\hat P_{\calE_t},\pi_t}(s,a),\qquad q^{\hat P_{\calE_t},\pi_t}(s,a)\leq2q^{P_{\calE_t},\pi_t}(s,a).
\end{align}
We now derive the remaining three bounds, where the last two use $\bar q_t(s,a)=\hat q_t(s,a)-q^{\hat P_{\calE_t},\pi_t}(s,a)+q^{P_{\calE_t},\pi_t}(s,a)$.
First, \cref{eq:effective-ratio,eq:effective-hat} give $q^{P_{\calE_t},\pi_t}(s,a)\leq2q^{\hat P_{\calE_t},\pi_t}(s,a)\leq2\hat q_t(s,a)$.
Second, \cref{eq:effective-hat} gives $\bar q_t(s,a)\geq q^{P_{\calE_t},\pi_t}(s,a)$.
Finally, the second inequality in \cref{eq:effective-ratio} and \cref{eq:effective-hat} give
\begin{align}
\bar q_t(s,a)\geq\hat q_t(s,a)-q^{\hat P_{\calE_t},\pi_t}(s,a)+\frac12q^{\hat P_{\calE_t},\pi_t}(s,a)=\hat q_t(s,a)-\frac12q^{\hat P_{\calE_t},\pi_t}(s,a)\geq\frac12\hat q_t(s,a).
\end{align}
\end{proof}

\begin{lemma}\label{lem:bounded-loss}
For $t<\tau$, $G_t(s,a)\leq c_0\iota$ and $V^{\hat P_{\calE_t},\pi_t}(s;G_t)\leq c_0H\iota$.
\end{lemma}
\begin{proof}
By the definition of $G_t$ and \cref{eq:good-event},
\begin{align}
G_t(s,a)=\frac{1}{c_0\iota}\sum_{k<t}w_{k,t}(s,a)\ell_k(s,a)\leq\frac{1}{c_0\iota}\sum_{k<t}w_{k,t}(s,a)\leq c_0\iota.
\end{align}
Summing over at most $H$ layers gives $V^{\hat P_{\calE_t},\pi_t}(s;G_t)\leq c_0H\iota$.
\end{proof}

\begin{lemma}\label{lem:curvature}
For an edge set $\calE$, $q\in\calQ(\calE)$, and a vector $u$ on the edges with $u(s,a)=\sum_{s'}u(s,a,s')$,
\begin{align}\label{eq:metric}
\begin{aligned}
&\sum_{s,a}\prn*{q(s,a)+\frac{H}{SA}}\prn*{\frac{u(s,a)}{q(s,a)}}^2
+H^2\sum_{(s,a,s')\in\calE}\prn*{\frac{q(s,a)u(s,a,s')-q(s,a,s')u(s,a)}{\Pbar(s'\mid s,a)q(s,a)^2}}^2\\
&\leq2\nabla^2\Psi_\calE(q)[u,u].
\end{aligned}
\end{align}
\end{lemma}
\begin{proof}
Since $b^\pm(s,a,s';q)$ is linear in $q$, differentiating $\Psi_\calE$ gives
\begin{align}
\nabla\Psi_\calE(q)[u]
&=\sum_{s,a}\prn*{\log q(s,a)+1-\frac{H}{SAq(s,a)}}u(s,a)
-\frac{H}{S^2A}\sum_{(s,a,s')}\frac{u(s,a,s')}{q(s,a,s')}\\
&\quad-\frac{H}{S^2A}\sum_{(s,a,s')\in\calE}\prn*{\frac{b^-(s,a,s';u)}{b^-(s,a,s';q)}+\frac{b^+(s,a,s';u)}{b^+(s,a,s';q)}},
\end{align}
and differentiating once more gives
\begin{align}
\nabla^2\Psi_\calE(q)[u,u]
&=\sum_{s,a}\prn*{q(s,a)+\frac{H}{SA}}\prn*{\frac{u(s,a)}{q(s,a)}}^2
+\frac{H}{S^2A}\sum_{(s,a,s')}\prn*{\frac{u(s,a,s')}{q(s,a,s')}}^2\\
&\quad+\frac{H}{S^2A}\sum_{(s,a,s')\in\calE}\prn*{\prn*{\frac{b^-(s,a,s';u)}{b^-(s,a,s';q)}}^2+\prn*{\frac{b^+(s,a,s';u)}{b^+(s,a,s';q)}}^2}.
\end{align}
For $(s,a,s')\in\calE$, substituting
\begin{align}
u(s,a,s')=\frac{q(s,a,s')}{q(s,a)}u(s,a)+\frac{q(s,a)u(s,a,s')-q(s,a,s')u(s,a)}{q(s,a)}
\end{align}
into the definitions of $b^\pm(s,a,s';u)$ gives
\begin{align}\label{eq:barrier-ratio}
\frac{b^\pm(s,a,s';u)}{b^\pm(s,a,s';q)}=\frac{u(s,a)}{q(s,a)}\mp\frac{q(s,a)u(s,a,s')-q(s,a,s')u(s,a)}{q(s,a)b^\pm(s,a,s';q)}.
\end{align}
By the definitions of $b^\pm$ and $\Pbar^\pm=(1\pm2\epsilon)\Pbar$ in \cref{eq:domain},
\begin{align}\label{eq:barrier-sum}
b^-(s,a,s';q)+b^+(s,a,s';q)=4\epsilon\Pbar(s'\mid s,a)q(s,a),
\end{align}
it holds that $b^-(s,a,s';q)b^+(s,a,s';q)\leq 4\epsilon^2\Pbar(s'\mid s,a)^2q(s,a)^2$.
Thus
\begin{align}
\abs*{\frac{q(s,a)u(s,a,s')-q(s,a,s')u(s,a)}{\epsilon\Pbar(s'\mid s,a)q(s,a)^2}}
&\leq\frac{4\epsilon\Pbar(s'\mid s,a)\abs{q(s,a)u(s,a,s')-q(s,a,s')u(s,a)}}{b^-(s,a,s';q)b^+(s,a,s';q)}\\
&=\abs*{\frac{b^-(s,a,s';u)}{b^-(s,a,s';q)}-\frac{b^+(s,a,s';u)}{b^+(s,a,s';q)}},
\end{align}
where the equality follows from subtracting the two equations in \cref{eq:barrier-ratio} and using \cref{eq:barrier-sum}.
Squaring and using $(y-z)^2\leq2y^2+2z^2$ for $y,z\in\R$,
\begin{align}
\prn*{\frac{q(s,a)u(s,a,s')-q(s,a,s')u(s,a)}{\epsilon\Pbar(s'\mid s,a)q(s,a)^2}}^2\leq2\prn*{\frac{b^-(s,a,s';u)}{b^-(s,a,s';q)}}^2+2\prn*{\frac{b^+(s,a,s';u)}{b^+(s,a,s';q)}}^2.
\end{align}
Summing this over $(s,a,s')\in\calE$,
\begin{align}
&H^2\sum_{(s,a,s')\in\calE}\prn*{\frac{q(s,a)u(s,a,s')-q(s,a,s')u(s,a)}{\Pbar(s'\mid s,a)q(s,a)^2}}^2\\
&=H^2\epsilon^2\sum_{(s,a,s')\in\calE}\prn*{\frac{q(s,a)u(s,a,s')-q(s,a,s')u(s,a)}{\epsilon\Pbar(s'\mid s,a)q(s,a)^2}}^2\\
&\leq\frac{2H}{S^2A}\sum_{(s,a,s')\in\calE}\prn*{\prn*{\frac{b^-(s,a,s';u)}{b^-(s,a,s';q)}}^2+\prn*{\frac{b^+(s,a,s';u)}{b^+(s,a,s';q)}}^2},
\end{align}
where the inequality follows from $H^2\epsilon^2\leq H/(S^2A)$.
Hence the left-hand side of \cref{eq:metric} is at most
\begin{align}
&\sum_{s,a}\prn*{q(s,a)+\frac{H}{SA}}\prn*{\frac{u(s,a)}{q(s,a)}}^2+\frac{2H}{S^2A}\sum_{(s,a,s')\in\calE}\prn*{\prn*{\frac{b^-(s,a,s';u)}{b^-(s,a,s';q)}}^2+\prn*{\frac{b^+(s,a,s';u)}{b^+(s,a,s';q)}}^2}\\
&\leq2\nabla^2\Psi_\calE(q)[u,u].
\end{align}
\end{proof}

\begin{lemma}\label{lem:local-ftrl}
For $t<\tau$, $\nrm{\hat g_t}_{t,*}\leq5c_0H\sqrt{SA}\iota$, and hence $6\eta\sqrt{S^2A/H}\,\nrm{\hat g_t}_{t,*}\leq1$.
\end{lemma}
\begin{proof}
Let $u$ be a vector on the edges.
By \cref{eq:gradient} and $\sum_{s'}u(s,a,s')=u(s,a)$,
\begin{align}
\inpr{\hat g_t,u}
&=\sum_{s,a}u(s,a)G_t(s,a)+\sum_{s,a}\frac{u(s,a)}{\hat q_t(s,a)+\gamma}\sum_{s'}\I_t(s,a,s';\calE_t)V^{\hat P_{\calE_t},\pi_t}(s';G_t)\\
&\quad-\sum_{(s,a,s')\in\calE_t}\frac{u(s,a,s')}{\hat q_t(s,a)}\I_t(s,a;\calE_t)V^{\hat P_{\calE_t},\pi_t}(s';G_t)\\
&=\sum_{s,a}\frac{u(s,a)}{\hat q_t(s,a)}\prn*{\hat q_t(s,a)G_t(s,a)+\sum_{s'}\prn*{\I_t(s,a,s';\calE_t)-\hat P_{\calE_t}(s'\mid s,a)\I_t(s,a;\calE_t)}V^{\hat P_{\calE_t},\pi_t}(s';G_t)}\\
&\quad-\sum_{(s,a,s')\in\calE_t}\I_t(s,a;\calE_t)V^{\hat P_{\calE_t},\pi_t}(s';G_t)\frac{u(s,a,s')-\hat P_{\calE_t}(s'\mid s,a)u(s,a)}{\hat q_t(s,a)}-\sum_{s,a}\frac{u(s,a)}{\hat q_t(s,a)}\beta_t(s,a)\\
&=\sum_{s,a}\frac{u(s,a)}{\hat q_t(s,a)}\prn*{\hat q_t(s,a)G_t(s,a)+\mathrm{TD}_t(s,a)-\beta_t(s,a)}\\
&\quad-\sum_{(s,a,s')\in\calE_t}\I_t(s,a;\calE_t)\Pbar(s'\mid s,a)V^{\hat P_{\calE_t},\pi_t}(s';G_t)\frac{\hat q_t(s,a)u(s,a,s')-\hat q_t(s,a,s')u(s,a)}{\Pbar(s'\mid s,a)\hat q_t(s,a)^2},\label{eq:gradient-pairing}
\end{align}
where the second equality uses $1/(\hat q_t(s,a)+\gamma)=(1-\gamma/(\hat q_t(s,a)+\gamma))/\hat q_t(s,a)$ and \cref{eq:ix-term}, and the last equality uses the definition of $\mathrm{TD}_t(s,a)$ in \cref{eq:td} and $\hat P_{\calE_t}(s'\mid s,a)=\hat q_t(s,a,s')/\hat q_t(s,a)$ on $\calE_t$.
Using $(x+y)^2\leq2x^2+2y^2$ for the two sums in \cref{eq:gradient-pairing}, and then the Cauchy--Schwarz inequality for each sum with $\I_t(s,a;\calE_t)^2=\I_t(s,a;\calE_t)$,
\begin{align}
\inpr{\hat g_t,u}^2
&\leq2\sum_{s,a}\frac{\prn*{\hat q_t(s,a)G_t(s,a)+\mathrm{TD}_t(s,a)-\beta_t(s,a)}^2}{\hat q_t(s,a)+H/(SA)}\cdot\sum_{s,a}\prn*{\hat q_t(s,a)+\frac{H}{SA}}\prn*{\frac{u(s,a)}{\hat q_t(s,a)}}^2\\
&\quad+\frac{2}{H^2}\sum_{(s,a,s')\in\calE_t}\I_t(s,a;\calE_t)\Pbar(s'\mid s,a)^2V^{\hat P_{\calE_t},\pi_t}(s';G_t)^2\\
&\qquad\cdot H^2\sum_{(s,a,s')\in\calE_t}\prn*{\frac{\hat q_t(s,a)u(s,a,s')-\hat q_t(s,a,s')u(s,a)}{\Pbar(s'\mid s,a)\hat q_t(s,a)^2}}^2.
\end{align}
By \cref{lem:curvature},
\begin{align}
\inpr{\hat g_t,u}^2
&\leq4\nabla^2\Psi_{\calE_t}(\hat q_t)[u,u]\Bigg(\sum_{s,a}\frac{\prn*{\hat q_t(s,a)G_t(s,a)+\mathrm{TD}_t(s,a)-\beta_t(s,a)}^2}{\hat q_t(s,a)+H/(SA)}\\
&\qquad+\frac{1}{H^2}\sum_{(s,a,s')\in\calE_t}\I_t(s,a;\calE_t)\Pbar(s'\mid s,a)^2V^{\hat P_{\calE_t},\pi_t}(s';G_t)^2\Bigg).
\end{align}
By the definitions $\nrm{u}_t^2=\nabla^2\Psi_{\calE_t}(\hat q_t)[u,u]$ and $\nrm{\hat g_t}_{t,*}=\sup\set{\inpr{\hat g_t,u}:\nrm{u}_t\leq1}$, we obtain
\begin{align}\label{eq:dual-coordinates}
\begin{aligned}
\nrm{\hat g_t}_{t,*}^2&\leq4\sum_{s,a}\frac{\prn*{\hat q_t(s,a)G_t(s,a)+\mathrm{TD}_t(s,a)-\beta_t(s,a)}^2}{\hat q_t(s,a)+H/(SA)}\\
&\quad+\frac{4}{H^2}\sum_{(s,a,s')\in\calE_t}\I_t(s,a;\calE_t)\Pbar(s'\mid s,a)^2V^{\hat P_{\calE_t},\pi_t}(s';G_t)^2.
\end{aligned}
\end{align}
For the first term in \cref{eq:dual-coordinates}, $(x+y)^2\leq2x^2+2y^2$, $\hat q_t(s,a)+H/(SA)\geq\max\set{\hat q_t(s,a),H/(SA)}$, and $G_t(s,a)\leq c_0\iota$ by \cref{lem:bounded-loss} give
\begin{align}
&\sum_{s,a}\frac{\prn*{\hat q_t(s,a)G_t(s,a)+\mathrm{TD}_t(s,a)-\beta_t(s,a)}^2}{\hat q_t(s,a)+H/(SA)}\notag\\
&\leq2\sum_{s,a}\hat q_t(s,a)G_t(s,a)^2+\frac{2SA}{H}\sum_{s,a}\prn*{\mathrm{TD}_t(s,a)-\beta_t(s,a)}^2\notag\\
&\leq2c_0^2H\iota^2+\frac{2SA}{H}\sum_{s,a}\prn*{\mathrm{TD}_t(s,a)-\beta_t(s,a)}^2.\label{eq:first-sum}
\end{align}
By \cref{eq:td,eq:ix-term},
\begin{align}
\mathrm{TD}_t(s,a)-\beta_t(s,a)&=\frac{\hat q_t(s,a)}{\hat q_t(s,a)+\gamma}\sum_{s'}\I_t(s,a,s';\calE_t)V^{\hat P_{\calE_t},\pi_t}(s';G_t)\\
&\quad-\I_t(s,a;\calE_t)\sum_{s'}\hat P_{\calE_t}(s'\mid s,a)V^{\hat P_{\calE_t},\pi_t}(s';G_t).
\end{align}
Then, using $0\leq V^{\hat P_{\calE_t},\pi_t}(\cdot;G_t)\leq c_0H\iota$ from \cref{lem:bounded-loss},
\begin{align}
\abs{\mathrm{TD}_t(s,a)-\beta_t(s,a)}\leq c_0H\iota\,\I_t(s,a;\calE_t).\label{eq:td-range}
\end{align}
Thus, we have
\begin{align}
\sum_{s,a}\prn*{\mathrm{TD}_t(s,a)-\beta_t(s,a)}^2\leq c_0^2H^2\iota^2\sum_{s,a}\I_t(s,a;\calE_t)\leq c_0^2H^3\iota^2.
\end{align}
For the second term in \cref{eq:dual-coordinates}, $\Pbar(s'\mid s,a)=B/n(s,a)\leq1$ and $\Pbar(s'\mid s,a)\leq P(s'\mid s,a)/(1-\epsilon)$ on $\calE_t$ by \cref{eq:good-event} give
\begin{align}
\sum_{s':(s,a,s')\in\calE_t}\Pbar(s'\mid s,a)^2\leq\sum_{s':(s,a,s')\in\calE_t}\Pbar(s'\mid s,a)\leq\frac{1}{1-\epsilon}\leq2.
\end{align}
Using this and $V^{\hat P_{\calE_t},\pi_t}(\cdot;G_t)\leq c_0H\iota$ from \cref{lem:bounded-loss}, we obtain
\begin{align}\label{eq:second-sum}
\frac{1}{H^2}\sum_{(s,a,s')\in\calE_t}\I_t(s,a;\calE_t)\Pbar(s'\mid s,a)^2V^{\hat P_{\calE_t},\pi_t}(s';G_t)^2\leq2c_0^2H\iota^2.
\end{align}
Substituting these bounds into \cref{eq:dual-coordinates},
\begin{align}
\nrm{\hat g_t}_{t,*}^2\leq4\prn*{2c_0^2H\iota^2+\frac{2SA}{H}c_0^2H^3\iota^2}+4\cdot2c_0^2H\iota^2\leq24c_0^2H^2SA\iota^2.
\end{align}
Hence $\nrm{\hat g_t}_{t,*}\leq5c_0H\sqrt{SA}\iota$, and the definition of $\eta$ in \cref{alg:main} gives $6\sqrt{S^2A/H}\,\eta\nrm{\hat g_t}_{t,*}\leq30c_0\eta \iota\sqrt{HS^3}A\leq1$.
\end{proof}

\begin{lemma}\label{lem:gradient-moments}
For $t<\tau$, $\E_t[\nrm{\hat g_t}_{t,*}^2]\leq104c_0^2HSA\iota^2$ and $\nrm{\E_t[\hat g_t]}_{t,*}^2\leq96c_0^2H\iota^2$.
\end{lemma}
\begin{proof}
We first prove the first inequality, for which we show
\begin{align}\label{eq:td-variance}
\E_t\brk*{\sum_{s,a}\mathrm{TD}_t(s,a)^2}\leq5c_0^2H^2\iota^2.
\end{align}
By \cref{eq:td},
\begin{align}
\mathrm{TD}_t(s,a)
&=\I_t(s,a;\calE_t)\ind\set{(s,a,s_{t,h(s)+1})\in\calE_t}V^{\hat P_{\calE_t},\pi_t}(s_{t,h(s)+1};G_t)\\
&\quad-\I_t(s,a;\calE_t)\sum_{s'}\hat P_{\calE_t}(s'\mid s,a)V^{\hat P_{\calE_t},\pi_t}(s';G_t).\label{eq:td-expansion}
\end{align}
Applying $\E[(X-b)^2]=\E[X^2]-\E[X]^2+(\E[X]-b)^2$ to \cref{eq:td-expansion},
\begin{align}
&\E_t\brk*{\mathrm{TD}_t(s,a)^2}\\
&=q^{P_{\calE_t},\pi_t}(s,a)\brk*{\sum_{s'}P_{\calE_t}(s'\mid s,a)V^{\hat P_{\calE_t},\pi_t}(s';G_t)^2-\prn*{\sum_{s'}P_{\calE_t}(s'\mid s,a)V^{\hat P_{\calE_t},\pi_t}(s';G_t)}^2}\\
&\quad+q^{P_{\calE_t},\pi_t}(s,a)\prn*{\sum_{s'}\prn*{P_{\calE_t}(s'\mid s,a)-\hat P_{\calE_t}(s'\mid s,a)}V^{\hat P_{\calE_t},\pi_t}(s';G_t)}^2.\label{eq:td-second-moment}
\end{align}

We first bound the second term in \cref{eq:td-second-moment}.
For $(s,a,s')\in\calE_t$, \cref{eq:good-event} and \cref{eq:domain} give $\abs{P(s'\mid s,a)-\hat P_{\calE_t}(s'\mid s,a)}\leq3\epsilon\Pbar(s'\mid s,a)$.
Then, $\sum_{s':(s,a,s')\in\calE_t}\Pbar(s'\mid s,a)\leq2$, and $V^{\hat P_{\calE_t},\pi_t}(\cdot;G_t)\leq c_0H\iota$ from \cref{lem:bounded-loss} give
\begin{align}\label{eq:td-bias}
\abs*{\sum_{s'}\prn*{P_{\calE_t}(s'\mid s,a)-\hat P_{\calE_t}(s'\mid s,a)}V^{\hat P_{\calE_t},\pi_t}(s';G_t)}\leq6\epsilon c_0H\iota.
\end{align}
Thus, by $\sum_{s,a}q^{P_{\calE_t},\pi_t}(s,a)\leq H$ and $36\epsilon^2H\leq1$, the second term summed over $(s,a)$ is at most $c_0^2H^2\iota^2$.

We next bound the first term in \cref{eq:td-second-moment}.
For every state $s$, $G_t(s,a)\leq c_0\iota$ from \cref{lem:bounded-loss}, \cref{eq:td-bias}, and $6\epsilon H\leq1$ give 
\begin{align}
&V^{\hat P_{\calE_t},\pi_t}(s;G_t)-\sum_a\pi_t(a\mid s)\sum_{s'}P_{\calE_t}(s'\mid s,a)V^{\hat P_{\calE_t},\pi_t}(s';G_t)\\
&=\sum_a\pi_t(a\mid s)\prn*{G_t(s,a)+\sum_{s'}\prn*{\hat P_{\calE_t}(s'\mid s,a)-P_{\calE_t}(s'\mid s,a)}V^{\hat P_{\calE_t},\pi_t}(s';G_t)}
\in[-2c_0\iota,2c_0\iota].
\end{align}
By $V^{\hat P_{\calE_t},\pi_t}(\cdot;G_t)\leq c_0H\iota$ from \cref{lem:bounded-loss}, we then have
\begin{align}
&V^{\hat P_{\calE_t},\pi_t}(s;G_t)^2-\prn*{\sum_a\pi_t(a\mid s)\sum_{s'}P_{\calE_t}(s'\mid s,a)V^{\hat P_{\calE_t},\pi_t}(s';G_t)}^2\notag\\
&=\prn*{V^{\hat P_{\calE_t},\pi_t}(s;G_t)-\sum_a\pi_t(a\mid s)\sum_{s'}P_{\calE_t}(s'\mid s,a)V^{\hat P_{\calE_t},\pi_t}(s';G_t)}\notag\\
&\qquad\times\prn*{V^{\hat P_{\calE_t},\pi_t}(s;G_t)+\sum_a\pi_t(a\mid s)\sum_{s'}P_{\calE_t}(s'\mid s,a)V^{\hat P_{\calE_t},\pi_t}(s';G_t)}\notag\\
&\leq2c_0\iota\cdot2c_0H\iota=4c_0^2H\iota^2.\label{eq:bellman-square}
\end{align}
Summing the first term over $(s,a)$,
\begin{align}
&\sum_{s,a}q^{P_{\calE_t},\pi_t}(s,a)\brk*{\sum_{s'}P_{\calE_t}(s'\mid s,a)V^{\hat P_{\calE_t},\pi_t}(s';G_t)^2-\prn*{\sum_{s'}P_{\calE_t}(s'\mid s,a)V^{\hat P_{\calE_t},\pi_t}(s';G_t)}^2}\\
&\leq\sum_{s\neq s_0}q^{P_{\calE_t},\pi_t}(s)V^{\hat P_{\calE_t},\pi_t}(s;G_t)^2-\sum_{s}q^{P_{\calE_t},\pi_t}(s)\prn*{\sum_a\pi_t(a\mid s)\sum_{s'}P_{\calE_t}(s'\mid s,a)V^{\hat P_{\calE_t},\pi_t}(s';G_t)}^2\\
&\leq\sum_{s\neq s_0}q^{P_{\calE_t},\pi_t}(s)V^{\hat P_{\calE_t},\pi_t}(s;G_t)^2-\sum_{s}q^{P_{\calE_t},\pi_t}(s)V^{\hat P_{\calE_t},\pi_t}(s;G_t)^2+4c_0^2H^2\iota^2 \tag{by \cref{eq:bellman-square}}\\
&=4c_0^2H^2\iota^2-V^{\hat P_{\calE_t},\pi_t}(s_0;G_t)^2\leq4c_0^2H^2\iota^2.
\end{align}
Combining the bounds on the two terms proves \cref{eq:td-variance}.

By \cref{eq:ix-term} and \cref{lem:bounded-loss},
\begin{align}\label{eq:h-bound}
0\leq \beta_t(s,a)\leq\frac{c_0\gamma H\iota\,\I_t(s,a;\calE_t)}{\hat q_t(s,a)+\gamma}.
\end{align}
Since $\E_t[\I_t(s,a;\calE_t)]=q^{P_{\calE_t},\pi_t}(s,a)\leq2\hat q_t(s,a)$ by \cref{eq:indicator-mean} and \cref{lem:effective}, this gives
\begin{align}
\E_t\brk*{\beta_t(s,a)^2}\leq\frac{2c_0^2\gamma^2H^2\iota^2\hat q_t(s,a)}{\prn*{\hat q_t(s,a)+\gamma}^2}\leq\frac{c_0^2\gamma H^2\iota^2}{2},
\end{align}
where the last inequality follows from $(\hat q_t(s,a)+\gamma)^2-4\gamma\hat q_t(s,a)=(\hat q_t(s,a)-\gamma)^2\geq0$.
Hence $(x+y)^2\leq2x^2+2y^2$, \cref{eq:td-variance}, and $\gamma SA\leq1$ give
\begin{align}\label{eq:td-h-variance}
\E_t\brk*{\sum_{s,a}\prn*{\mathrm{TD}_t(s,a)-\beta_t(s,a)}^2}\leq10c_0^2H^2\iota^2+c_0^2\gamma SAH^2\iota^2\leq11c_0^2H^2\iota^2.
\end{align}
Taking expectation in \cref{eq:dual-coordinates} and using \cref{eq:first-sum}, \cref{eq:second-sum}, and \cref{eq:td-h-variance},
\begin{align}
\E_t\brk*{\nrm{\hat g_t}_{t,*}^2}\leq4\prn*{2c_0^2H\iota^2+\frac{2SA}{H}\cdot11c_0^2H^2\iota^2}+4\cdot2c_0^2H\iota^2\leq104c_0^2HSA\iota^2.
\end{align}
We next prove the second inequality.
Since $\E_t[\I_t(s,a;\calE_t)]=q^{P_{\calE_t},\pi_t}(s,a)$, taking $\E_t$ in \cref{eq:gradient-pairing} gives, for every vector $u$ on the edges,
\begin{align}
\inpr{\E_t[\hat g_t],u}
&=\sum_{s,a}\frac{u(s,a)}{\hat q_t(s,a)}\prn*{\hat q_t(s,a)G_t(s,a)+\E_t[\mathrm{TD}_t(s,a)]-\E_t[\beta_t(s,a)]}\\
&\quad-\sum_{(s,a,s')\in\calE_t}q^{P_{\calE_t},\pi_t}(s,a)\Pbar(s'\mid s,a)V^{\hat P_{\calE_t},\pi_t}(s';G_t)\frac{\hat q_t(s,a)u(s,a,s')-\hat q_t(s,a,s')u(s,a)}{\Pbar(s'\mid s,a)\hat q_t(s,a)^2}.
\end{align}
In the same way as \cref{eq:dual-coordinates} is derived from \cref{eq:gradient-pairing}, we obtain
\begin{align}
\nrm{\E_t[\hat g_t]}_{t,*}^2
&\leq4\sum_{s,a}\frac{\prn*{\hat q_t(s,a)G_t(s,a)+\E_t[\mathrm{TD}_t(s,a)]-\E_t[\beta_t(s,a)]}^2}{\hat q_t(s,a)+H/(SA)}\\
&\quad+\frac{4}{H^2}\sum_{(s,a,s')\in\calE_t}q^{P_{\calE_t},\pi_t}(s,a)^2\Pbar(s'\mid s,a)^2V^{\hat P_{\calE_t},\pi_t}(s';G_t)^2\\
&\leq8\sum_{s,a}\frac{\prn*{\hat q_t(s,a)G_t(s,a)+\E_t[\mathrm{TD}_t(s,a)]}^2}{\hat q_t(s,a)+H/(SA)}+8\sum_{s,a}\frac{\E_t[\beta_t(s,a)]^2}{\hat q_t(s,a)+H/(SA)}\\
&\quad+16c_0^2\iota^2\sum_{s,a}\hat q_t(s,a)\sum_{s':(s,a,s')\in\calE_t}\Pbar(s'\mid s,a)^2\\
&\leq32c_0^2\iota^2\sum_{s,a}\hat q_t(s,a)+32c_0^2\gamma^2HS^2A^2\iota^2+16c_0^2\iota^2\sum_{s,a}\hat q_t(s,a)\sum_{s':(s,a,s')\in\calE_t}\Pbar(s'\mid s,a)^2\\
&\leq32c_0^2H\iota^2+32c_0^2H\iota^2+32c_0^2H\iota^2=96c_0^2H\iota^2,\tag{by $\sum_{s,a}\hat q_t(s,a)=H$, $\sum_{s'}\Pbar(s'\mid s,a)^2\leq2$, $\gamma SA\leq1$}
\end{align}
where the second inequality follows from $(x+y)^2\leq2x^2+2y^2$, $q^{P_{\calE_t},\pi_t}(s,a)\leq2\hat q_t(s,a)$ by \cref{lem:effective}, $\hat q_t(s,a)\leq1$, and $V^{\hat P_{\calE_t},\pi_t}(s';G_t)\leq c_0H\iota$ by \cref{lem:bounded-loss}, and the third from $\hat q_t(s,a)+H/(SA)\geq\max\set{\hat q_t(s,a),H/(SA)}$, $\E_t[\beta_t(s,a)]\leq2c_0\gamma H\iota$ by \cref{eq:h-bound} with $\E_t[\I_t(s,a;\calE_t)]\leq2\hat q_t(s,a)$, and
\begin{align}\label{eq:td-mean}
\abs*{\hat q_t(s,a)G_t(s,a)+\E_t[\mathrm{TD}_t(s,a)]}\leq c_0\iota\hat q_t(s,a)+6\epsilon c_0H\iota q^{P_{\calE_t},\pi_t}(s,a)\leq2c_0\iota\hat q_t(s,a).
\end{align}
The first inequality in \cref{eq:td-mean} follows from \cref{lem:bounded-loss} and \cref{eq:td-bias} with $\E_t[\mathrm{TD}_t(s,a)]=q^{P_{\calE_t},\pi_t}(s,a)\sum_{s'}(P_{\calE_t}(s'\mid s,a)-\hat P_{\calE_t}(s'\mid s,a))V^{\hat P_{\calE_t},\pi_t}(s';G_t)$.
The second follows from $6\epsilon Hq^{P_{\calE_t},\pi_t}(s,a)\leq\hat q_t(s,a)$, which holds since $q^{P_{\calE_t},\pi_t}(s,a)\leq2\hat q_t(s,a)$ by \cref{lem:effective} and $12\epsilon H\leq1$.
\end{proof}

\begin{lemma}\label{lem:update}
For $t<\tau$,
\begin{align}\label{eq:expected-update}
\nrm{\E_t[\hat q_t^+-\hat q_t]}_t\leq\eta\nrm{\E_t[\hat g_t]}_{t,*}+48\eta^2\sqrt{\frac{S^2A}{H}}\E_t\brk*{\nrm{\hat g_t}_{t,*}^2}.
\end{align}
\end{lemma}
\begin{proof}
Since $\Psi_{\calE_t}$ is $\frac32\sqrt{S^2A/H}$-self-concordant by \cref{lem:third}, for $\theta\in[0,1]$, we have
\begin{align}
\frac32\sqrt{\frac{S^2A}{H}}\nrm{\theta(\hat q_t^+-\hat q_t)}_t\leq3\sqrt{\frac{S^2A}{H}}\,\eta\nrm{\hat g_t}_{t,*}\leq\frac12
\end{align}
by $\theta\leq1$, $\nrm{\hat q_t^+-\hat q_t}_t\leq2\eta\nrm{\hat g_t}_{t,*}$ in \cref{eq:stability}, and \cref{lem:local-ftrl}.
Hence the second statement of \cref{lem:self-concordant} with $x=\hat q_t$ and $y=\hat q_t+\theta(\hat q_t^+-\hat q_t)$ gives, for every vector $u$ on the edges,
\begin{align}\label{eq:hessian-comparison}
\nabla^2\Psi_{\calE_t}(\hat q_t+\theta(\hat q_t^+-\hat q_t))[u,u]\leq4\nabla^2\Psi_{\calE_t}(\hat q_t)[u,u].
\end{align}

Since $\hat q_t$ and $\E_t[\hat q_t^+]$ lie in the convex set $\calQ(\calE_t)$, $\hat q_t+\lambda\E_t[\hat q_t^+-\hat q_t]$ and $\hat q_t^++\lambda\E_t[\hat q_t^+-\hat q_t]$ lie in $\calQ(\calE_t)$ for small $\abs\lambda$, as the equality constraints in \cref{eq:domain} are preserved and $\hat q_t$ and $\hat q_t^+$ satisfy the inequality constraints in \cref{eq:domain}.
At a minimizer over $\calQ(\calE_t)$, the gradient of the objective has zero inner product with such a direction, and hence
\begin{align}\label{eq:update-optimality}
\inpr*{\nabla\Psi_{\calE_t}(\hat q_t)+\eta\sum_{k<t}\hat g_k,\E_t[\hat q_t^+-\hat q_t]}=0,\qquad\inpr*{\nabla\Psi_{\calE_t}(\hat q_t^+)+\eta\sum_{k\leq t}\hat g_k,\E_t[\hat q_t^+-\hat q_t]}=0.
\end{align}

Then
\begin{align}\label{eq:update-identity}
\begin{aligned}
\nrm{\E_t[\hat q_t^+-\hat q_t]}_t^2
&=\E_t\brk*{\nabla^2\Psi_{\calE_t}(\hat q_t)[\hat q_t^+-\hat q_t,\E_t[\hat q_t^+-\hat q_t]]}\\
&=\E_t\brk*{\inpr*{\nabla\Psi_{\calE_t}(\hat q_t^+)-\nabla\Psi_{\calE_t}(\hat q_t),\E_t[\hat q_t^+-\hat q_t]}}\\
&\quad-\E_t\brk*{\int_0^1(1-\theta)\nabla^3\Psi_{\calE_t}(\hat q_t+\theta(\hat q_t^+-\hat q_t))[\hat q_t^+-\hat q_t,\hat q_t^+-\hat q_t,\E_t[\hat q_t^+-\hat q_t]]\,d\theta}\\
&=-\eta\inpr*{\E_t[\hat g_t],\E_t[\hat q_t^+-\hat q_t]}\\
&\quad-\E_t\brk*{\int_0^1(1-\theta)\nabla^3\Psi_{\calE_t}(\hat q_t+\theta(\hat q_t^+-\hat q_t))[\hat q_t^+-\hat q_t,\hat q_t^+-\hat q_t,\E_t[\hat q_t^+-\hat q_t]]\,d\theta},
\end{aligned}
\end{align}
where the second equality follows from Taylor's theorem $f(1)=f(0)+f'(0)+\int_0^1(1-\theta)f''(\theta)\,d\theta$ for $f(\theta)=\inpr{\nabla\Psi_{\calE_t}(\hat q_t+\theta(\hat q_t^+-\hat q_t)),\E_t[\hat q_t^+-\hat q_t]}$, whose derivatives are
\begin{align}
f'(\theta)&=\nabla^2\Psi_{\calE_t}(\hat q_t+\theta(\hat q_t^+-\hat q_t))[\hat q_t^+-\hat q_t,\E_t[\hat q_t^+-\hat q_t]],\\
f''(\theta)&=\nabla^3\Psi_{\calE_t}(\hat q_t+\theta(\hat q_t^+-\hat q_t))[\hat q_t^+-\hat q_t,\hat q_t^+-\hat q_t,\E_t[\hat q_t^+-\hat q_t]],
\end{align}
and the third from \cref{eq:update-optimality}.

We bound the two terms in \cref{eq:update-identity}.
For the first term, $-\eta\inpr{\E_t[\hat g_t],\E_t[\hat q_t^+-\hat q_t]}\leq\eta\nrm{\E_t[\hat g_t]}_{t,*}\nrm{\E_t[\hat q_t^+-\hat q_t]}_t$.
For the second term, for $\theta\in[0,1]$,
\begin{align}
&\abs*{\nabla^3\Psi_{\calE_t}(\hat q_t+\theta(\hat q_t^+-\hat q_t))[\hat q_t^+-\hat q_t,\hat q_t^+-\hat q_t,\E_t[\hat q_t^+-\hat q_t]]}\\
&\leq3\sqrt{\frac{S^2A}{H}}\nrm{\hat q_t^+-\hat q_t}_{\hat q_t+\theta(\hat q_t^+-\hat q_t),\Psi_{\calE_t}}^2\nrm{\E_t[\hat q_t^+-\hat q_t]}_{\hat q_t+\theta(\hat q_t^+-\hat q_t),\Psi_{\calE_t}}\\
&\leq24\sqrt{\frac{S^2A}{H}}\nrm{\hat q_t^+-\hat q_t}_t^2\nrm{\E_t[\hat q_t^+-\hat q_t]}_t,
\end{align}
where the first inequality follows from the first statement of \cref{lem:self-concordant} with \cref{lem:third}, and the second from \cref{eq:hessian-comparison}.
Substituting these bounds into \cref{eq:update-identity} with $\int_0^1(1-\theta)\,d\theta=\frac12$,
\begin{align}
\nrm{\E_t[\hat q_t^+-\hat q_t]}_t^2
&\leq\prn*{\eta\nrm{\E_t[\hat g_t]}_{t,*}+12\sqrt{\frac{S^2A}{H}}\E_t\brk*{\nrm{\hat q_t^+-\hat q_t}_t^2}}\nrm{\E_t[\hat q_t^+-\hat q_t]}_t\\
&\leq\prn*{\eta\nrm{\E_t[\hat g_t]}_{t,*}+48\eta^2\sqrt{\frac{S^2A}{H}}\E_t\brk*{\nrm{\hat g_t}_{t,*}^2}}\nrm{\E_t[\hat q_t^+-\hat q_t]}_t,
\end{align}
where the second inequality follows from $\nrm{\hat q_t^+-\hat q_t}_t\leq2\eta\nrm{\hat g_t}_{t,*}$ in \cref{eq:stability}.
This completes the proof.
\end{proof}

\begin{lemma}\label{lem:ratio}
For $t<\tau$ and a vector $u$ on the edges with $u(s,a)=\sum_{s'}u(s,a,s')$,
\begin{align}
&\sum_{s,a}\prn*{\frac{u(s,a)}{\hat q_t(s,a)}}^2\leq\frac{2SA}{H}\nrm{u}_t^2,\label{eq:relative-coordinates}\\
&\sum_{(s,a,s')\in\calE_t}\prn*{\frac{u(s,a,s')}{\hat q_t(s,a,s')}-\frac{u(s,a)}{\hat q_t(s,a)}}^2\leq\frac{8}{H^2}\nrm{u}_t^2,\label{eq:relative-edges}\\
&\sum_{h<H}\max_{(s,a)\in\calS_h\times\calA}\abs*{\frac{u(s,a)}{\hat q_t(s,a)}}\leq\sqrt{2SA}\nrm{u}_t,\label{eq:layer-sums}\\
&\sum_{h<H}\max_{(s,a,s')\in\calE_t\cap(\calS_h\times\calA\times\calS_{h+1})}\abs*{\frac{u(s,a,s')}{\hat q_t(s,a,s')}-\frac{u(s,a)}{\hat q_t(s,a)}}\leq\frac{2\sqrt2}{\sqrt H}\nrm{u}_t.\label{eq:layer-edges}
\end{align}
\end{lemma}
\begin{proof}

For \cref{eq:relative-coordinates},
\begin{align}
\sum_{s,a}\prn*{\frac{u(s,a)}{\hat q_t(s,a)}}^2\leq\frac{SA}{H}\sum_{s,a}\prn*{\hat q_t(s,a)+\frac{H}{SA}}\prn*{\frac{u(s,a)}{\hat q_t(s,a)}}^2\leq\frac{2SA}{H}\nrm{u}_t^2,
\end{align}
where the first inequality follows from $\hat q_t(s,a)+H/(SA)\geq H/(SA)$, and the second from \cref{lem:curvature}.

For \cref{eq:relative-edges},
\begin{align}
\sum_{(s,a,s')\in\calE_t}\prn*{\frac{u(s,a,s')}{\hat q_t(s,a,s')}-\frac{u(s,a)}{\hat q_t(s,a)}}^2
&=\sum_{(s,a,s')\in\calE_t}\prn*{\frac{\hat q_t(s,a)u(s,a,s')-\hat q_t(s,a,s')u(s,a)}{\hat q_t(s,a)\hat q_t(s,a,s')}}^2\\
&\leq4\sum_{(s,a,s')\in\calE_t}\prn*{\frac{\hat q_t(s,a)u(s,a,s')-\hat q_t(s,a,s')u(s,a)}{\Pbar(s'\mid s,a)\hat q_t(s,a)^2}}^2\leq\frac{8}{H^2}\nrm{u}_t^2,
\end{align}
where the first inequality follows from $\hat q_t(s,a,s')\geq(1-2\epsilon)\Pbar(s'\mid s,a)\hat q_t(s,a)\geq\Pbar(s'\mid s,a)\hat q_t(s,a)/2$ on $\calE_t$ by \cref{eq:domain} and $\epsilon\leq 1/8$, and the second from \cref{lem:curvature}.

For \cref{eq:layer-sums},
\begin{align}
\sum_{h<H}\max_{(s,a)\in\calS_h\times\calA}\abs*{\frac{u(s,a)}{\hat q_t(s,a)}}
\leq\sum_{h<H}\sqrt{\sum_{(s,a)\in\calS_h\times\calA}\prn*{\frac{u(s,a)}{\hat q_t(s,a)}}^2}
\leq\sqrt{H\sum_{s,a}\prn*{\frac{u(s,a)}{\hat q_t(s,a)}}^2}
\leq\sqrt{2SA}\nrm{u}_t,
\end{align}
where the second inequality follows from the Cauchy--Schwarz inequality, and the third from \cref{eq:relative-coordinates}.

In the same way, for \cref{eq:layer-edges},
\begin{align}
&\sum_{h<H}\max_{(s,a,s')\in\calE_t\cap(\calS_h\times\calA\times\calS_{h+1})}\abs*{\frac{u(s,a,s')}{\hat q_t(s,a,s')}-\frac{u(s,a)}{\hat q_t(s,a)}}\\
&\leq\sum_{h<H}\sqrt{\sum_{(s,a,s')\in\calE_t\cap(\calS_h\times\calA\times\calS_{h+1})}\prn*{\frac{u(s,a,s')}{\hat q_t(s,a,s')}-\frac{u(s,a)}{\hat q_t(s,a)}}^2}\\
&\leq\sqrt{H\sum_{(s,a,s')\in\calE_t}\prn*{\frac{u(s,a,s')}{\hat q_t(s,a,s')}-\frac{u(s,a)}{\hat q_t(s,a)}}^2}
\leq\frac{2\sqrt2}{\sqrt H}\nrm{u}_t,
\end{align}
where the last inequality follows from \cref{eq:relative-edges}.
\end{proof}

\begin{lemma}\label{lem:motion}
For $t<\tau$ and all $(s,a)$,
\begin{align}
&\abs{\bar q_t^+(s,a)-\bar q_t(s,a)}\leq\frac{\bar q_t(s,a)}4,\label{eq:pointwise-motion}\\
&\sum_{s,a}\frac{\abs{\E_t[\bar q_t^+(s,a)-\bar q_t(s,a)]}}{\hat q_t(s,a)}\lesssim\eta SA\iota+\eta^2HS^3A^3\iota^2,\label{eq:drift}\\
&\sum_{s,a}\frac{\E_t[(\bar q_t^+(s,a)-\bar q_t(s,a))^2]}{\hat q_t(s,a)^2}\lesssim\eta^2S^2A^2\iota^2.\label{eq:qv}
\end{align}
\end{lemma}
\begin{proof}

We first decompose $\bar q_t^+-\bar q_t$.
Fix $(s,a)\in\calS_h\times\calA$.
Since $P_{\calE_t}$, $\hat P_{\calE_t}$, and $\hat P^+_{\calE_t}$ vanish outside $\calE_t$, writing the occupancy measures as in \cref{eq:trajectory-sum} gives
\begin{align}
&\bar q_t^+(s,a)-\bar q_t(s,a)-\prn*{\hat q_t^+(s,a)-\hat q_t(s,a)}\\
&=q^{P_{\calE_t},\pi_t^+}(s,a)-q^{P_{\calE_t},\pi_t}(s,a)-q^{\hat P^+_{\calE_t},\pi_t^+}(s,a)+q^{\hat P_{\calE_t},\pi_t}(s,a)\\
&=\prn*{q^{P_{\calE_t},\pi_t^+}(s,a)-q^{P_{\calE_t},\pi_t}(s,a)-q^{\hat P_{\calE_t},\pi_t^+}(s,a)+q^{\hat P_{\calE_t},\pi_t}(s,a)}
-\prn*{q^{\hat P^+_{\calE_t},\pi_t^+}(s,a)-q^{\hat P_{\calE_t},\pi_t^+}(s,a)}\\
&=\sum_{s_0,a_0,\ldots,s_{h-1},a_{h-1}}\Biggl(\prn*{\prod_{h'=0}^{h-1}P_{\calE_t}(s_{h'+1}\mid s_{h'},a_{h'})-\prod_{h'=0}^{h-1}\hat P_{\calE_t}(s_{h'+1}\mid s_{h'},a_{h'})}\\
&\hspace{8em}\times\prn*{\prod_{h'=0}^{h}\pi_t^+(a_{h'}\mid s_{h'})-\prod_{h'=0}^{h}\pi_t(a_{h'}\mid s_{h'})}\\
&\hspace{6em}-\prn*{\prod_{h'=0}^{h-1}\hat P^+_{\calE_t}(s_{h'+1}\mid s_{h'},a_{h'})-\prod_{h'=0}^{h-1}\hat P_{\calE_t}(s_{h'+1}\mid s_{h'},a_{h'})}\prod_{h'=0}^{h}\pi_t^+(a_{h'}\mid s_{h'})\Biggr)\\
&=\sum_{s_0,a_0,\ldots,s_{h-1},a_{h-1}}\prod_{h'=0}^{h}\pi_t(a_{h'}\mid s_{h'})\prn*{\prod_{h'=0}^{h-1}P_{\calE_t}(s_{h'+1}\mid s_{h'},a_{h'})-\prod_{h'=0}^{h-1}\hat P_{\calE_t}(s_{h'+1}\mid s_{h'},a_{h'})}\\
&\hspace{8em}\times\prn*{\prod_{h'=0}^{h}\frac{\pi_t^+(a_{h'}\mid s_{h'})}{\pi_t(a_{h'}\mid s_{h'})}-1}\\
&\quad-\sum_{s_0,a_0,\ldots,s_{h-1},a_{h-1}}\prod_{h'=0}^{h}\pi_t(a_{h'}\mid s_{h'})\prod_{h'=0}^{h-1}\hat P_{\calE_t}(s_{h'+1}\mid s_{h'},a_{h'})\\
&\hspace{8em}\times\prn*{\prod_{h'=0}^{h-1}\frac{\hat P^+_{\calE_t}(s_{h'+1}\mid s_{h'},a_{h'})}{\hat P_{\calE_t}(s_{h'+1}\mid s_{h'},a_{h'})}-1}\prod_{h'=0}^{h}\frac{\pi_t^+(a_{h'}\mid s_{h'})}{\pi_t(a_{h'}\mid s_{h'})},\label{eq:motion-split}
\end{align}
where the sums are over trajectories with $(s_{h'},a_{h'},s_{h'+1})\in\calE_t$ for $h'<h$.

By \cref{eq:kernel-comparison}, $\mathrm{e}^{-4\epsilon}\leq P_{\calE_t}(s'\mid s,a)/\hat P_{\calE_t}(s'\mid s,a)\leq \mathrm{e}^{4\epsilon}$ for $(s,a,s')\in\calE_t$, and hence
\begin{align}
\abs*{\prod_{h'=0}^{h-1}P_{\calE_t}(s_{h'+1}\mid s_{h'},a_{h'})-\prod_{h'=0}^{h-1}\hat P_{\calE_t}(s_{h'+1}\mid s_{h'},a_{h'})}
&\leq(\mathrm{e}^{4H\epsilon}-1)\prod_{h'=0}^{h-1}\hat P_{\calE_t}(s_{h'+1}\mid s_{h'},a_{h'})\\
&\leq8H\epsilon\prod_{h'=0}^{h-1}\hat P_{\calE_t}(s_{h'+1}\mid s_{h'},a_{h'}),\label{eq:kernel-difference}
\end{align}
where the last inequality follows from $\mathrm{e}^x-1\leq2x$ for $0\leq x\leq1$ and $4H\epsilon\leq1$.

We next bound the ratios $\pi_t^+(a\mid s)/\pi_t(a\mid s)$ and $\hat P^+_{\calE_t}(s'\mid s,a)/\hat P_{\calE_t}(s'\mid s,a)$ in \cref{eq:motion-split}.
By \cref{eq:stability} and \cref{lem:local-ftrl},
\begin{align}\label{eq:step-size}
\nrm{\hat q_t^+-\hat q_t}_t\leq2\eta\nrm{\hat g_t}_{t,*}\leq10c_0\eta H\sqrt{SA}\iota.
\end{align}
Then, for every $(s,a)$,
\begin{align}\label{eq:relative-step}
\abs*{\frac{\hat q_t^+(s,a)}{\hat q_t(s,a)}-1}&\leq\sqrt{\sum_{s',a'}\prn*{\frac{\hat q_t^+(s',a')}{\hat q_t(s',a')}-1}^2}\\
&\leq\sqrt{\frac{2SA}{H}}\nrm{\hat q_t^+-\hat q_t}_t\leq10\sqrt2c_0\eta SA\sqrt H\iota\leq\frac18,
\end{align}
where the second inequality follows from \cref{eq:relative-coordinates} in \cref{lem:ratio} with $u=\hat q_t^+-\hat q_t$, the third from \cref{eq:step-size}, and the last from the definition of $\eta$ in \cref{alg:main}.
Moreover, for every state $s$,
\begin{align}\label{eq:relative-state}
\abs*{\frac{\sum_b\hat q_t^+(s,b)}{\sum_b\hat q_t(s,b)}-1}=\abs*{\sum_b\pi_t(b\mid s)\prn*{\frac{\hat q_t^+(s,b)}{\hat q_t(s,b)}-1}}\leq\max_b\abs*{\frac{\hat q_t^+(s,b)}{\hat q_t(s,b)}-1}\leq\frac18,
\end{align}
where the equality follows from $\pi_t(b\mid s)=\hat q_t(s,b)/\sum_{b'}\hat q_t(s,b')$ in \cref{eq:policy-model}, and the last inequality from \cref{eq:relative-step}.

Since $\pi_t(a\mid s)=\hat q_t(s,a)/\sum_b\hat q_t(s,b)$ and $\hat P_{\calE_t}(s'\mid s,a)=\hat q_t(s,a,s')/\hat q_t(s,a)$ for $(s,a,s')\in\calE_t$, and similarly for $\pi_t^+$ and $\hat P^+_{\calE_t}$,
\begin{align}
\frac{\pi_t^+(a\mid s)}{\pi_t(a\mid s)}=\frac{\hat q_t^+(s,a)/\hat q_t(s,a)}{\sum_b\hat q_t^+(s,b)/\sum_b\hat q_t(s,b)},\qquad
\frac{\hat P^+_{\calE_t}(s'\mid s,a)}{\hat P_{\calE_t}(s'\mid s,a)}=\frac{\hat q_t^+(s,a,s')/\hat q_t(s,a,s')}{\hat q_t^+(s,a)/\hat q_t(s,a)}.
\end{align}
Hence
\begin{align}
&\abs*{\frac{\pi_t^+(a\mid s)}{\pi_t(a\mid s)}-1}\leq4\max_b\abs*{\frac{\hat q_t^+(s,b)}{\hat q_t(s,b)}-1},\label{eq:policy-ratio}\\
&\abs*{\frac{\pi_t^+(a\mid s)}{\pi_t(a\mid s)}-1-\prn*{\frac{\hat q_t^+(s,a)}{\hat q_t(s,a)}-\frac{\sum_b\hat q_t^+(s,b)}{\sum_b\hat q_t(s,b)}}}\leq4\max_b\prn*{\frac{\hat q_t^+(s,b)}{\hat q_t(s,b)}-1}^2,\label{eq:policy-ratio-linear}\\
&\abs*{\frac{\hat P^+_{\calE_t}(s'\mid s,a)}{\hat P_{\calE_t}(s'\mid s,a)}-1}\leq2\abs*{\frac{\hat q_t^+(s,a,s')}{\hat q_t(s,a,s')}-\frac{\hat q_t^+(s,a)}{\hat q_t(s,a)}},\label{eq:kernel-ratio}\\
&\abs*{\frac{\hat P^+_{\calE_t}(s'\mid s,a)}{\hat P_{\calE_t}(s'\mid s,a)}-1-\prn*{\frac{\hat q_t^+(s,a,s')}{\hat q_t(s,a,s')}-\frac{\hat q_t^+(s,a)}{\hat q_t(s,a)}}}\leq\prn*{\frac{\hat q_t^+(s,a,s')}{\hat q_t(s,a,s')}-\frac{\hat q_t^+(s,a)}{\hat q_t(s,a)}}^2+\prn*{\frac{\hat q_t^+(s,a)}{\hat q_t(s,a)}-1}^2,\label{eq:kernel-ratio-linear}
\end{align}
where, writing each ratio above as $x/y$ with $\abs{y-1}\leq\frac18$ by \cref{eq:relative-step,eq:relative-state}, \cref{eq:policy-ratio} follows from $\abs{x/y-1}=\abs{x-y}/y\leq2\abs{x-y}\leq2(\abs{x-1}+\abs{y-1})$ and $\abs{x-1},\abs{y-1}\leq\max_b\abs{\hat q_t^+(s,b)/\hat q_t(s,b)-1}$ by \cref{eq:relative-state}, \cref{eq:policy-ratio-linear} from $\abs{x/y-1-(x-y)}=\abs{x-y}\abs{y-1}/y\leq2\abs{x-y}\abs{y-1}$ and the same bounds on $\abs{x-1}$ and $\abs{y-1}$, \cref{eq:kernel-ratio} from $\abs{x/y-1}=\abs{x-y}/y\leq2\abs{x-y}$, and \cref{eq:kernel-ratio-linear} from $\abs{x/y-1-(x-y)}=\abs{x-y}\abs{y-1}/y\leq2\abs{x-y}\abs{y-1}\leq(x-y)^2+(y-1)^2$.

We next bound the products of the ratios in \cref{eq:motion-split}.
Fix $(s_0,a_0,\ldots,s_{h-1},a_{h-1})$ such that $(s_{h'},a_{h'},s_{h'+1})\in\calE_t$ for $h'<h$, where $(s_h,a_h)=(s,a)$.
Then
\begin{align}
&\sum_{h'=0}^{h}\abs*{\frac{\pi_t^+(a_{h'}\mid s_{h'})}{\pi_t(a_{h'}\mid s_{h'})}-1}\leq4\sum_{h'=0}^{h}\max_b\abs*{\frac{\hat q_t^+(s_{h'},b)}{\hat q_t(s_{h'},b)}-1}\leq4\sqrt{2SA}\nrm{\hat q_t^+-\hat q_t}_t\leq1,\label{eq:policy-sum}\\
&\sum_{h'=0}^{h-1}\abs*{\frac{\hat P^+_{\calE_t}(s_{h'+1}\mid s_{h'},a_{h'})}{\hat P_{\calE_t}(s_{h'+1}\mid s_{h'},a_{h'})}-1}\leq2\sum_{h'=0}^{h-1}\abs*{\frac{\hat q_t^+(s_{h'},a_{h'},s_{h'+1})}{\hat q_t(s_{h'},a_{h'},s_{h'+1})}-\frac{\hat q_t^+(s_{h'},a_{h'})}{\hat q_t(s_{h'},a_{h'})}}\leq\frac{4\sqrt2}{\sqrt H}\nrm{\hat q_t^+-\hat q_t}_t\leq1,\label{eq:kernel-sum}
\end{align}
where the first inequalities follow from \cref{eq:policy-ratio,eq:kernel-ratio}, the second from \cref{eq:layer-sums,eq:layer-edges} in \cref{lem:ratio} with $u=\hat q_t^+-\hat q_t$, and the last from \cref{eq:step-size} and the definition of $\eta$ in \cref{alg:main}.

By \cref{eq:policy-sum,eq:kernel-sum}, we also have
\begin{align}
\abs*{\prod_{h'=0}^{h}\frac{\pi_t^+(a_{h'}\mid s_{h'})}{\pi_t(a_{h'}\mid s_{h'})}-1}\leq8\sqrt{2SA}\nrm{\hat q_t^+-\hat q_t}_t,\qquad
\abs*{\prod_{h'=0}^{h-1}\frac{\hat P^+_{\calE_t}(s_{h'+1}\mid s_{h'},a_{h'})}{\hat P_{\calE_t}(s_{h'+1}\mid s_{h'},a_{h'})}-1}\leq\frac{8\sqrt2}{\sqrt H}\nrm{\hat q_t^+-\hat q_t}_t,\label{eq:policy-product}
\end{align}
where we used $\abs{\prod_{i=1}^m(1+x_i)-1}\leq r+r^2\leq2r$ for $x_1,\ldots,x_m\in\R$ with $r=\sum_{i=1}^m\abs{x_i}\leq1$, which follows from
\begin{align}\label{eq:product}
\abs*{\prod_{i=1}^m(1+x_i)-1-\sum_{i=1}^mx_i}\leq\prod_{i=1}^m(1+\abs{x_i})-1-r\leq \mathrm{e}^r-1-r\leq r^2,
\end{align}
in which the second inequality follows from $1+z\leq \mathrm{e}^z$ for $z\in\R$, and the last from $\mathrm{e}^r-1-r=\sum_{k\geq2}r^k/k!\leq r^2\sum_{k\geq2}1/k!$ for $0\leq r\leq1$ and $\sum_{k\geq2}1/k!=\mathrm{e}-2\leq1$.

Thus, by \cref{eq:motion-split}, we obtain
\begin{align}
&\abs*{\bar q_t^+(s,a)-\bar q_t(s,a)-\prn*{\hat q_t^+(s,a)-\hat q_t(s,a)}}\\
&\leq\sum_{s_0,a_0,\ldots,s_{h-1},a_{h-1}}\prod_{h'=0}^{h}\pi_t(a_{h'}\mid s_{h'})\prod_{h'=0}^{h-1}\hat P_{\calE_t}(s_{h'+1}\mid s_{h'},a_{h'})\\
&\hspace{6em}\times\prn*{8H\epsilon\cdot8\sqrt{2SA}\nrm{\hat q_t^+-\hat q_t}_t+\frac{8\sqrt2}{\sqrt H}\nrm{\hat q_t^+-\hat q_t}_t\prn*{1+8\sqrt{2SA}\nrm{\hat q_t^+-\hat q_t}_t}}\\
&=\prn*{8H\epsilon\cdot8\sqrt{2SA}+\frac{8\sqrt2}{\sqrt H}\prn*{1+8\sqrt{2SA}\nrm{\hat q_t^+-\hat q_t}_t}}\nrm{\hat q_t^+-\hat q_t}_tq^{\hat P_{\calE_t},\pi_t}(s,a)\\
&\leq\frac{24}{\sqrt H}\nrm{\hat q_t^+-\hat q_t}_t\hat q_t(s,a),\label{eq:motion-pointwise}
\end{align}
where the first inequality follows from \cref{eq:kernel-difference,eq:policy-product}, and the last inequality from $q^{\hat P_{\calE_t},\pi_t}(s,a)\leq\hat q_t(s,a)$ by \cref{lem:effective}, $8\sqrt{2SA}\nrm{\hat q_t^+-\hat q_t}_t\leq1$ by \cref{eq:step-size} and the definition of $\eta$ in \cref{alg:main}, and $64\sqrt2H^{3/2}\epsilon\sqrt{SA}\leq1$ by the definition of $\epsilon$ in \cref{alg:main}.
Hence
\begin{align}
\abs{\bar q_t^+(s,a)-\bar q_t(s,a)}&\leq\prn*{\sqrt{\frac{2SA}{H}}+\frac{24}{\sqrt H}}\nrm{\hat q_t^+-\hat q_t}_t\hat q_t(s,a)\leq255c_0\eta SA\sqrt H\iota\,\hat q_t(s,a)\\
&\leq\frac{\hat q_t(s,a)}8\leq\frac{\bar q_t(s,a)}4,
\end{align}
where the first inequality follows from \cref{eq:relative-step,eq:motion-pointwise}, the second from \cref{eq:step-size}, the third from the definition of $\eta$ in \cref{alg:main}, and the last from $\bar q_t(s,a)\geq\hat q_t(s,a)/2$ by \cref{lem:effective}.
This proves \cref{eq:pointwise-motion}.

We next prove \cref{eq:drift}. Taking $\E_t$ in \cref{eq:motion-split} gives
\begin{align}
&\E_t\brk*{\bar q_t^+(s,a)-\bar q_t(s,a)-\prn*{\hat q_t^+(s,a)-\hat q_t(s,a)}}\\
&=\sum_{s_0,a_0,\ldots,s_{h-1},a_{h-1}}\prod_{h'=0}^{h}\pi_t(a_{h'}\mid s_{h'})\prn*{\prod_{h'=0}^{h-1}P_{\calE_t}(s_{h'+1}\mid s_{h'},a_{h'})-\prod_{h'=0}^{h-1}\hat P_{\calE_t}(s_{h'+1}\mid s_{h'},a_{h'})}\\
&\hspace{8em}\times\E_t\brk*{\prod_{h'=0}^{h}\frac{\pi_t^+(a_{h'}\mid s_{h'})}{\pi_t(a_{h'}\mid s_{h'})}-1}\\
&\quad-\sum_{s_0,a_0,\ldots,s_{h-1},a_{h-1}}\prod_{h'=0}^{h}\pi_t(a_{h'}\mid s_{h'})\prod_{h'=0}^{h-1}\hat P_{\calE_t}(s_{h'+1}\mid s_{h'},a_{h'})\\
&\hspace{8em}\times\E_t\brk*{\prn*{\prod_{h'=0}^{h-1}\frac{\hat P^+_{\calE_t}(s_{h'+1}\mid s_{h'},a_{h'})}{\hat P_{\calE_t}(s_{h'+1}\mid s_{h'},a_{h'})}-1}\prod_{h'=0}^{h}\frac{\pi_t^+(a_{h'}\mid s_{h'})}{\pi_t(a_{h'}\mid s_{h'})}}.\label{eq:mean-split}
\end{align}
We bound the two conditional means on the right-hand side by approximating the products of the ratios by linear functions of $\hat q_t^+$, whose conditional means depend only on $\E_t[\hat q_t^+]$.
For the product of the ratios of the policies, 
\begin{align}
&\abs*{\prod_{h'=0}^{h}\frac{\pi_t^+(a_{h'}\mid s_{h'})}{\pi_t(a_{h'}\mid s_{h'})}-1-\sum_{h'=0}^{h}\prn*{\frac{\hat q_t^+(s_{h'},a_{h'})}{\hat q_t(s_{h'},a_{h'})}-\frac{\sum_b\hat q_t^+(s_{h'},b)}{\sum_b\hat q_t(s_{h'},b)}}}\\
&\leq\abs*{\prod_{h'=0}^{h}\frac{\pi_t^+(a_{h'}\mid s_{h'})}{\pi_t(a_{h'}\mid s_{h'})}-1-\sum_{h'=0}^{h}\prn*{\frac{\pi_t^+(a_{h'}\mid s_{h'})}{\pi_t(a_{h'}\mid s_{h'})}-1}}\\
&\quad+\sum_{h'=0}^{h}\abs*{\frac{\pi_t^+(a_{h'}\mid s_{h'})}{\pi_t(a_{h'}\mid s_{h'})}-1-\prn*{\frac{\hat q_t^+(s_{h'},a_{h'})}{\hat q_t(s_{h'},a_{h'})}-\frac{\sum_b\hat q_t^+(s_{h'},b)}{\sum_b\hat q_t(s_{h'},b)}}}\\
&\leq\prn*{\sum_{h'=0}^{h}\abs*{\frac{\pi_t^+(a_{h'}\mid s_{h'})}{\pi_t(a_{h'}\mid s_{h'})}-1}}^2+4\sum_{h'=0}^{h}\max_b\prn*{\frac{\hat q_t^+(s_{h'},b)}{\hat q_t(s_{h'},b)}-1}^2\\
&\leq32SA\nrm{\hat q_t^+-\hat q_t}_t^2+4\sum_{s,a}\prn*{\frac{\hat q_t^+(s,a)}{\hat q_t(s,a)}-1}^2\lesssim SA\nrm{\hat q_t^+-\hat q_t}_t^2,\label{eq:policy-linear}
\end{align}
where the first inequality follows from the triangle inequality, the second from \cref{eq:product} with \cref{eq:policy-sum} and from \cref{eq:policy-ratio-linear}, the third from \cref{eq:policy-sum}, and the last from \cref{eq:relative-coordinates} in \cref{lem:ratio} with $u=\hat q_t^+-\hat q_t$.
Similarly, for the product of the ratios of the estimated kernels,
\begin{align}
&\abs*{\prod_{h'=0}^{h-1}\frac{\hat P^+_{\calE_t}(s_{h'+1}\mid s_{h'},a_{h'})}{\hat P_{\calE_t}(s_{h'+1}\mid s_{h'},a_{h'})}-1-\sum_{h'=0}^{h-1}\prn*{\frac{\hat q_t^+(s_{h'},a_{h'},s_{h'+1})}{\hat q_t(s_{h'},a_{h'},s_{h'+1})}-\frac{\hat q_t^+(s_{h'},a_{h'})}{\hat q_t(s_{h'},a_{h'})}}}\\
&\leq\prn*{\sum_{h'=0}^{h-1}\abs*{\frac{\hat P^+_{\calE_t}(s_{h'+1}\mid s_{h'},a_{h'})}{\hat P_{\calE_t}(s_{h'+1}\mid s_{h'},a_{h'})}-1}}^2\\
&\quad+\sum_{h'=0}^{h-1}\prn*{\prn*{\frac{\hat q_t^+(s_{h'},a_{h'},s_{h'+1})}{\hat q_t(s_{h'},a_{h'},s_{h'+1})}-\frac{\hat q_t^+(s_{h'},a_{h'})}{\hat q_t(s_{h'},a_{h'})}}^2+\prn*{\frac{\hat q_t^+(s_{h'},a_{h'})}{\hat q_t(s_{h'},a_{h'})}-1}^2}\\
&\leq\frac{32}H\nrm{\hat q_t^+-\hat q_t}_t^2+\sum_{(s,a,s')\in\calE_t}\prn*{\frac{\hat q_t^+(s,a,s')}{\hat q_t(s,a,s')}-\frac{\hat q_t^+(s,a)}{\hat q_t(s,a)}}^2+\sum_{s,a}\prn*{\frac{\hat q_t^+(s,a)}{\hat q_t(s,a)}-1}^2\lesssim SA\nrm{\hat q_t^+-\hat q_t}_t^2,\label{eq:kernel-linear}
\end{align}
where the first inequality follows from the triangle inequality, \cref{eq:product} with \cref{eq:kernel-sum}, and \cref{eq:kernel-ratio-linear}, the second from \cref{eq:kernel-sum}, and the last from \cref{eq:relative-coordinates,eq:relative-edges} in \cref{lem:ratio} with $u=\hat q_t^+-\hat q_t$.
For the second term of \cref{eq:motion-split}, we also obtain
\begin{align}
&\prn*{\prod_{h'=0}^{h-1}\frac{\hat P^+_{\calE_t}(s_{h'+1}\mid s_{h'},a_{h'})}{\hat P_{\calE_t}(s_{h'+1}\mid s_{h'},a_{h'})}-1}\prod_{h'=0}^{h}\frac{\pi_t^+(a_{h'}\mid s_{h'})}{\pi_t(a_{h'}\mid s_{h'})}\\
&=\prod_{h'=0}^{h-1}\frac{\hat P^+_{\calE_t}(s_{h'+1}\mid s_{h'},a_{h'})}{\hat P_{\calE_t}(s_{h'+1}\mid s_{h'},a_{h'})}-1+\prn*{\prod_{h'=0}^{h-1}\frac{\hat P^+_{\calE_t}(s_{h'+1}\mid s_{h'},a_{h'})}{\hat P_{\calE_t}(s_{h'+1}\mid s_{h'},a_{h'})}-1}\prn*{\prod_{h'=0}^{h}\frac{\pi_t^+(a_{h'}\mid s_{h'})}{\pi_t(a_{h'}\mid s_{h'})}-1}\\
&=\prod_{h'=0}^{h-1}\frac{\hat P^+_{\calE_t}(s_{h'+1}\mid s_{h'},a_{h'})}{\hat P_{\calE_t}(s_{h'+1}\mid s_{h'},a_{h'})}-1+O\prn*{SA\nrm{\hat q_t^+-\hat q_t}_t^2},\label{eq:cross-term}
\end{align}
where the last equality follows from \cref{eq:policy-product}.
Here, by \cref{eq:layer-sums,eq:layer-edges} in \cref{lem:ratio} with $u=\E_t[\hat q_t^+-\hat q_t]$, the linear parts in \cref{eq:policy-linear,eq:kernel-linear} satisfy
\begin{align}
\abs*{\E_t\brk*{\sum_{h'=0}^{h}\prn*{\frac{\hat q_t^+(s_{h'},a_{h'})}{\hat q_t(s_{h'},a_{h'})}-\frac{\sum_b\hat q_t^+(s_{h'},b)}{\sum_b\hat q_t(s_{h'},b)}}}}&\leq2\sqrt{2SA}\nrm{\E_t[\hat q_t^+-\hat q_t]}_t,\label{eq:linear-mean1}\\
\abs*{\E_t\brk*{\sum_{h'=0}^{h-1}\prn*{\frac{\hat q_t^+(s_{h'},a_{h'},s_{h'+1})}{\hat q_t(s_{h'},a_{h'},s_{h'+1})}-\frac{\hat q_t^+(s_{h'},a_{h'})}{\hat q_t(s_{h'},a_{h'})}}}}&\leq\frac{2\sqrt2}{\sqrt H}\nrm{\E_t[\hat q_t^+-\hat q_t]}_t.\label{eq:linear-mean2}
\end{align}
Hence, by \cref{eq:mean-split},
\begin{align}
&\abs*{\E_t[\bar q_t^+(s,a)-\bar q_t(s,a)]-\E_t[\hat q_t^+(s,a)-\hat q_t(s,a)]}\\
&\lesssim\sum_{s_0,a_0,\ldots,s_{h-1},a_{h-1}}\prod_{h'=0}^{h}\pi_t(a_{h'}\mid s_{h'})\prod_{h'=0}^{h-1}\hat P_{\calE_t}(s_{h'+1}\mid s_{h'},a_{h'})\\
&\quad\times\prn*{H\epsilon\prn*{\sqrt{SA}\nrm{\E_t[\hat q_t^+-\hat q_t]}_t+SA\,\E_t\brk*{\nrm{\hat q_t^+-\hat q_t}_t^2}}+\frac{\nrm{\E_t[\hat q_t^+-\hat q_t]}_t}{\sqrt H}+SA\,\E_t\brk*{\nrm{\hat q_t^+-\hat q_t}_t^2}}\\
&=\prn*{\prn*{H\epsilon\sqrt{SA}+\frac1{\sqrt H}}\nrm{\E_t[\hat q_t^+-\hat q_t]}_t+(H\epsilon+1)SA\,\E_t\brk*{\nrm{\hat q_t^+-\hat q_t}_t^2}} q^{\hat P_{\calE_t},\pi_t}(s,a)\\
&\lesssim\prn*{\frac{\nrm{\E_t[\hat q_t^+-\hat q_t]}_t}{\sqrt H}+SA\,\E_t\brk*{\nrm{\hat q_t^+-\hat q_t}_t^2}}\hat q_t(s,a),\label{eq:motion-mean}
\end{align}
where the first inequality follows from \cref{eq:kernel-difference,eq:policy-linear,eq:kernel-linear,eq:linear-mean1,eq:linear-mean2,eq:cross-term}, and the last inequality from $q^{\hat P_{\calE_t},\pi_t}(s,a)\leq\hat q_t(s,a)$ by \cref{lem:effective} and the definition of $\epsilon$ in \cref{alg:main}.
Thus
\begin{align}
&\sum_{s,a}\frac{\abs{\E_t[\bar q_t^+(s,a)-\bar q_t(s,a)]}}{\hat q_t(s,a)}\\
&\leq\sum_{s,a}\frac{\abs{\E_t[\hat q_t^+(s,a)-\hat q_t(s,a)]}}{\hat q_t(s,a)}+\sum_{s,a}\frac{\abs{\E_t[\bar q_t^+(s,a)-\bar q_t(s,a)]-\E_t[\hat q_t^+(s,a)-\hat q_t(s,a)]}}{\hat q_t(s,a)}\\
&\lesssim\sqrt{SA\sum_{s,a}\prn*{\frac{\E_t[\hat q_t^+(s,a)-\hat q_t(s,a)]}{\hat q_t(s,a)}}^2}+SA\prn*{\frac{\nrm{\E_t[\hat q_t^+-\hat q_t]}_t}{\sqrt H}+SA\,\E_t\brk*{\nrm{\hat q_t^+-\hat q_t}_t^2}}\\
&\lesssim\frac{SA}{\sqrt H}\nrm{\E_t[\hat q_t^+-\hat q_t]}_t+S^2A^2\E_t\brk*{\nrm{\hat q_t^+-\hat q_t}_t^2}\\
&\lesssim\frac{SA}{\sqrt H}\prn*{\eta\nrm{\E_t[\hat g_t]}_{t,*}+\eta^2\sqrt{\frac{S^2A}{H}}\E_t\brk*{\nrm{\hat g_t}_{t,*}^2}}+\eta^2S^2A^2\E_t\brk*{\nrm{\hat g_t}_{t,*}^2}\\
&\lesssim\eta SA\iota+\eta^2S^3A^{5/2}\iota^2+\eta^2HS^3A^3\iota^2\lesssim\eta SA\iota+\eta^2HS^3A^3\iota^2,
\end{align}
where the first inequality follows from the triangle inequality, the second from the Cauchy--Schwarz inequality and \cref{eq:motion-mean}, the third from \cref{eq:relative-coordinates} in \cref{lem:ratio}, the fourth from \cref{lem:update} and \cref{eq:step-size}, and the fifth from \cref{lem:gradient-moments}.
This proves \cref{eq:drift}.

Finally, we prove \cref{eq:qv}. We have
\begin{align}
&\E_t\brk*{\sum_{s,a}\prn*{\frac{\bar q_t^+(s,a)-\bar q_t(s,a)}{\hat q_t(s,a)}}^2}\\
&\leq2\E_t\brk*{\sum_{s,a}\prn*{\frac{\hat q_t^+(s,a)-\hat q_t(s,a)}{\hat q_t(s,a)}}^2}+2\E_t\brk*{\sum_{s,a}\prn*{\frac{\bar q_t^+(s,a)-\bar q_t(s,a)-(\hat q_t^+(s,a)-\hat q_t(s,a))}{\hat q_t(s,a)}}^2}\\
&\lesssim \E_t\brk*{\sum_{s,a}\prn*{\frac{\hat q_t^+(s,a)-\hat q_t(s,a)}{\hat q_t(s,a)}}^2}+\frac{SA}{H}\E_t\brk*{\nrm{\hat q_t^+-\hat q_t}_t^2}\\
&\lesssim\frac{SA}{H}\E_t\brk*{\nrm{\hat q_t^+-\hat q_t}_t^2}\lesssim\frac{SA}{H}\eta^2\E_t\brk*{\nrm{\hat g_t}_{t,*}^2}\lesssim\eta^2S^2A^2\iota^2,
\end{align}
where the first inequality follows from $(x+y)^2\leq2x^2+2y^2$, the second from \cref{eq:motion-pointwise}, the third from \cref{eq:relative-coordinates} in \cref{lem:ratio} with $u=\hat q_t^+-\hat q_t$, the fourth from \cref{eq:step-size}, and the last from \cref{lem:gradient-moments}.
This completes the proof.
\end{proof}

\begin{lemma}\label{lem:policy-motion}
For $t<\tau$,
$\E_t\brk*{\log R(\pi_t^+,\pi_t)}\lesssim\eta\sqrt HSA\iota$.
\end{lemma}
\begin{proof}
By the definition of $R$ in \cref{eq:epoch}, we have
\begin{align}
\log R(\pi_t^+,\pi_t)&=\sum_{h<H}\max_{(s,a)\in\calS_h\times\calA}\abs*{\log\frac{\pi_t^+(a\mid s)}{\pi_t(a\mid s)}}\leq2\sum_{h<H}\max_{(s,a)\in\calS_h\times\calA}\abs*{\frac{\pi_t^+(a\mid s)}{\pi_t(a\mid s)}-1}\\
&\leq8\sum_{h<H}\max_{(s,a)\in\calS_h\times\calA}\abs*{\frac{\hat q_t^+(s,a)}{\hat q_t(s,a)}-1}\leq8\sqrt{2SA}\nrm{\hat q_t^+-\hat q_t}_t\leq16\sqrt{2SA}\,\eta\nrm{\hat g_t}_{t,*},\label{eq:log-ratio}
\end{align}
where the first inequality follows from $\abs{\log(1+z)}\leq2\abs z$ for $\abs z\leq\frac12$ since $\abs{\pi_t^+(a\mid s)/\pi_t(a\mid s)-1}\leq4\cdot\frac18=\frac12$ by \cref{eq:policy-ratio,eq:relative-step}, the second from \cref{eq:policy-ratio}, the third from \cref{eq:layer-sums} in \cref{lem:ratio} with $u=\hat q_t^+-\hat q_t$, and the last from \cref{eq:step-size}.
Hence
\begin{align}
\E_t\brk*{\log R(\pi_t^+,\pi_t)}\leq16\sqrt{2SA}\,\eta\,\E_t\brk*{\nrm{\hat g_t}_{t,*}}\leq16\sqrt{2SA}\,\eta\sqrt{\E_t\brk*{\nrm{\hat g_t}_{t,*}^2}}\lesssim\eta\sqrt HSA\iota,
\end{align}
where the second inequality follows from the Jensen inequality, and the last from \cref{lem:gradient-moments}.
\end{proof}

\begin{lemma}\label{lem:delayed-estimate}
For $t<\tau$ and every $(s,a)$, $\E_t[\hat\ell_t(s,a)]\leq\ell_t(s,a)$.
\end{lemma}
\begin{proof}
By the update of $w_{t,u}(s,a)$,
\begin{align}\label{eq:weight-identity}
\I_t(s,a;\calE_t)-w_{t,\tau}(s,a)=\sum_{u=t+1}^{\tau-1}\prn*{w_{t,u}(s,a)-w_{t,u+1}(s,a)}=\sum_{u=t+1}^{\tau-1}\frac{w_{t,u}(s,a)\prn*{\tilde q_{t,u}(s,a)+\gamma}}{c_0\iota}.
\end{align}
Hence
\begin{align}
q^{P_{\calE_t},\pi_t}(s,a)\E_t\brk*{\sum_{u=t+1}^{\tau-1}\frac{w_{t,u}(s,a)}{c_0\iota}}
&\leq\E_t\brk*{\sum_{u=t+1}^{\tau-1}\frac{w_{t,u}(s,a)\prn*{\tilde q_{t,u}(s,a)+\gamma}}{c_0\iota}}\\
&=\E_t\brk*{\I_t(s,a;\calE_t)-w_{t,\tau}(s,a)}\\
&\leq q^{P_{\calE_t},\pi_t}(s,a),
\end{align}
where the first inequality follows from $\E_u[\tilde q_{t,u}(s,a)]\geq q^{P_{\calE_t},\pi_t}(s,a)$ in \cref{lem:service}, and the last from $w_{t,\tau}(s,a)\geq0$ and $\E_t[\I_t(s,a;\calE_t)]=q^{P_{\calE_t},\pi_t}(s,a)$.
If $q^{P_{\calE_t},\pi_t}(s,a)>0$, dividing both sides by $q^{P_{\calE_t},\pi_t}(s,a)$ gives
\begin{align}
\E_t[\hat\ell_t(s,a)]=\ell_t(s,a)\E_t\brk*{\sum_{u=t+1}^{\tau-1}\frac{w_{t,u}(s,a)}{c_0\iota}}\leq\ell_t(s,a).
\end{align}
If $q^{P_{\calE_t},\pi_t}(s,a)=0$, then $\I_t(s,a;\calE_t)=0$ almost surely, and hence $\hat\ell_t(s,a)=0$.
\end{proof}

\section{Proof of \texorpdfstring{\cref{thm:main}}{Theorem 5.1}}\label{app:terms}
This section reduces the regret to five terms (\cref{lem:decomposition}), bounds them with high probability, and proves \cref{thm:main}.

\subsection{Regret decomposition}\label{app:decomposition}
\begin{lemma}\label{lem:decomposition}
Suppose that $\tau=T+1$ holds. Then $\Reg_T=\text{(I)}+\text{(II)}+\text{(III)}+\text{(IV)}+\text{(V)}$, where
\begin{align}
&\text{(I)}=\sum_{t<\tau}\inpr*{\bar q_t-q^{\pio},G_t},
&&\text{(II)}=\sum_{t<\tau}\prn*{\inpr*{q^{P_{\calE_t},\pi_t},\hat\ell_t}-\inpr*{\bar q_t,G_t}},
&&\text{(III)}=\sum_{t<\tau}\inpr*{q^{P_{\calE_t},\pi_t},\ell_t-\hat\ell_t},\\
&\text{(IV)}=\sum_{t<\tau}\inpr*{q^{\pio},\hat\ell_t-\ell_t},
&&\text{(V)}=\sum_{t<\tau}\inpr*{q^{\pi_t}-q^{P_{\calE_t},\pi_t},\ell_t}.
\end{align}
\end{lemma}
\begin{proof}
By $V^{\pi}(s_0;\ell_t)=\inpr{q^{\pi},\ell_t}$ and $\tau=T+1$, we have
\begin{align}
\Reg_T=\sum_{t=1}^T\inpr*{q^{\pi_t}-q^{\pio},\ell_t}=\sum_{t<\tau}\inpr*{q^{P_{\calE_t},\pi_t}-q^{\pio},\ell_t}+\underbrace{\sum_{t<\tau}\inpr*{q^{\pi_t}-q^{P_{\calE_t},\pi_t},\ell_t}}_{\text{(V)}}.
\end{align}
For the first term on the right-hand side, we have
\begin{align}
\sum_{t<\tau}\inpr*{q^{P_{\calE_t},\pi_t}-q^{\pio},\ell_t}
&=\sum_{t<\tau}\inpr*{q^{P_{\calE_t},\pi_t}-q^{\pio},\hat\ell_t}+\underbrace{\sum_{t<\tau}\inpr*{q^{P_{\calE_t},\pi_t},\ell_t-\hat\ell_t}}_{\text{(III)}}+\underbrace{\sum_{t<\tau}\inpr*{q^{\pio},\hat\ell_t-\ell_t}}_{\text{(IV)}}.\label{eq:decomp2-rhs}
\end{align}

For the first term in \cref{eq:decomp2-rhs}, \cref{eq:source-estimate,eq:received} give $\sum_{t<\tau}\inpr{q^{\pio},\hat\ell_t}=\sum_{t<\tau}\inpr{q^{\pio},G_t}$ by exchanging the order of summation, and hence
\begin{align}
\sum_{t<\tau}\inpr*{q^{P_{\calE_t},\pi_t}-q^{\pio},\hat\ell_t}
=\underbrace{\sum_{t<\tau}\inpr*{\bar q_t-q^{\pio},G_t}}_{\text{(I)}}+\underbrace{\sum_{t<\tau}\prn*{\inpr*{q^{P_{\calE_t},\pi_t},\hat\ell_t}-\inpr*{\bar q_t,G_t}}}_{\text{(II)}}.
\end{align}
\end{proof}

\subsection{Bounding term (I)}\label{app:ftrl}
\begin{proof}[Proof of \cref{lem:pairing}]
By the definition of $\hat g_t$ in \cref{eq:gradient}, $1/(\hat q_t(s,a)+\gamma)=(1-\gamma/(\hat q_t(s,a)+\gamma))/\hat q_t(s,a)$, \cref{eq:ix-term}, and $\hat q_t(s,a,s')=\hat q_t(s,a)\hat P_{\calE_t}(s'\mid s,a)$ for $(s,a,s')\in\calE_t$,
\begin{align}
\inpr{\hat q_t,\hat g_t}-\inpr{\hat q_t,G_t}+\sum_{s,a}\beta_t(s,a)
=\sum_{s,a}\sum_{s'}\prn*{\I_t(s,a,s';\calE_t)-\I_t(s,a;\calE_t)\hat P_{\calE_t}(s'\mid s,a)}V^{\hat P_{\calE_t},\pi_t}(s';G_t).
\end{align}
Taking expectation and using $\E_t[\I_t(s,a,s';\calE_t)]=q^{P_{\calE_t},\pi_t}(s,a)P_{\calE_t}(s'\mid s,a)$ and $\E_t[\I_t(s,a;\calE_t)]=q^{P_{\calE_t},\pi_t}(s,a)$,
\begin{align}
&\E_t\brk*{\inpr{\hat q_t,\hat g_t}}-\inpr{\hat q_t,G_t}+\sum_{s,a}\E_t[\beta_t(s,a)]\\
&=\sum_{s,a}q^{P_{\calE_t},\pi_t}(s,a)\sum_{s'}\prn*{P_{\calE_t}(s'\mid s,a)-\hat P_{\calE_t}(s'\mid s,a)}V^{\hat P_{\calE_t},\pi_t}(s';G_t)\\
&=\sum_{s\neq s_0}q^{P_{\calE_t},\pi_t}(s)V^{\hat P_{\calE_t},\pi_t}(s;G_t)-\sum_{s,a}q^{P_{\calE_t},\pi_t}(s,a)\prn*{Q^{\hat P_{\calE_t},\pi_t}(s,a;G_t)-G_t(s,a)}\\
&=\sum_{s\neq s_0}q^{P_{\calE_t},\pi_t}(s)V^{\hat P_{\calE_t},\pi_t}(s;G_t)-\sum_{s}q^{P_{\calE_t},\pi_t}(s)V^{\hat P_{\calE_t},\pi_t}(s;G_t)+\inpr{q^{P_{\calE_t},\pi_t},G_t}\\
&=\inpr{q^{P_{\calE_t},\pi_t}-q^{\hat P_{\calE_t},\pi_t},G_t},
\end{align}
where the second equality follows from the definitions of $q^{P_{\calE_t},\pi_t}(s)$ and $Q^{\hat P_{\calE_t},\pi_t}$, the third from $q^{P_{\calE_t},\pi_t}(s,a)=q^{P_{\calE_t},\pi_t}(s)\pi_t(a\mid s)$, and the last from $q^{P_{\calE_t},\pi_t}(s_0)=1$ and $V^{\hat P_{\calE_t},\pi_t}(s_0;G_t)=\inpr{q^{\hat P_{\calE_t},\pi_t},G_t}$.
\end{proof}

\begin{lemma}\label{lem:ftrl}
Term (I) in \cref{lem:decomposition} satisfies, with probability at least $1-3\delta$,
\begin{align}
\text{(I)}=\sum_{t<\tau}\inpr*{\bar q_t-q^{\pio},G_t}\lesssim\frac{H\iota}{\eta}+\eta HSAT\iota^2+H\sqrt{SAT}\iota^2+HSA\iota^2.
\end{align}
\end{lemma}
\begin{proof}
With $\qo$ as in \cref{lem:ftrl-regret}, we decompose
\begin{align}
\text{(I)}
&=\sum_{t<\tau}\inpr{\bar q_t-q^{\pio},G_t}\\
&=\sum_{t<\tau}\inpr{\hat q_t-q^{\pio},\hat g_t}+\sum_{t<\tau}\prn*{\inpr{\bar q_t,G_t}-\inpr{\hat q_t,\hat g_t}}+\sum_{t<\tau}\inpr{q^{\pio},\hat g_t-G_t}\\
&\lesssim\frac{H\iota}{\eta}+\eta\underbrace{\sum_{t<\tau}\nrm{\hat g_t}_{t,*}^2}_{\text{(Ia)}}+\underbrace{\sum_{t<\tau}\prn*{\inpr{\bar q_t,G_t}-\inpr{\hat q_t,\hat g_t}}}_{\text{(Ib)}}+\underbrace{\sum_{t<\tau}\inpr{\qo,\hat g_t-G_t}}_{\text{(Ic)}}+\underbrace{\sum_{t<\tau}\inpr{\qo-q^{\pio},G_t}}_{\text{(Id)}},\label{eq:term-I}
\end{align}
where the inequality follows from \cref{lem:ftrl-regret}, whose assumption holds by \cref{lem:local-ftrl}.
We bound (Ia)--(Id) in turn.

For (Ia), since $\nrm{\hat g_t}_{t,*}^2\leq25c_0^2H^2SA\iota^2$ by \cref{lem:local-ftrl}, \cref{lem:bounded-sum} and \cref{lem:gradient-moments} give, with probability at least $1-\delta$,
\begin{align}
\text{(Ia)}=\sum_{t<\tau}\nrm{\hat g_t}_{t,*}^2\leq\frac32\sum_{t<\tau}\E_t\brk*{\nrm{\hat g_t}_{t,*}^2}+50c_0^2H^2SA\iota^2\log\frac1\delta\lesssim HSAT\iota^2+H^2SA\iota^3.
\end{align}
By the definition of $\eta$ in \cref{alg:main} and $H\leq S$, we have $\eta H^2SA\iota^3\lesssim H^{3/2}\iota^2/\sqrt S\leq H\iota^2$, and hence
\begin{align}
    \eta\cdot\text{(Ia)}\lesssim\eta HSAT\iota^2+H\iota^2.
\end{align}

For (Ib), let $Z_t=\sum_{s,a}(\mathrm{TD}_t(s,a)-\beta_t(s,a))$.
Then
\begin{align}
\text{(Ib)}
&=\sum_{t<\tau}\prn*{\inpr{\bar q_t,G_t}-\inpr{\hat q_t,G_t}-Z_t}\\
&=\sum_{t<\tau}\prn*{\E_t\brk*{\inpr{\hat q_t,\hat g_t}}-\inpr{\hat q_t,G_t}-Z_t}+\sum_{t<\tau}\sum_{s,a}\E_t[\beta_t(s,a)]\\
&=\sum_{t<\tau}\prn*{\E_t[Z_t]-Z_t}+\sum_{t<\tau}\sum_{s,a}\E_t[\beta_t(s,a)],\label{eq:term-Ib}
\end{align}
where the first and last equalities follow from $\inpr{\hat q_t,\hat g_t}=\inpr{\hat q_t,G_t}+Z_t$, which is \cref{eq:gradient-pairing} with $u=\hat q_t$, and the second from \cref{lem:pairing} and the definition of $\bar q_t$.

For the first sum in \cref{eq:term-Ib}, \cref{eq:td-range} and $\sum_{s,a}\I_t(s,a;\calE_t)\leq H$ give, for $t<\tau$,
\begin{align}
\abs{Z_t}\leq c_0H^2\iota,\qquad\E_t[Z_t^2]\leq H\,\E_t\brk*{\sum_{s,a}\prn*{\mathrm{TD}_t(s,a)-\beta_t(s,a)}^2}\leq11c_0^2H^3\iota^2,
\end{align}
where the second inequality follows from the Cauchy--Schwarz inequality, and the third from \cref{eq:td-h-variance}.
Thus \cref{lem:martingale} with $R=2c_0H^2\iota$ and $V=11c_0^2H^3\iota^2T$ gives, with probability at least $1-\delta$,
\begin{align}
\sum_{t<\tau}\prn*{\E_t[Z_t]-Z_t}\leq\sqrt{22c_0^2H^3\iota^2T\log(1/\delta)}+\frac43c_0H^2\iota\log(1/\delta)\lesssim H^{3/2}\sqrt T \iota^{3/2}+H^2\iota^2,
\end{align}
where the last inequality follows from $\log(1/\delta)\leq \iota$.

For the second sum in \cref{eq:term-Ib},
\begin{align}
\sum_{t<\tau}\sum_{s,a}\E_t[\beta_t(s,a)]
&\leq\sum_{t<\tau}\sum_{s,a}\frac{c_0\gamma H\iota\,\E_t[\I_t(s,a;\calE_t)]}{\hat q_t(s,a)+\gamma}
\leq\sum_{t<\tau}\sum_{s,a}\frac{2c_0\gamma H\iota\,\hat q_t(s,a)}{\hat q_t(s,a)+\gamma}\\
&\leq2c_0\gamma HSAT\iota\leq2c_0H\sqrt{SAT}\iota,
\end{align}
where the first inequality follows from \cref{eq:h-bound}, the second from $\E_t[\I_t(s,a;\calE_t)]=q^{P_{\calE_t},\pi_t}(s,a)\leq2\hat q_t(s,a)$ by \cref{eq:indicator-mean} and \cref{lem:effective}, and the last from $\gamma\leq1/\sqrt{SAT}$.
Then, with probability at least $1-\delta$,
\begin{align}
\text{(Ib)}\lesssim H\sqrt{SAT}\iota^{3/2}+H^2\iota^2.
\end{align}

For (Ic), fix $t<\tau$.
By \cref{eq:gradient} and $\qo(s,a,s')=\qo(s,a)P^\zeta(s'\mid s,a)$,
\begin{align}
&\inpr{\qo,\hat g_t-G_t}\\
&=\sum_{s,a}\qo(s,a)\prn*{\frac{\sum_{s'}\I_t(s,a,s';\calE_t)V^{\hat P_{\calE_t},\pi_t}(s';G_t)}{\hat q_t(s,a)+\gamma}-\frac{\I_t(s,a;\calE_t)\sum_{s'}P^\zeta_{\calE_t}(s'\mid s,a)V^{\hat P_{\calE_t},\pi_t}(s';G_t)}{\hat q_t(s,a)}}\\
&\leq\sum_{s,a}\qo(s,a)\prn*{\frac{\sum_{s'}\I_t(s,a,s';\calE_t)V^{\hat P_{\calE_t},\pi_t}(s';G_t)}{\hat q_t(s,a)+\gamma}-\frac{(1-\zeta)\I_t(s,a;\calE_t)\sum_{s'}P_{\calE_t}(s'\mid s,a)V^{\hat P_{\calE_t},\pi_t}(s';G_t)}{\hat q_t(s,a)}},\label{eq:term-Ic}
\end{align}
where the inequality follows from $P^\zeta(s'\mid s,a)\geq(1-\zeta)P(s'\mid s,a)$ and $V^{\hat P_{\calE_t},\pi_t}(\cdot;G_t)\geq0$.
Let $\lambda=\gamma/(2c_0H\iota)$. Then, for each $(s,a)$,
{\small\begin{align}
&\E_t\Bigg[\exp\Bigg(\lambda\qo(s,a)\Bigg(\frac{\sum_{s'}\I_t(s,a,s';\calE_t)V^{\hat P_{\calE_t},\pi_t}(s';G_t)}{\hat q_t(s,a)+\gamma}\\
&\hspace{8em}-\frac{(1-\zeta)\sum_{s'}P_{\calE_t}(s'\mid s,a)V^{\hat P_{\calE_t},\pi_t}(s';G_t)}{\hat q_t(s,a)}\Bigg)\Bigg)\Biggm|\I_t(s,a;\calE_t)=1\Bigg]\\
&\leq\prn*{1+\frac{\lambda\qo(s,a)\sum_{s'}P_{\calE_t}(s'\mid s,a)V^{\hat P_{\calE_t},\pi_t}(s';G_t)}{\hat q_t(s,a)+\gamma}\prn*{1+\frac{\gamma}{2(\hat q_t(s,a)+\gamma)}}}\\
&\qquad\times\exp\prn*{-\frac{\lambda\qo(s,a)(1-\zeta)\sum_{s'}P_{\calE_t}(s'\mid s,a)V^{\hat P_{\calE_t},\pi_t}(s';G_t)}{\hat q_t(s,a)}}\\
&\leq\exp\Bigg(\lambda\qo(s,a)\sum_{s'}P_{\calE_t}(s'\mid s,a)V^{\hat P_{\calE_t},\pi_t}(s';G_t)\prn*{\frac{1}{\hat q_t(s,a)+\gamma}+\frac{\gamma}{2(\hat q_t(s,a)+\gamma)^2}-\frac{1-\zeta}{\hat q_t(s,a)}}\Bigg)\leq1,\label{eq:ix-step}
\end{align}}
where the first inequality follows from $\mathrm{e}^x\leq1+x+x^2$ for $x\leq1$ and $x^2\leq\gamma x/(2(\hat q_t(s,a)+\gamma))$ for $x=\lambda\qo(s,a)\sum_{s'}\I_t(s,a,s';\calE_t)V^{\hat P_{\calE_t},\pi_t}(s';G_t)/(\hat q_t(s,a)+\gamma)$, which lies in $[0,\gamma/(2(\hat q_t(s,a)+\gamma))]\subseteq[0,1/2]$ by \cref{lem:bounded-loss} and $\qo(s,a)\leq1$, the second from $1+y\leq \mathrm{e}^y$, and the last from
\begin{align}
\frac{1}{\hat q_t(s,a)+\gamma}+\frac{\gamma}{2(\hat q_t(s,a)+\gamma)^2}-\frac{1-\zeta}{\hat q_t(s,a)}
&\leq\frac{\hat q_t(s,a)+\gamma/2}{\hat q_t(s,a)(\hat q_t(s,a)+\gamma)}-\frac{1-\zeta}{\hat q_t(s,a)}\\
&=\frac{\zeta(\hat q_t(s,a)+\gamma)-\gamma/2}{\hat q_t(s,a)(\hat q_t(s,a)+\gamma)}\leq0.
\end{align}
Here the second inequality follows from $\zeta(\hat q_t(s,a)+\gamma)\leq2\zeta\leq\gamma/2$ with $\zeta=\epsilon/(2T)\leq\gamma/4$ by \cref{alg:main}.

Since $\I_t(s,a;\calE_t)=1$ for at most one $(s,a)$ in each layer, taking conditional expectations in \cref{eq:term-Ic} from the last layer backward and applying \cref{eq:ix-step} gives $\E_t[\exp(\lambda\inpr{\qo,\hat g_t-G_t})]\leq1$.
Thus \cref{lem:supermartingale} gives, with probability at least $1-\delta$,
\begin{align}
\text{(Ic)}\leq\frac{2c_0H\iota}{\gamma}\log\frac1\delta\lesssim H\sqrt{SAT}\iota^2+HSA\iota^2.
\end{align}

For (Id), we bound $\sum_{s,a}\abs{\qo(s,a)-q^{\pio}(s,a)}$ layer by layer.
For $(s',a')\in\calS_{h+1}\times\calA$, the recursion of the occupancy measures gives
\begin{align}
\qo(s',a')-q^{\pio}(s',a')
&=\pi^\zeta(a'\mid s')\sum_{(s,a)\in\calS_h\times\calA}\prn*{\qo(s,a)-q^{\pio}(s,a)}P^\zeta(s'\mid s,a)\\
&\quad+\pi^\zeta(a'\mid s')\sum_{(s,a)\in\calS_h\times\calA}q^{\pio}(s,a)\prn*{P^\zeta(s'\mid s,a)-P(s'\mid s,a)}\\
&\quad+\prn*{\pi^\zeta(a'\mid s')-\pio(a'\mid s')}q^{\pio}(s').
\end{align}
Then, using $\sum_{s'}\abs{P^\zeta(s'\mid s,a)-P(s'\mid s,a)}\leq2\zeta$, $\sum_{a'}\abs{\pi^\zeta(a'\mid s')-\pio(a'\mid s')}\leq2\zeta$, and $\sum_{(s,a)\in\calS_h\times\calA}q^{\pio}(s,a)\leq1$,
\begin{align}
\sum_{(s,a)\in\calS_{h+1}\times\calA}\abs*{\qo(s,a)-q^{\pio}(s,a)}\leq\sum_{(s,a)\in\calS_h\times\calA}\abs*{\qo(s,a)-q^{\pio}(s,a)}+4\zeta,
\end{align}
and the sum at $h=0$ is at most $2\zeta$.
Hence $\sum_{s,a}\abs{\qo(s,a)-q^{\pio}(s,a)}\leq\sum_{h<H}(4h+2)\zeta=2H^2\zeta$.
This and $G_t\leq c_0\iota$ from \cref{lem:bounded-loss} give $\inpr{\qo-q^{\pio},G_t}\leq2c_0\zeta H^2\iota$.
Hence
\begin{align}
\text{(Id)}=\sum_{t<\tau}\inpr{\qo-q^{\pio},G_t}\leq2c_0\zeta H^2T\iota=c_0\epsilon H^2\iota\lesssim H\iota,
\end{align}
where the equality follows from $\zeta=\epsilon/(2T)$, and the last inequality from $\epsilon H\leq1$.

Combining the bounds on (Ia)--(Id) with a union bound completes the proof.
\end{proof}

\subsection{Bounding term (II)}\label{app:delay}
By \cref{eq:source-estimate,eq:received}, exchanging the sums over $k$ and $t$ gives
\begin{align}
\text{(II)}=\sum_{k<t<\tau}\sum_{s,a}\frac{w_{k,t}(s,a)\ell_k(s,a)}{c_0\iota}\prn*{q^{P_{\calE_k},\pi_k}(s,a)-\bar q_t(s,a)}.
\end{align}
Thus, in episode $t$, the loss stored in episode $k$ at $(s,a)$ contributes
\begin{align}
\frac{w_{k,t}(s,a)\ell_k(s,a)q^{P_{\calE_k},\pi_k}(s,a)}{c_0\iota}\prn*{1-\frac{\bar q_t(s,a)}{q^{P_{\calE_k},\pi_k}(s,a)}}
\end{align}
to term (II).
To expand it to second order in $x=\bar q_t(s,a)/q^{P_{\calE_k},\pi_k}(s,a)$, we bound $1-x$ by the following smoothed version $f$ of $\max\set{1-x,0}$.
Let $\rho=c_0\eta SA\iota$, which is at most $1/4$ by the definition of $\eta$ in \cref{alg:main}, and let
\begin{align}
f(x)=
\begin{cases}
1-x,&0\leq x\leq1-\rho,\\
(1+\rho-x)^2/(4\rho),&1-\rho<x<1+\rho,\\
0,&x\geq1+\rho,
\end{cases}
\end{align}
which satisfies $\max\set{1-x,0}\leq f(x)\leq1$.

\begin{lemma}\label{lem:delay-ramp}
Let $x\geq0$ and $y\in\R$ with $\abs{y}\leq x/4$.
Then $f(x)\leq\rho/4$ if $x\geq1$, and
\begin{align}
\abs{f(x+y)-f(x)}\leq\abs{y}\ind\set*{x<\frac53},\qquad
0\leq f(x+y)-f(x)-f'(x)y\leq\frac{y^2}{4\rho}\ind\set*{x<\frac53}.
\end{align}
\end{lemma}
\begin{proof}
The function $f$ is convex, continuously differentiable, and nonincreasing with $-1\leq f'\leq0$, and $f''=1/(2\rho)$ on $(1-\rho,1+\rho)$ and $f''=0$ elsewhere.
Hence $f(x)\leq f(1)=\rho/4$ for $x\geq1$.
If $x\geq5/3$, then $x+y\geq3x/4\geq5/4\geq1+\rho$, so that $f(x+y)=f(x)=f'(x)=0$.
If $x<5/3$, then $\abs{f'}\leq1$ and Taylor's theorem with $0\leq f''\leq1/(2\rho)$ give the two inequalities.
\end{proof}

\begin{lemma}\label{lem:delay}
Term (II) in \cref{lem:decomposition} satisfies, with probability at least $1-\delta$,
\begin{align}
\text{(II)}\lesssim\eta HSAT\iota^3+\eta^2HS^3A^3T\iota^4+S^3A^2\iota^2+\prn*{H\iota+\eta\sqrt HS^{3/2}A^{3/2}\iota^3}\prn*{\sqrt{T\iota}+\iota}.
\end{align}
\end{lemma}

\begin{proof}[Proof of \cref{lem:delay}]
The arguments of $f$ below are nonnegative, since $\bar q_t^+(s,a)\geq q^{P_{\calE_t},\pi_t^+}(s,a)\geq0$ by \cref{eq:effective-hat} with $\hat q_t^+$ and $\pi_t^+$ in place of $\hat q_t$ and $\pi_t$, and $q^{P_{\calE_k},\pi_k}(s,a)>0$ whenever $w_{k,t}(s,a)>0$.
Let
\begin{align}
U_t=\sum_{k<t}\sum_{s,a}\ell_k(s,a)\prn*{w_{k,t}(s,a)-w_{k,t+1}(s,a)}f\prn*{\frac{\bar q_t(s,a)}{q^{P_{\calE_k},\pi_k}(s,a)}}.
\end{align}
Then
\begin{align}
\text{(II)}
&\leq\sum_{k<t<\tau}\sum_{s,a}\ell_k(s,a)\prn*{w_{k,t}(s,a)-w_{k,t+1}(s,a)}f\prn*{\frac{\bar q_{t}(s,a)}{q^{P_{\calE_k},\pi_k}(s,a)}}\\
&\quad+\sum_{t<\tau}\prn*{\E_t[U_t]-U_t}\\
&\leq\sum_{k<\tau}\sum_{s,a}\ell_k(s,a)w_{k,k+1}(s,a)f\prn*{\frac{\bar q_{k}(s,a)}{q^{P_{\calE_k},\pi_k}(s,a)}}\\
&\quad+\sum_{t\leq\tau-2}\sum_{k\leq t}\sum_{s,a}\ell_k(s,a)w_{k,t+1}(s,a)\prn*{f\prn*{\frac{\bar q_{t+1}(s,a)}{q^{P_{\calE_k},\pi_k}(s,a)}}-f\prn*{\frac{\bar q_t^+(s,a)}{q^{P_{\calE_k},\pi_k}(s,a)}}}\\
&\quad+\sum_{k\leq t<\tau}\sum_{s,a}\ell_k(s,a)w_{k,t+1}(s,a)\prn*{f\prn*{\frac{\bar q_t^+(s,a)}{q^{P_{\calE_k},\pi_k}(s,a)}}-f\prn*{\frac{\bar q_{t}(s,a)}{q^{P_{\calE_k},\pi_k}(s,a)}}}\\
&\quad+\sum_{t<\tau}\prn*{\E_t[U_t]-U_t},\label{eq:delay-split}
\end{align}
where the first inequality follows from $q^{P_{\calE_k},\pi_k}(s,a)-\bar q_t(s,a)\leq q^{P_{\calE_k},\pi_k}(s,a)f(\bar q_t(s,a)/q^{P_{\calE_k},\pi_k}(s,a))$, $f\geq0$ and $\E_t[w_{k,t}(s,a)-w_{k,t+1}(s,a)]=w_{k,t}(s,a)(\E_t[\tilde q_{k,t}(s,a)]+\gamma)/(c_0\iota)\geq w_{k,t}(s,a)q^{P_{\calE_k},\pi_k}(s,a)/(c_0\iota)$ by \cref{lem:service}, and the second from the equation
\begin{align}
\sum_{t=k+1}^{\tau-1}(w_t-w_{t+1})z_t=w_{k+1}z_k+\sum_{t=k}^{\tau-2}w_{t+1}\prn*{z_{t+1}-z_t^+}+\sum_{t=k}^{\tau-1}w_{t+1}\prn*{z_t^+-z_t}-w_\tau z_{\tau-1}^+
\end{align}
with $w_t=w_{k,t}(s,a)$, $z_t=f(\bar q_t(s,a)/q^{P_{\calE_k},\pi_k}(s,a))$, and $z_t^+=f(\bar q_t^+(s,a)/q^{P_{\calE_k},\pi_k}(s,a))$, together with $w_\tau z_{\tau-1}^+\geq0$, which follows from $f\geq0$ and $w_\tau\geq0$ by $0\leq1-(\tilde q_{k,t}(s,a)+\gamma)/(c_0\iota)\leq1$ in \cref{lem:service}.

For the first term in \cref{eq:delay-split},
\begin{align}
\sum_{k<\tau}\sum_{s,a}\ell_k(s,a)w_{k,k+1}(s,a)f\prn*{\frac{\bar q_{k}(s,a)}{q^{P_{\calE_k},\pi_k}(s,a)}}
&\leq\frac\rho4\sum_{k<\tau}\sum_{s,a}\I_k(s,a;\calE_k)\\
&\leq\frac{\rho HT}{4}=\frac{c_0\eta HSAT\iota}{4},
\end{align}
where the first inequality follows from $f(\bar q_k(s,a)/q^{P_{\calE_k},\pi_k}(s,a))\leq\rho/4$ by $\bar q_k(s,a)\geq q^{P_{\calE_k},\pi_k}(s,a)$ in \cref{lem:effective} and \cref{lem:delay-ramp}, and $w_{k,k+1}(s,a)=\I_k(s,a;\calE_k)$.

For the second term in \cref{eq:delay-split}, let $t\leq\tau-2$.
If no edge is activated in episode $t$, then $\calE_{t+1}=\calE_t$ and $\hat q_{t+1}=\hat q_t^+$, so that $\bar q_{t+1}=\bar q_t^+$ and episode $t$ does not contribute to the second term.
If an edge is activated in episode $t$, then
\begin{align}
&\sum_{k\leq t}\sum_{s,a}\ell_k(s,a)w_{k,t+1}(s,a)\prn*{f\prn*{\frac{\bar q_{t+1}(s,a)}{q^{P_{\calE_k},\pi_k}(s,a)}}-f\prn*{\frac{\bar q_t^+(s,a)}{q^{P_{\calE_k},\pi_k}(s,a)}}}\\
&\leq\sum_{s,a}\sum_{k\leq t}w_{k,t+1}(s,a)\leq c_0^2SA\iota^2,
\end{align}
where the first inequality follows from $\ell_k\leq1$ and $0\leq f\leq1$, and the second from \cref{eq:good-event} for $t+1<\tau$.
Thus, the second term is at most $c_0^2S^3A^2\iota^2$, because an edge is activated in at most $S^2A$ episodes $t$.

For the third term in \cref{eq:delay-split}, let
\begin{align}
M_t=\sum_{k\leq t}\sum_{s,a}\ell_k(s,a)w_{k,t+1}(s,a)\prn*{f\prn*{\frac{\bar q_t^+(s,a)}{q^{P_{\calE_k},\pi_k}(s,a)}}-f\prn*{\frac{\bar q_{t}(s,a)}{q^{P_{\calE_k},\pi_k}(s,a)}}}.
\end{align}
Then the third term is $\sum_{t<\tau}M_t$, and we first bound $\sum_{t<\tau}\E_t[M_t]$.
To this end, fix $k\leq t<\tau$ and $(s,a)$, and denote
\begin{align}
x=\frac{\bar q_t(s,a)}{q^{P_{\calE_k},\pi_k}(s,a)},\qquad y=\frac{\bar q_t^+(s,a)-\bar q_t(s,a)}{q^{P_{\calE_k},\pi_k}(s,a)},
\end{align}
which satisfy the condition $\abs{y}\leq x/4$ of \cref{lem:delay-ramp} by \cref{eq:pointwise-motion} in \cref{lem:motion}.
For $t=k$,
\begin{align}
&\E_k\brk*{\ell_k(s,a)w_{k,k+1}(s,a)\prn*{f\prn*{\frac{\bar q_k^+(s,a)}{q^{P_{\calE_k},\pi_k}(s,a)}}-f\prn*{\frac{\bar q_{k}(s,a)}{q^{P_{\calE_k},\pi_k}(s,a)}}}}\\
&\leq\frac{\E_k\brk*{\I_k(s,a;\calE_k)\abs{\bar q_k^+(s,a)-\bar q_k(s,a)}}}{q^{P_{\calE_k},\pi_k}(s,a)}\ind\set*{x<\frac53}\\
&\leq\frac{\sqrt{\E_k[\I_k(s,a;\calE_k)^2]\E_k[(\bar q_k^+(s,a)-\bar q_k(s,a))^2]}}{q^{P_{\calE_k},\pi_k}(s,a)}\ind\set*{x<\frac53}\tag{by the Cauchy--Schwarz inequality}\\
&=\sqrt{\frac{\E_k[(\bar q_k^+(s,a)-\bar q_k(s,a))^2]}{q^{P_{\calE_k},\pi_k}(s,a)}}\ind\set*{x<\frac53}\\
&\leq\sqrt{\frac{10}3\hat q_k(s,a)}\sqrt{\frac{\E_k[(\bar q_k^+(s,a)-\bar q_k(s,a))^2]}{\hat q_k(s,a)^2}},
\end{align}
where the first inequality follows from $\ell_k\leq1$, $w_{k,k+1}(s,a)=\I_k(s,a;\calE_k)$, and \cref{lem:delay-ramp}, and the last from $q^{P_{\calE_k},\pi_k}(s,a)>\frac35\bar q_k(s,a)\geq\frac3{10}\hat q_k(s,a)$ for $x<5/3$ by \cref{lem:effective}.
Summing over $(s,a)$ and $k<\tau$ gives
\begin{align}
&\sum_{k<\tau}\sum_{s,a}\sqrt{\frac{10}3\hat q_k(s,a)}\sqrt{\frac{\E_k[(\bar q_k^+(s,a)-\bar q_k(s,a))^2]}{\hat q_k(s,a)^2}}\\
&\leq\sum_{k<\tau}\sqrt{\frac{10}3\sum_{s,a}\hat q_k(s,a)}\sqrt{\sum_{s,a}\frac{\E_k[(\bar q_k^+(s,a)-\bar q_k(s,a))^2]}{\hat q_k(s,a)^2}}\tag{by the Cauchy--Schwarz inequality}\\
&\lesssim\eta\sqrt HSAT\iota,
\end{align}
where the last inequality follows from \cref{lem:motion}.
This bounds the contribution of $t=k$ to $\sum_{t<\tau}\E_t[M_t]$.

For $t>k$,
\begin{align}
&\E_t\brk*{\ell_k(s,a)w_{k,t+1}(s,a)\prn*{f\prn*{\frac{\bar q_t^+(s,a)}{q^{P_{\calE_k},\pi_k}(s,a)}}-f\prn*{\frac{\bar q_{t}(s,a)}{q^{P_{\calE_k},\pi_k}(s,a)}}}}\\
&=\ell_k(s,a)w_{k,t}(s,a)\E_t\brk*{\prn*{1-\frac{\tilde q_{k,t}(s,a)+\gamma}{c_0\iota}}\prn*{f(x+y)-f(x)}}\\
&\leq\ell_k(s,a)w_{k,t}(s,a)\ind\set*{x<\frac53}\E_t\brk*{\prn*{1-\frac{\tilde q_{k,t}(s,a)+\gamma}{c_0\iota}}\prn*{f'(x)y+\frac{y^2}{4\rho}}}\\
&\leq\ell_k(s,a)w_{k,t}(s,a)\ind\set*{x<\frac53}\prn*{f'(x)\E_t[y]-\frac{f'(x)\E_t[(\tilde q_{k,t}(s,a)+\gamma)y]}{c_0\iota}+\frac{\E_t[y^2]}{4\rho}}\\
&\leq w_{k,t}(s,a)\ind\set*{x<\frac53}\prn*{2\abs{\E_t[y]}+\frac{\sqrt{2q^{P_{\calE_k},\pi_k}(s,a)\E_t[y^2]}}{c_0\iota}+\frac{\E_t[y^2]}{4\rho}}\\
&\leq w_{k,t}(s,a)\Biggl(\frac{20\abs{\E_t[\bar q_t^+(s,a)-\bar q_t(s,a)]}}{3\hat q_t(s,a)}+\frac{\sqrt{\frac{20}3\hat q_t(s,a)}}{c_0\iota}\sqrt{\frac{\E_t[(\bar q_t^+(s,a)-\bar q_t(s,a))^2]}{\hat q_t(s,a)^2}}\\
&\hspace{6em}+\frac{25\E_t[(\bar q_t^+(s,a)-\bar q_t(s,a))^2]}{9\rho\hat q_t(s,a)^2}\Biggr),
\end{align}
where the first inequality follows from \cref{lem:delay-ramp}, $f'(x)=0$ for $x\geq5/3\geq1+\rho$, and $1-(\tilde q_{k,t}(s,a)+\gamma)/(c_0\iota)\geq0$ by \cref{lem:service}, the second from $1-(\tilde q_{k,t}(s,a)+\gamma)/(c_0\iota)\leq1$, the third from $\ell_k\leq1$, $\abs{f'}\leq1$, $\gamma\leq c_0\iota$, the Cauchy--Schwarz inequality, and $\E_t[\tilde q_{k,t}(s,a)^2]\leq2q^{P_{\calE_k},\pi_k}(s,a)$ by \cref{lem:service}, and the fourth from $q^{P_{\calE_k},\pi_k}(s,a)>\frac35\bar q_t(s,a)\geq\frac3{10}\hat q_t(s,a)$ for $x<5/3$ by \cref{lem:effective}.
Summing over $(s,a)$ and $k<t<\tau$ gives
\begin{align}
&\sum_{t<\tau}\sum_{k<t}\sum_{s,a}w_{k,t}(s,a)\Biggl(\frac{20\abs{\E_t[\bar q_t^+(s,a)-\bar q_t(s,a)]}}{3\hat q_t(s,a)}+\frac{\sqrt{\frac{20}3\hat q_t(s,a)}}{c_0\iota}\sqrt{\frac{\E_t[(\bar q_t^+(s,a)-\bar q_t(s,a))^2]}{\hat q_t(s,a)^2}}\\
&\hspace{6em}+\frac{25\E_t[(\bar q_t^+(s,a)-\bar q_t(s,a))^2]}{9\rho\hat q_t(s,a)^2}\Biggr)\\
&\lesssim\sum_{t<\tau}\Biggl(\iota^2\sum_{s,a}\frac{\abs{\E_t[\bar q_t^+(s,a)-\bar q_t(s,a)]}}{\hat q_t(s,a)}+\iota\sqrt H\sqrt{\sum_{s,a}\frac{\E_t[(\bar q_t^+(s,a)-\bar q_t(s,a))^2]}{\hat q_t(s,a)^2}}\\
&\hspace{6em}+\frac{\iota^2}{\rho}\sum_{s,a}\frac{\E_t[(\bar q_t^+(s,a)-\bar q_t(s,a))^2]}{\hat q_t(s,a)^2}\Biggr)\\
&\lesssim\prn*{\iota^2\prn*{\eta SA\iota+\eta^2HS^3A^3\iota^2}+\iota\sqrt H\eta SA\iota+\frac{\iota}{\eta SA}\eta^2S^2A^2\iota^2}T\\
&\lesssim\eta\sqrt HSAT\iota^2+\eta SAT\iota^3+\eta^2HS^3A^3T\iota^4,
\end{align}
where the first inequality follows from $\sum_{k<t}w_{k,t}(s,a)\leq c_0^2\iota^2$ by \cref{eq:good-event} and the Cauchy--Schwarz inequality, and the second from \cref{lem:motion} and $\rho=c_0\eta SA\iota$.
This bounds the contribution of $t>k$ to $\sum_{t<\tau}\E_t[M_t]$.

It remains to bound $\sum_{t<\tau}\prn*{(M_t-\E_t[M_t])-(U_t-\E_t[U_t])}$ by \cref{lem:martingale}.
We first bound $\abs{U_t-\E_t[U_t]}$ and $\abs{M_t}$ for $t<\tau$.
Since $w_{k,t}(s,a)-w_{k,t+1}(s,a)=w_{k,t}(s,a)(\tilde q_{k,t}(s,a)+\gamma)/(c_0\iota)$,
\begin{align}
U_t-\E_t[U_t]=\frac1{c_0\iota}\sum_{k<t}\sum_{s,a}\ell_k(s,a)w_{k,t}(s,a)f\prn*{\frac{\bar q_t(s,a)}{q^{P_{\calE_k},\pi_k}(s,a)}}\prn*{\tilde q_{k,t}(s,a)-\E_t[\tilde q_{k,t}(s,a)]}.
\end{align}
For $k<r_t$, $\tilde q_{k,t}(s,a)=\E_t[\tilde q_{k,t}(s,a)]$ by \cref{eq:upper-probability}.
For $k\geq r_t$, \cref{eq:service} and \cref{lem:service} give $0\leq\tilde q_{k,t}(s,a)\leq2\I_t(s,a;\calE_k)\leq2\ind\set{(s_{t,h(s)},a_{t,h(s)})=(s,a)}$.

Then, by $\ell_k\leq1$ and $0\leq f\leq1$, we obtain
\begin{align}
0&\leq\frac1{c_0\iota}\sum_{r_t\leq k<t}\sum_{s,a}\ell_k(s,a)w_{k,t}(s,a)f\prn*{\frac{\bar q_t(s,a)}{q^{P_{\calE_k},\pi_k}(s,a)}}\tilde q_{k,t}(s,a)\\
&\leq\frac2{c_0\iota}\sum_{s,a}\ind\set{(s_{t,h(s)},a_{t,h(s)})=(s,a)}\sum_{k<t}w_{k,t}(s,a)
\leq2c_0H\iota,
\end{align}
where the last inequality follows from \cref{eq:good-event} and $\sum_{s,a}\ind\set{(s_{t,h(s)},a_{t,h(s)})=(s,a)}=H$.
Thus $\abs{U_t-\E_t[U_t]}\leq2c_0H\iota$.
For $M_t$, with $x$ and $y$ as above, we have
\begin{align}
\abs{M_t}
&\leq\sum_{s,a}\sum_{k\leq t}w_{k,t+1}(s,a)\abs{f(x+y)-f(x)}\\
&\leq\sum_{s,a}\sum_{k\leq t}w_{k,t+1}(s,a)\frac{\abs{\bar q_t^+(s,a)-\bar q_t(s,a)}}{q^{P_{\calE_k},\pi_k}(s,a)}\ind\set*{x<\frac53}\\
&\leq\frac{10}{3}\sum_{s,a}\sum_{k\leq t}w_{k,t+1}(s,a)\frac{\abs{\bar q_t^+(s,a)-\bar q_t(s,a)}}{\hat q_t(s,a)}\\
&\leq\frac{10}{3}\prn*{c_0^2\iota^2+1}\sum_{s,a}\prn*{\frac{\abs{\hat q_t^+(s,a)-\hat q_t(s,a)}}{\hat q_t(s,a)}+\frac{\abs{\bar q_t^+(s,a)-\bar q_t(s,a)-\prn*{\hat q_t^+(s,a)-\hat q_t(s,a)}}}{\hat q_t(s,a)}}\\
&\leq\frac{10}{3}\prn*{c_0^2\iota^2+1}\prn*{\sqrt{SA\sum_{s,a}\prn*{\frac{\hat q_t^+(s,a)-\hat q_t(s,a)}{\hat q_t(s,a)}}^2}+\frac{24SA}{\sqrt H}\nrm{\hat q_t^+-\hat q_t}_t}\\
&\leq\frac{10}{3}\prn*{c_0^2\iota^2+1}\prn*{\sqrt{\frac{2}{H}}+\frac{24}{\sqrt H}}SA\nrm{\hat q_t^+-\hat q_t}_t\\
&\lesssim\eta\sqrt HS^{3/2}A^{3/2}\iota^3,
\end{align}
where the second inequality follows from \cref{lem:delay-ramp}, the third from $q^{P_{\calE_k},\pi_k}(s,a)>\frac35\bar q_t(s,a)\geq\frac3{10}\hat q_t(s,a)$ for $x<5/3$ by \cref{lem:effective}, the fourth from $\sum_{k\leq t}w_{k,t+1}(s,a)\leq\sum_{k<t}w_{k,t}(s,a)+w_{t,t+1}(s,a)\leq c_0^2\iota^2+1$ by $w_{k,t+1}(s,a)\leq w_{k,t}(s,a)$, \cref{eq:good-event}, and $w_{t,t+1}(s,a)\leq1$, the fifth from the Cauchy--Schwarz inequality and \cref{eq:motion-pointwise}, the sixth from \cref{eq:relative-coordinates} in \cref{lem:ratio}, and the last from \cref{eq:step-size}.

Thus $\abs{(M_t-\E_t[M_t])-(U_t-\E_t[U_t])}\leq R$ for $t<\tau$ with $R\lesssim H\iota+\eta\sqrt HS^{3/2}A^{3/2}\iota^3$, and \cref{lem:martingale} with $V=R^2T$ and $\log(1/\delta)\leq \iota$ gives, with probability at least $1-\delta$,
\begin{align}
\sum_{t<\tau}\prn*{(M_t-\E_t[M_t])-(U_t-\E_t[U_t])}&\leq R\sqrt{2T\iota}+\frac23R\iota\\
&\lesssim\prn*{H\iota+\eta\sqrt HS^{3/2}A^{3/2}\iota^3}\prn*{\sqrt{T\iota}+\iota}.
\end{align}

Substituting the four bounds into \cref{eq:delay-split} proves the lemma.
\end{proof}

\subsection{Bounding term (III)}\label{app:source}
\begin{lemma}\label{lem:source}
Term (III) in \cref{lem:decomposition} satisfies, with probability at least $1-2\delta$,
\begin{align}
\text{(III)}\lesssim\eta HSAT\iota^3+S^3A^2\iota^2+H\sqrt T\iota^{3/2}+\eta H^{3/2}SA\iota^4+\sqrt{SAT}\iota.
\end{align}
\end{lemma}
\begin{proof}
By the definition of (III) and \cref{eq:source-estimate},
\begin{align}
\text{(III)}
&=\sum_{k<\tau}\sum_{s,a}\ell_k(s,a)q^{P_{\calE_k},\pi_k}(s,a)\prn*{1-\sum_{t=k+1}^{\tau-1}\frac{w_{k,t}(s,a)}{c_0\iota}}\\
&=\sum_{k<\tau}\sum_{s,a}\ell_k(s,a)w_{k,\tau}(s,a)\\
&\quad+\frac{1}{c_0\iota}\sum_{k<\tau}\sum_{s,a}\ell_k(s,a)\sum_{t=k+1}^{\tau-1}w_{k,t}(s,a)\prn*{\E_t[\tilde q_{k,t}(s,a)]-q^{P_{\calE_k},\pi_k}(s,a)}\\
&\quad+\sum_{t<\tau}Y_t+\gamma\sum_{t<\tau}\sum_{s,a}G_t(s,a),\label{eq:source-split}
\end{align}
where
\begin{align}
Y_t=\sum_{s,a}\ell_t(s,a)\prn*{q^{P_{\calE_t},\pi_t}(s,a)-\I_t(s,a;\calE_t)}+\frac{1}{c_0\iota}\sum_{k<t}\sum_{s,a}\ell_k(s,a)w_{k,t}(s,a)\prn*{\tilde q_{k,t}(s,a)-\E_t[\tilde q_{k,t}(s,a)]},
\end{align}
and the second equality follows from the identity
\begin{align}
&q^{P_{\calE_k},\pi_k}(s,a)\prn*{1-\sum_{t=k+1}^{\tau-1}\frac{w_{k,t}(s,a)}{c_0\iota}}\\
&=\prn*{q^{P_{\calE_k},\pi_k}(s,a)-\I_k(s,a;\calE_k)}+\I_k(s,a;\calE_k)-\sum_{t=k+1}^{\tau-1}\frac{w_{k,t}(s,a)q^{P_{\calE_k},\pi_k}(s,a)}{c_0\iota}\\
&=\prn*{q^{P_{\calE_k},\pi_k}(s,a)-\I_k(s,a;\calE_k)}+w_{k,\tau}(s,a)+\sum_{t=k+1}^{\tau-1}\frac{w_{k,t}(s,a)\prn*{\tilde q_{k,t}(s,a)+\gamma-q^{P_{\calE_k},\pi_k}(s,a)}}{c_0\iota},\tag{by \cref{eq:weight-identity}}
\end{align}
together with the exchange $\sum_{k<\tau}\sum_{t=k+1}^{\tau-1}=\sum_{t<\tau}\sum_{k<t}$, the definition of $G_t$ in \cref{eq:received}, and the decomposition $\tilde q_{k,t}(s,a)=\E_t[\tilde q_{k,t}(s,a)]+(\tilde q_{k,t}(s,a)-\E_t[\tilde q_{k,t}(s,a)])$.

For the first term in \cref{eq:source-split},
\begin{align}
\sum_{k<\tau}\sum_{s,a}\ell_k(s,a)w_{k,\tau}(s,a)\leq\sum_{s,a}\prn*{\sum_{k<\tau-1}w_{k,\tau-1}(s,a)+w_{\tau-1,\tau}(s,a)}\leq SA(c_0^2\iota^2+1),
\end{align}
where the first inequality follows from $w_{k,\tau}(s,a)\leq w_{k,\tau-1}(s,a)$, and the second from \cref{eq:good-event} and $w_{\tau-1,\tau}(s,a)=\I_{\tau-1}(s,a;\calE_{\tau-1})\leq1$.

For the second term in \cref{eq:source-split}, fix $k<\tau$ and $(s,a)$.
If the reference episode is not updated in any episode $k<t<\tau$, then $r_t\leq k$ and $\E_t[\tilde q_{k,t}(s,a)]=q^{P_{\calE_k},\pi_k}(s,a)$ for all such $t$ by \cref{lem:service}, and hence the sum over $t$ in the second term is zero.
Otherwise, let $u$ be the first episode with $k<u<\tau$ and $r_u=u$.
Then
\begin{align}
&\sum_{t=k+1}^{\tau-1}\frac{w_{k,t}(s,a)\prn*{\E_t[\tilde q_{k,t}(s,a)]-q^{P_{\calE_k},\pi_k}(s,a)}}{c_0\iota}\\
&=\sum_{t=k+1}^{u-1}\frac{w_{k,t}(s,a)\prn*{\E_t[\tilde q_{k,t}(s,a)]-q^{P_{\calE_k},\pi_k}(s,a)}}{c_0\iota}+\sum_{t=u}^{\tau-1}\frac{w_{k,t}(s,a)\prn*{\E_t[\tilde q_{k,t}(s,a)]-q^{P_{\calE_k},\pi_k}(s,a)}}{c_0\iota}\\
&=\sum_{t=u}^{\tau-1}\frac{w_{k,t}(s,a)\prn*{\mathrm{e}^{4H\epsilon}q^{\hat P_{\calE_k},\pi_k}(s,a)-q^{P_{\calE_k},\pi_k}(s,a)}}{c_0\iota}\\
&=\frac{w_{k,u}(s,a)\prn*{\mathrm{e}^{4H\epsilon}q^{\hat P_{\calE_k},\pi_k}(s,a)-q^{P_{\calE_k},\pi_k}(s,a)}}{c_0\iota}\sum_{t=u}^{\tau-1}\prn*{1-\frac{\mathrm{e}^{4H\epsilon}q^{\hat P_{\calE_k},\pi_k}(s,a)+\gamma}{c_0\iota}}^{t-u}\\
&\leq\frac{w_{k,u}(s,a)\prn*{\mathrm{e}^{4H\epsilon}q^{\hat P_{\calE_k},\pi_k}(s,a)-q^{P_{\calE_k},\pi_k}(s,a)}}{\mathrm{e}^{4H\epsilon}q^{\hat P_{\calE_k},\pi_k}(s,a)+\gamma}\\
&\leq w_{k,u}(s,a)\prn*{1-\frac{q^{P_{\calE_k},\pi_k}(s,a)}{\mathrm{e}^{4H\epsilon}q^{\hat P_{\calE_k},\pi_k}(s,a)}}\leq w_{k,u}(s,a)\prn*{1-\mathrm{e}^{-8H\epsilon}}\leq8H\epsilon w_{k,u}(s,a),
\end{align}
where the second equality follows from \cref{lem:service} for $k<t<u$ and the definition \cref{eq:upper-probability} for $u\leq t<\tau$, the third from the update of $w_{k,t}(s,a)$, the third inequality from \cref{lem:upper-comparison}, and the fourth from $1-\mathrm{e}^{-x}\leq x$.
Thus, summing over $k<\tau$ and $(s,a)$ gives, with $N=\abs{\set{u<\tau:r_u=u}}$,
\begin{align}
&\frac{1}{c_0\iota}\sum_{k<\tau}\sum_{s,a}\ell_k(s,a)\sum_{t=k+1}^{\tau-1}w_{k,t}(s,a)\prn*{\E_t[\tilde q_{k,t}(s,a)]-q^{P_{\calE_k},\pi_k}(s,a)}\\
&\leq8H\epsilon\sum_{u<\tau:r_u=u}\sum_{s,a}\sum_{k<u}w_{k,u}(s,a)\leq8c_0^2H\epsilon SA\iota^2N,
\end{align}
where the last inequality used \cref{eq:good-event} for $u<\tau$.

To bound $N$, let $u<u'<\tau$ be consecutive elements of $\set{t:r_t=t}$ with $\calE_u=\calE_{u'}$.
Then
\begin{align}
\frac{\log2}{2}<\log R(\pi_{u'},\pi_u)\leq\sum_{u\leq t<u'}\log R(\pi_{t+1},\pi_t)=\sum_{u\leq t<u'}\log R(\pi_t^+,\pi_t),\label{eq:reference-gap}
\end{align}
where the first inequality follows from \cref{line:epoch} with $r_{u'-1}=u$, the second from the definition of $R$ in \cref{eq:epoch} and the triangle inequality $\abs{\log\pi_{u'}(a\mid s)-\log\pi_u(a\mid s)}\leq\sum_{t=u}^{u'-1}\abs{\log\pi_{t+1}(a\mid s)-\log\pi_t(a\mid s)}$ for each $(s,a)$, and the equality from $\calE_{t+1}=\calE_t$.
Then, using $0\leq\log R(\pi_t^+,\pi_t)\leq80\sqrt2c_0\eta HSA\iota$ by \cref{eq:log-ratio} and \cref{lem:local-ftrl}, with probability at least $1-\delta$,
\begin{align}
N&\leq S^2A+1+\frac{2}{\log2}\sum_{t<\tau}\log R(\pi_t^+,\pi_t)\\
&\leq S^2A+1+\frac{2}{\log2}\prn*{\frac32\sum_{t<\tau}\E_t\brk*{\log R(\pi_t^+,\pi_t)}+160\sqrt2c_0\eta HSA\iota^2}\tag{by \cref{lem:bounded-sum}}\\
&\lesssim S^2A+\eta\sqrt HSAT\iota+\eta HSA\iota^2,\tag{by \cref{lem:policy-motion}}
\end{align}
where the first inequality follows from \cref{eq:reference-gap} for consecutive pairs and the fact that an edge is activated in at most $S^2A$ episodes.

For the third term in \cref{eq:source-split}, let $t<\tau$.
For $k<r_t$, $\tilde q_{k,t}(s,a)=\E_t[\tilde q_{k,t}(s,a)]$ by \cref{eq:upper-probability}.
For $k\geq r_t$, \cref{eq:service} and \cref{lem:service} give $0\leq\tilde q_{k,t}(s,a)\leq2\I_t(s,a;\calE_k)\leq2\ind\set{(s_{t,h(s)},a_{t,h(s)})=(s,a)}$, and hence
\begin{align}
0&\leq\frac1{c_0\iota}\sum_{r_t\leq k<t}\sum_{s,a}\ell_k(s,a)w_{k,t}(s,a)\tilde q_{k,t}(s,a)\\
&\leq\frac2{c_0\iota}\sum_{s,a}\ind\set{(s_{t,h(s)},a_{t,h(s)})=(s,a)}\sum_{k<t}w_{k,t}(s,a)\leq2c_0H\iota.\tag{by \cref{eq:good-event}}
\end{align}
Therefore
\begin{align}
\abs{Y_t}
&\leq\abs*{\sum_{s,a}\ell_t(s,a)\prn*{q^{P_{\calE_t},\pi_t}(s,a)-\I_t(s,a;\calE_t)}}\\
&\quad+\frac1{c_0\iota}\abs*{\sum_{k<t}\sum_{s,a}\ell_k(s,a)w_{k,t}(s,a)\prn*{\tilde q_{k,t}(s,a)-\E_t[\tilde q_{k,t}(s,a)]}}\\
&\leq H+2c_0H\iota\leq3c_0H\iota.
\end{align}
Thus \cref{lem:martingale} with $R=3c_0H\iota$, $V=9c_0^2H^2\iota^2T$, and $\log(1/\delta)\leq \iota$ gives, with probability at least $1-\delta$, $\sum_{t<\tau}Y_t\leq3\sqrt2c_0H\sqrt T\iota^{3/2}+2c_0H\iota^2$.
For the fourth term in \cref{eq:source-split}, $G_t(s,a)\leq c_0\iota$ by \cref{lem:bounded-loss} and $\gamma\leq1/\sqrt{SAT}$ give $\gamma\sum_{t<\tau}\sum_{s,a}G_t(s,a)\leq c_0\gamma SAT\iota\leq c_0\sqrt{SAT}\iota$.

Substituting these four bounds into \cref{eq:source-split}, we obtain
\begin{align}
\text{(III)}
&\leq SA(c_0^2\iota^2+1)+8c_0^2H\epsilon SA\iota^2N+3\sqrt2c_0H\sqrt T\iota^{3/2}+2c_0H\iota^2+c_0\sqrt{SAT}\iota\\
&\lesssim SA\iota^2+\epsilon HS^3A^2\iota^2+\eta\epsilon H\sqrt HS^2A^2T\iota^3+\eta\epsilon H^2S^2A^2\iota^4+H\sqrt T\iota^{3/2}+H\iota^2+\sqrt{SAT}\iota\\
&\lesssim S^3A^2\iota^2+\eta HSAT\iota^3+\eta H^{3/2}SA\iota^4+H\sqrt T\iota^{3/2}+\sqrt{SAT}\iota,
\end{align}
where we use $\epsilon\sqrt HSA\lesssim 1$ by the definition of $\epsilon$ in \cref{alg:main} and $H\leq S$.
\end{proof}

\subsection{Bounding term (IV)}\label{app:comparator}
\begin{lemma}\label{lem:comparator}
Term (IV) in \cref{lem:decomposition} satisfies, with probability at least $1-\delta$,
\begin{align}
\text{(IV)}\lesssim H\sqrt{SAT}\iota^2+HSA\iota^2.
\end{align}
\end{lemma}
\begin{proof}
By \cref{eq:source-estimate,eq:received},
\begin{align}
\text{(IV)}
=\sum_{t<\tau}\inpr*{q^{\pio},\hat\ell_t-\ell_t}
&=\sum_{t<\tau}\inpr*{q^{\pio},G_t-\ell_t}.
\end{align}
We show below that, with probability at least $1-\delta$, for all $(s,a)$,
\begin{align}\label{eq:comparator-coordinate}
\sum_{t<\tau}\prn*{G_t(s,a)-\ell_t(s,a)}\leq\frac{16c_0\iota}{\gamma}\log\frac{SA}{\delta}.
\end{align}
Fix $(s,a)$ and $k<\tau$.
First, by \cref{eq:weight-identity} and $w_{k,\tau}(s,a)\geq0$,
\begin{align}
&\sum_{t=k+1}^{\tau-1}\frac{w_{k,t}(s,a)}{c_0\iota}\prn*{q^{P_{\calE_k},\pi_k}(s,a)+\frac\gamma2}+\sum_{t=k+1}^{\tau-1}\frac{w_{k,t}(s,a)}{c_0\iota}\prn*{\tilde q_{k,t}(s,a)-q^{P_{\calE_k},\pi_k}(s,a)+\frac\gamma2}\\
&=\sum_{t=k+1}^{\tau-1}\frac{w_{k,t}(s,a)\prn*{\tilde q_{k,t}(s,a)+\gamma}}{c_0\iota}\leq\I_k(s,a;\calE_k).
\end{align}
Dividing both sides by $q^{P_{\calE_k},\pi_k}(s,a)+\gamma/2$ gives
\begin{align}
\sum_{t=k+1}^{\tau-1}\frac{w_{k,t}(s,a)}{c_0\iota}
&\leq\frac{\I_k(s,a;\calE_k)}{q^{P_{\calE_k},\pi_k}(s,a)+\gamma/2}-\sum_{t=k+1}^{\tau-1}\frac{w_{k,t}(s,a)}{c_0\iota}\cdot\frac{\tilde q_{k,t}(s,a)-q^{P_{\calE_k},\pi_k}(s,a)+\gamma/2}{q^{P_{\calE_k},\pi_k}(s,a)+\gamma/2}.
\end{align}
Multiplying by $\ell_k(s,a)$, subtracting $\ell_k(s,a)$, and summing over $k<\tau$ give
\begin{align}\label{eq:comparator-step}
\sum_{t<\tau}\prn*{G_t(s,a)-\ell_t(s,a)}\leq\sum_{t<\tau}Z_t,
\end{align}
where
\begin{align}
Z_t&=\ell_t(s,a)\prn*{\frac{\I_t(s,a;\calE_t)}{q^{P_{\calE_t},\pi_t}(s,a)+\gamma/2}-1}-\sum_{k<t}\frac{\ell_k(s,a)w_{k,t}(s,a)}{c_0\iota}\cdot\frac{\tilde q_{k,t}(s,a)-q^{P_{\calE_k},\pi_k}(s,a)+\gamma/2}{q^{P_{\calE_k},\pi_k}(s,a)+\gamma/2}.
\end{align}
Let $t<\tau$ and
\begin{align}
\alpha_t=\frac{\ell_t(s,a)}{q^{P_{\calE_t},\pi_t}(s,a)+\gamma/2}+\sum_{k<t}\frac{\ell_k(s,a)w_{k,t}(s,a)}{c_0\iota\prn*{q^{P_{\calE_k},\pi_k}(s,a)+\gamma/2}}.
\end{align}
By $\E_t[\I_t(s,a;\calE_t)]=q^{P_{\calE_t},\pi_t}(s,a)$ and $\E_t[\tilde q_{k,t}(s,a)]\geq q^{P_{\calE_k},\pi_k}(s,a)$ in \cref{lem:service}, $\E_t[Z_t]\leq-\gamma \alpha_t/2$.
By $0\leq\tilde q_{k,t}(s,a)\leq2$, $\ell_t(s,a)\leq1$, and $G_t(s,a)\leq c_0\iota$ in \cref{lem:bounded-loss}, $\abs{Z_t}\leq(2+5c_0\iota)/\gamma\leq6c_0\iota/\gamma$.
Moreover, by $q^{P_{\calE_t},\pi_t}(s,a)+\gamma/2\leq5/4$,
\begin{align}
\E_t\brk*{\prn*{\frac{\I_t(s,a;\calE_t)}{q^{P_{\calE_t},\pi_t}(s,a)+\gamma/2}-1}^2}
&=\frac{q^{P_{\calE_t},\pi_t}(s,a)}{\prn*{q^{P_{\calE_t},\pi_t}(s,a)+\gamma/2}^2}-\frac{2q^{P_{\calE_t},\pi_t}(s,a)}{q^{P_{\calE_t},\pi_t}(s,a)+\gamma/2}+1\\
&\leq\frac{1}{q^{P_{\calE_t},\pi_t}(s,a)+\gamma/2}+1\leq\frac{9/4}{q^{P_{\calE_t},\pi_t}(s,a)+\gamma/2},
\end{align}
and, by $\E_t[\tilde q_{k,t}(s,a)]\geq q^{P_{\calE_k},\pi_k}(s,a)$ and $\E_t[\tilde q_{k,t}(s,a)^2]\leq2q^{P_{\calE_k},\pi_k}(s,a)$ in \cref{lem:service},
\begin{align}
\E_t\brk*{\prn*{\tilde q_{k,t}(s,a)-q^{P_{\calE_k},\pi_k}(s,a)+\frac\gamma2}^2}
&\leq2\E_t\brk*{\prn*{\tilde q_{k,t}(s,a)-q^{P_{\calE_k},\pi_k}(s,a)}^2}+\frac{\gamma^2}2\\
&\leq2\E_t[\tilde q_{k,t}(s,a)^2]+\frac{\gamma^2}2\leq4\prn*{q^{P_{\calE_k},\pi_k}(s,a)+\frac\gamma2}.
\end{align}
Then, by $(x+y)^2\leq2x^2+2y^2$ and the Cauchy--Schwarz inequality with $G_t(s,a)\leq c_0\iota$,
\begin{align}
\E_t[Z_t^2]&\leq\frac{9\ell_t(s,a)}{2\prn*{q^{P_{\calE_t},\pi_t}(s,a)+\gamma/2}}+8c_0\iota\sum_{k<t}\frac{\ell_k(s,a)w_{k,t}(s,a)}{c_0\iota\prn*{q^{P_{\calE_k},\pi_k}(s,a)+\gamma/2}}\leq8c_0\iota \alpha_t.
\end{align}
Let $\lambda=\gamma/(16c_0\iota)$.
Then $\abs{\lambda Z_t}\leq1$ and $\mathrm{e}^z\leq1+z+z^2$ for $\abs z\leq1$ give
\begin{align}
\E_t\brk*{\mathrm{e}^{\lambda Z_t}}\leq1+\lambda\E_t[Z_t]+\lambda^2\E_t[Z_t^2]\leq1-\lambda \alpha_t\prn*{\frac\gamma2-8c_0\iota\lambda}=1.
\end{align}
Thus \cref{lem:supermartingale} with $\delta/(SA)$ in place of $\delta$ gives, with probability at least $1-\delta/(SA)$,
\begin{align}
\sum_{t<\tau}\prn*{G_t(s,a)-\ell_t(s,a)}\leq\sum_{t<\tau}Z_t\leq\frac1\lambda\log\frac{SA}{\delta}=\frac{16c_0\iota}{\gamma}\log\frac{SA}{\delta}.
\end{align}
Taking a union bound over $(s,a)$, we obtain \cref{eq:comparator-coordinate} for all $(s,a)$ with probability at least $1-\delta$, and hence
\begin{align}
\text{(IV)}
\leq\frac{16c_0\iota}{\gamma}\log\frac{SA}{\delta}\sum_{s,a}q^{\pio}(s,a)\leq16c_0H\prn*{\sqrt{SAT}+SA}\iota^2,
\end{align}
where the last inequality follows from $\sum_{s,a}q^{\pio}(s,a)=H$, $\gamma=1/(\sqrt{SAT}+SA)$, and $\log(SA/\delta)\leq \iota$.
\end{proof}

\subsection{Bounding term (V)}\label{app:boundary}
\begin{lemma}\label{lem:inactive}
Term (V) in \cref{lem:decomposition} satisfies, with probability at least $1-\delta$,
\begin{align}
\text{(V)}\lesssim HS^2AB+H\iota.
\end{align}
\end{lemma}
\begin{proof}
Let $X_t$ be the indicator that the trajectory of episode $t$ traverses an edge outside $\calE_t$.
Each of the $S^2A$ edges is traversed at most $B$ times while it is outside $\calE_t$, so we have $\sum_{t\leq T}X_t\leq S^2AB$.
Hence, with probability at least $1-\delta$,
\begin{align}
\text{(V)}=\sum_{t<\tau}\inpr*{q^{\pi_t}-q^{P_{\calE_t},\pi_t},\ell_t}
&\leq\sum_{t<\tau}\sum_{s,a}\prn*{q^{\pi_t}(s,a)-q^{P_{\calE_t},\pi_t}(s,a)}\\
&=\sum_{t<\tau}\E_t\brk*{\sum_{s,a}\prn*{\ind\set{(s_{t,h(s)},a_{t,h(s)})=(s,a)}-\I_t(s,a;\calE_t)}}\\
&\leq H\sum_{t<\tau}\E_t[X_t]\\
&\leq H\prn*{2\sum_{t<\tau}X_t+4\iota}\tag{by \cref{lem:bounded-sum}}\\
&\lesssim HS^2AB+H\iota.
\end{align}
\end{proof}

\subsection{Proof of the main theorem}\label{app:main-proof}
\begin{proof}[Proof of \cref{thm:main}]
If $T\leq SA\iota^4$, then $\Reg_T\leq HT\leq H\sqrt{SAT}\iota^2$, and hence we assume $T>SA\iota^4$.
By \cref{lem:stopping,lem:decomposition,lem:ftrl,lem:delay,lem:source,lem:comparator,lem:inactive} and a union bound, with probability at least $1-11\delta$,
\begin{align}
&\Reg_T\\
&\lesssim\underbrace{\frac{H\iota}{\eta}+\eta HSAT\iota^2+H\sqrt{SAT}\iota^2+HSA\iota^2}_{\text{(I)}}\\
&\quad+\underbrace{\eta HSAT\iota^3+\eta^2HS^3A^3T\iota^4+S^3A^2\iota^2+\prn{H\iota+\eta\sqrt HS^{3/2}A^{3/2}\iota^3}\prn{\sqrt{T\iota}+\iota}}_{\text{(II)}}\\
&\quad+\underbrace{\eta HSAT\iota^3+S^3A^2\iota^2+H\sqrt T\iota^{3/2}+\eta H^{3/2}SA\iota^4+\sqrt{SAT}\iota}_{\text{(III)}}\\
&\quad+\underbrace{H\sqrt{SAT}\iota^2+HSA\iota^2}_{\text{(IV)}}+\underbrace{HS^2AB+H\iota}_{\text{(V)}}\\
&\lesssim\frac{H\iota}{\eta}+\eta HSAT\iota^3+\eta^2HS^3A^3T\iota^4+HS^2AB+S^3A^2\iota^2+H\sqrt{SAT}\iota^2,\label{eq:raw-rate}
\end{align}
where we used the following bounds from $\sqrt T>\sqrt{SA}\iota^2$, $\eta\leq1/(c_0\sqrt{SAT}\iota)$, and $H\leq S$,
\begin{align}
&\eta\sqrt HS^{3/2}A^{3/2}\sqrt T\iota^{7/2}\lesssim\sqrt HSA\iota^{5/2}\leq H\sqrt{SAT}\iota^2,\\
&\eta\sqrt HS^{3/2}A^{3/2}\iota^4+\eta H^{3/2}SA\iota^4\lesssim\frac{\prn*{\sqrt HSA+H^{3/2}\sqrt{SA}}\iota^3}{\sqrt T}\lesssim H\sqrt{SA}\iota\leq H\sqrt{SAT}\iota^2.
\end{align}
Since $\eta=1/(2500c_0(\sqrt{SAT}+\sqrt{HS^3}A)\iota)$ and $B=O(HSA\max\set{SA,H^2}\iota)$ by \cref{alg:main}, we have
\begin{align}
&\frac{H\iota}{\eta}+\eta HSAT\iota^3+\eta^2HS^3A^3T\iota^4\lesssim H\sqrt{SAT}\iota^2+H^{3/2}S^{3/2}A\iota^2+HS^2A^2\iota^2,\\
&HS^2AB\lesssim H^2S^3A^2\max\set{SA,H^2}\iota.
\end{align}
Substituting these into \cref{eq:raw-rate} and using $H\leq S$, we have
\begin{align}\label{eq:separated-rate}
\Reg_T\lesssim H\sqrt{SAT}\iota^2+H^2S^3A^2\max\set{SA,H^2}\iota+S^3A^2\iota^2.
\end{align}

If $H^2S^3A^2\max\set{SA,H^2}\leq H\sqrt{SAT}$, then $S^3A^2\leq H\sqrt{SAT}$, and \cref{eq:separated-rate} gives $\Reg_T\lesssim H\sqrt{SAT}\iota^2$.
Otherwise, $T\leq H^2S^5A^3\max\set{SA,H^2}^2$, so $\iota\lesssim\log(SA/\delta)$ and the last term $S^3A^2\iota^2$ is at most $H\sqrt{SAT}\iota^2$ if $S^3A^2\leq H\sqrt{SAT}$.
If instead $S^3A^2>H\sqrt{SAT}$, then $T>SA\iota^4$ gives $HSA\iota^2<H\sqrt{SAT}<S^3A^2$ and $S^3A^2\iota^2\leq H^2S^3A^2\max\set{SA,H^2}\iota$.
Hence \cref{eq:separated-rate} gives $\Reg_T\lesssim H\sqrt{SAT}\iota^2+H^2S^3A^2\max\set{SA,H^2}\log(SA/\delta)$.
Therefore, with probability at least $1-11\delta$,
\begin{align}
\Reg_T\lesssim H\sqrt{SAT}\log^2\prn*{\frac{SAT}{\delta}}+H^2S^3A^2\max\set{SA,H^2}\log\prn*{\frac{SA}{\delta}}.
\end{align}
\end{proof}

\end{document}